\documentclass[lettersize,journal]{IEEEtran}
\usepackage{amsmath,amsfonts}
\usepackage{algorithmic}
\usepackage{array}
\usepackage[caption=false,font=normalsize,labelfont=sf,textfont=sf]{subfig}
\usepackage{textcomp}
\usepackage{stfloats}
\usepackage{url}
\usepackage{verbatim}
\usepackage{graphicx}
\usepackage{booktabs}
\usepackage{hyperref}
\usepackage[table,xcdraw]{xcolor}
\usepackage{cite}
\usepackage{soul, color, xcolor}
\usepackage{tcolorbox} 
\usepackage{multirow}
\usepackage{overpic}
\usepackage{algorithm,algorithmic}
\usepackage{subfig} 
\usepackage{colortbl}  
\usepackage{xpatch}
\usepackage{balance}
\usepackage{amssymb}
\usepackage{makecell}
\usepackage{siunitx} 
\begin{document}

\title{PaAgent: Portrait-Aware Image Restoration Agent via Subjective-Objective Reinforcement Learning}

\author{

Yijian~Wang,~
Qingsen~Yan,~
Jiantao~Zhou,~~
Duwei Dai, ~
Wei~Dong~

\thanks{*Corresponding author: Qingsen Yan.}

\thanks{Yijian Wang and Qingsen Yan are with the School of Computer Science, Northwestern Polytechnical University, Xi’an 710072, China, and Qingsen Yan is also with the Shenzhen Research Institute of Northwestern Polytechnical University, Shenzhen 518057, China (e-mail:  wangyijian1017@163.com; qingsenyan@nwpu.edu.cn).}

\thanks{Jiantao Zhou is with the State Key Laboratory of Internet of Things for Smart City, University of Macau, Macau 999078, China (e-mail: jtzhou@um.edu.mo).}

\thanks{Duwei Dai is with the National-Local Joint Engineering Research Center of Biodiagnosis and Biotherapy, the Second Affiliated Hospital of Xi’an Jiaotong University, Xi’an 710006, China, and also with Xi’an Jiaotong University, Xi’an 710049, China (e-mail: duweidai@xjtu.edu.cn).}

\thanks{Wei Dong is with the College of Information and Control Engineering, Xi’an University of Architecture and Technology, Xi’an 710064, China (e-mail: dongwei156@outlook.com).}

}

\maketitle

\begin{abstract}

Image Restoration (IR) agents, leveraging multimodal large language models to perceive degradation and invoke restoration tools, have shown promise in automating IR tasks.
However, existing IR agents typically lack an insight summarization mechanism for past interactions, which results in an exhaustive search for the optimal IR tool. 
To address this limitation, we propose a portrait-aware IR agent, dubbed PaAgent, which incorporates a self-evolving portrait bank for IR tools and Retrieval-Augmented Generation (RAG) to select a suitable IR tool for input.
Specifically, to construct and evolve the portrait bank, the PaAgent continuously enriches it by summarizing the characteristics of various IR tools with restored images, selected IR tools, and degraded images.
In addition, the RAG is employed to select the optimal IR tool for the input image by retrieving relevant insights from the portrait bank.
Furthermore, to enhance PaAgent's ability to perceive degradation in complex scenes, we propose a subjective-objective reinforcement learning strategy that considers both image quality scores and semantic insights in reward generation, which accurately provides the degradation information even under partial and non-uniform degradation.
Extensive experiments across 8 IR benchmarks, covering six single-degradation and eight mixed-degradation scenarios, validate PaAgent's superiority in addressing complex IR tasks.
Our project page is \href{https://wyjgr.github.io/PaAgent.html}{PaAgent}.
\end{abstract}


\section{Introduction}
\IEEEPARstart{I}{mage} Restoration (IR) aims to recover high-quality images from low-quality ones that are degraded by 
snow \cite{CSD}, noise \cite{BSD68}, haze \cite{RESIDE}, rain \cite{MPRNet}, low-light \cite{LOL-v1}, blur \cite{Zhao_2}, or composite degradation patterns \cite{OneRestore}.
Task-specific IR methods \cite{MPRNet, Restormer, DehazeFormer, OKNet, FocalNet, Dehamer, DRBNet, InvDSNet, CIDNet, Zero-DCE, Zero-DCE++, LLFlow} focus on handling a single degradation; however, real-world images often suffer from multiple degradations \cite{Zhao_1}, which significantly limits the practicality of these methods.
To address this, numerous All-in-One (AiO) IR methods \cite{PromptIR, GridFormer, RAM, NDR-Restore, DGSolver, AWRaCLe, DFPIR, DA-RCOT, CPLIR, AdaIR} have been presented, which can handle multiple degradations simultaneously within a unified model by prompt learning or instruction tuning.
Unfortunately, these AiO IR methods \cite{PromptIR, GridFormer, RAM, NDR-Restore, DGSolver, AWRaCLe, DFPIR, DA-RCOT, CPLIR, AdaIR} struggle with a trade-off between generalization and restoration accuracy due to potential conflicts among various optimization objectives.

\begin{figure}[t]
    \centering
    \includegraphics[width=1\linewidth]{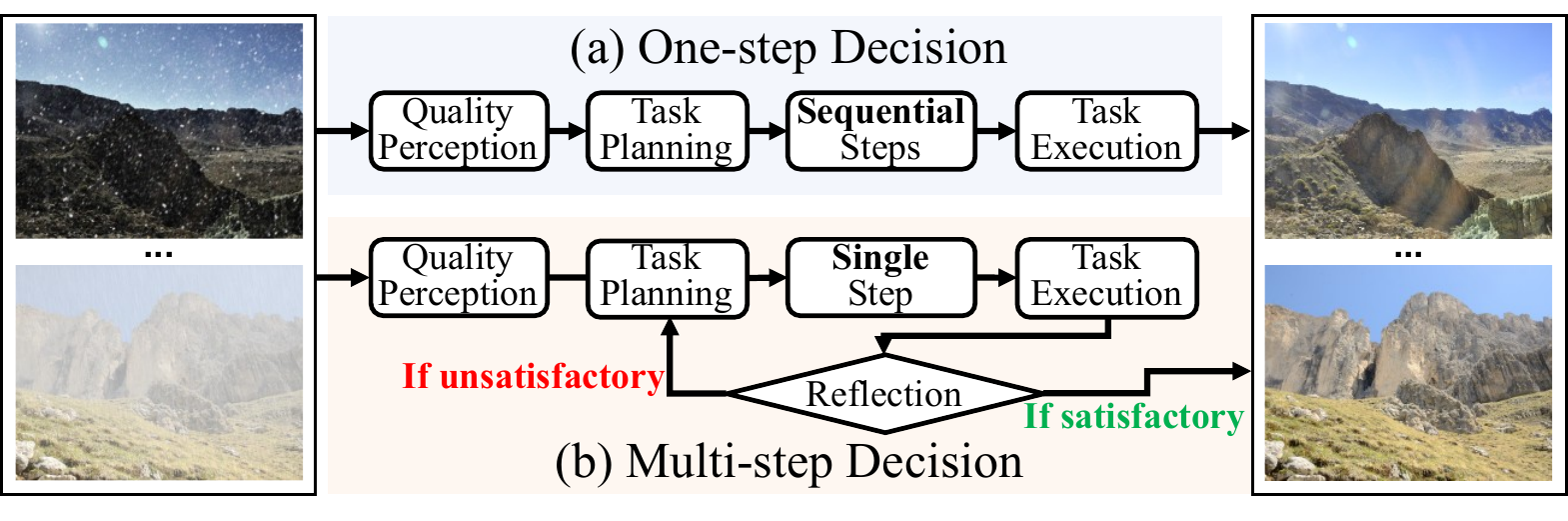}
    \caption{Execution workflows of existing agent-based IR methods \cite{RestoreAgent, AgenticIR, HAIR, MAIR, Q-Agent, 4KAgent, JarvisIR, HFLS-Weather}. (a) One-step decision methods follow a rigid sequential pipeline, which is prone to error accumulation. (b) Multi-step decision methods introduce a reflection-based iterative mechanism for evaluating and refining restoration results.}
    \label{Motivation}
\end{figure}

\begin{figure}[t]
    \centering
    \includegraphics[width=1\linewidth]{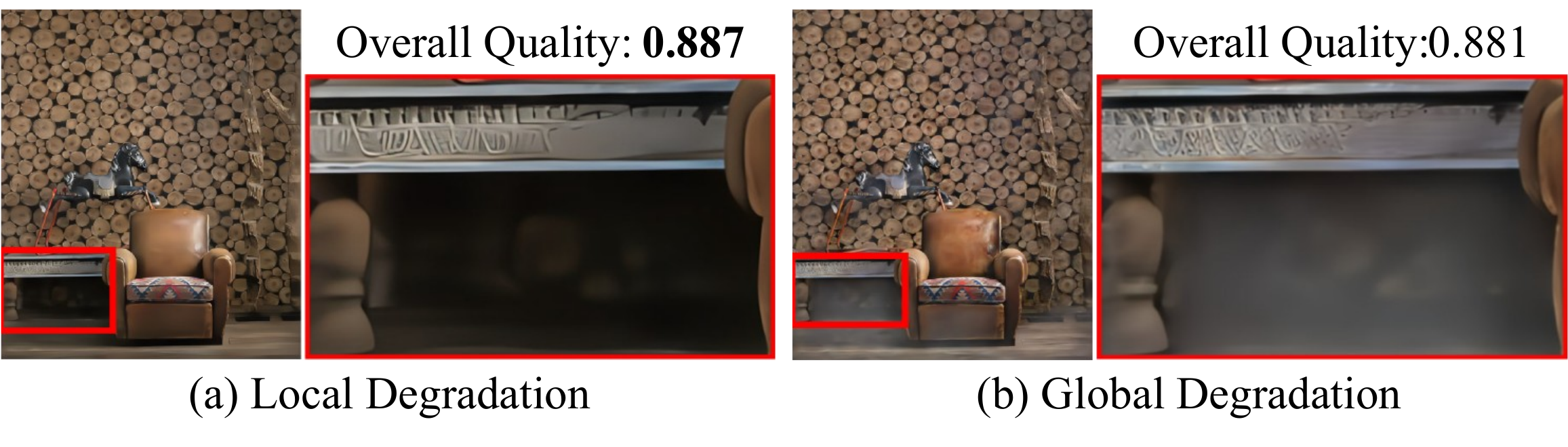}
    \caption{The illustration of quality score ambiguity. Despite possessing similar overall quality scores (calculated following \cite{Q-Agent}), these two images exhibit distinct spatial degradation patterns.}
    \label{Motivation2}
\end{figure}

Recently, agent-based IR methods  \cite{RestoreAgent, AgenticIR, HAIR, MAIR, Q-Agent, 4KAgent, JarvisIR, HFLS-Weather} have demonstrated remarkable performance in complex scenarios.
As shown in Fig. \ref{Motivation}, agent-based IR methods \cite{RestoreAgent, AgenticIR, HAIR, MAIR, Q-Agent, 4KAgent, JarvisIR, HFLS-Weather} can be broadly categorized into one-step \cite{MAIR, Q-Agent, 4KAgent, JarvisIR, HFLS-Weather} and multi-step decision \cite{RestoreAgent, AgenticIR, HAIR} methods.
One-step decision methods \cite{MAIR, Q-Agent, 4KAgent, JarvisIR, HFLS-Weather} mimics human expertise to perceive quality degradation, arrange a sequence of restoration tasks, and invoke IR tools.
However, any misjudgment at the quality assessment can cause error accumulation throughout the entire execution chain.
Multi-step decision methods \cite{RestoreAgent, AgenticIR, HAIR} employ an iterative framework with a rollback mechanism, which evaluates the restored output after each step to decide whether to execute the next task or rollback the current operation for a retry.
This iterative framework provides better scalability and flexibility in handling complex degradations.
Nevertheless, the aforementioned methods \cite{RestoreAgent, AgenticIR, HAIR, MAIR, Q-Agent, 4KAgent, JarvisIR, HFLS-Weather} force the agent into exhaustive exploration of IR tools in the task execution, even when for simple degradations, because they treat each restoration instance in isolation and fail to leverage insights from historical interactions.
Furthermore, existing agent-based IR methods \cite{RestoreAgent, AgenticIR, HAIR, MAIR, Q-Agent, 4KAgent, JarvisIR, HFLS-Weather} typically integrate multiple NR-IQA scores into an overall quality score as optimization targets when improving degradation perception capabilities \cite{Q-Agent}.
Although the overall quality score represents the visual quality of images to a certain extent, a single score is insufficient to capture the spatial distribution of degradations.
As shown in Fig. \ref{Motivation2}, two images possess a similar overall quality score, yet one exhibits significant degradation in local regions while the other shows slight hazing globally.
Such ambiguity in quality assessment often leads to erroneous decision-making, as the agent cannot distinguish between localized severe degradations and globally minor ones.

In this paper, we propose PaAgent, a portrait-aware IR agent designed to overcome the above limitations through two key innovations: 
1) a tool portrait bank that enables PaAgent to choose the suitable IR tool by leveraging past interaction insights.
2) a Subjective-Objective Reinforcement Learning (SORL) strategy that enhances PaAgent's degradation perception capabilities through reward signals derived from both image quality scores and semantic insights.
Specifically, we construct a tool portrait bank that stores extensive historical interaction insights from various IR tools. 
The portrait bank encompasses 14 restoration tools, 14 degradation scenarios, and over 100,000 interaction insights.
By retrieving relevant insights via RAG \cite{RAG}, PaAgent preserves the scalability and flexibility of multi-step decision methods while avoiding excessive IR trials and rollbacks.
In addition, each restoration record generated by the PaAgent is dynamically stored in the tool portrait bank, allowing PaAgent to improve its restoration performance over prolonged usage.
To improve PaAgent’s degradation perception, we design a SORL strategy comprising two core components: a Multimodal Large Language Model (MLLM)-based reward generator and the Group Relative Policy Optimization (GRPO) \cite{DeepSeekMath} algorithm.  
The MLLM-based reward generator is designed to generate fine-grained reward signals by synthesizing both MLLM’s semantic insights and quality scores from No-Reference Image Quality Assessment (NR-IQA) metrics \cite{CLIP-IQA, Hyper-IQA, NIQE, CPBD, BRISQUE, MUSIQ, LIQE}.
The GRPO \cite{DeepSeekMath} algorithm is employed to translate the fine-grained reward signals into intra-group relative advantages, which allows PaAgent to refine its policy (i.e., the next-task selection and degradation information) by comparing multiple restoration trajectories, ensuring that the policy updates are driven by a nuanced understanding of both quality scores and semantic insights.

In summary, our main contributions are as follows:

\begin{itemize}
    \item  We propose an efficient PaAgent method, which is the first work to tackle image restoration using an agent with insight summarization mechanism. Benefiting from RAG, PaAgent achieves informed tool invocation by retrieving historical interaction insights, effectively avoiding excessive tool trials.
    
    \item We design a Subjective-Objective Reinforcement Learning (SORL) strategy to sharpen the degradation perception capability of PaAgent. SORL enables PaAgent not only to perceive the overall quality of images but also to understand the semantic characteristics of local degradation, thereby achieving precise decision-making in complex scenarios.
    
    \item We conduct extensive experiments to demonstrate that the proposed PaAgent method outperforms several leading approaches in both subjective results and objective assessments. Additionally, we show that PaAgent effectively overcomes error accumulation and premature termination across diverse degradation scenarios.
\end{itemize}

\section{Related Work}
\subsection{All-in-One Image Restoration}
To handle multiple distortions with a unified model weight, numerous AiO IR models have been proposed.
Inspired by prompt learning in natural language processing, Potlapalli \emph{et al.} \cite{PromptIR} presented a prompt-based learning approach for dynamically handling diverse image degradations.
Qin \emph{et al.} \cite{RAM} developed a mask image modeling framework that effectively learns the inherent content of images while ensuring high-quality reconstruction results.
By leveraging the advantages of ordinary differential equation solvers, Wang \emph{et al.} \cite{DGSolver} introduced a diffusion generalist solver with universal posterior sampling for image restoration.
Rajagopalan \emph{et al.} \cite{AWRaCLe} proposed a visual context learning to replace the prompt-based learning in \cite{PromptIR}, thereby extracting degradation features more effectively.
Motivated by the observation that different degradation types uniquely impact image content across distinct frequency subbands, Cui \emph{et al.} \cite{AdaIR} proposed an AiO IR network based on frequency mining and modulation.
Although these AiO IR methods achieve significant improvements by unifying multiple degradation tasks within a single framework, training these methods is inherently difficult due to the optimization conflicts arising from diverse restoration objectives.
As a result, it often leads to a sub-optimal balance between generalization ability and reconstruction performance.

\subsection{Agent-based Image Restoration}
With the rise of agents, IR tasks have ushered in a new research line.
As a pioneering work, Chen \emph{et al.} \cite{RestoreAgent} fine-tuned an MLLM on synthetic datasets to enable autonomous degradation assessment and tool execution.
Another parallel work by Zhu \emph{et al.} \cite{AgenticIR} introduced a rollback-based agent framework to select a suitable IR tool at each decision step from a large tool repository.
To balance execution efficiency and restoration performance, Li \emph{et al.} \cite{HAIR} proposed a hybrid IR agent that adaptively invokes perception agents of varying complexity based on user prompts.
Jiang \emph{et al.} \cite{MAIR} proposed a three-stage IR Agent based on the inherent physical order of image degradation. 
Zhou \emph{et al.} \cite{Q-Agent} proposed a greedy strategy driven by NR-IQA metrics \cite{Q-Agent} to explore high-quality restoration sequences.
Lin \emph{et al.} \cite{JarvisIR} designed a training strategy based on Supervised Fine-Tuning (SFT) and human feedback alignment to enhance the agent's robustness in autonomous driving systems.
Liu \emph{et al.} \cite{HFLS-Weather} proposed a perturbation-driven image quality optimization strategy to improve the agent's decision-making capabilities in terms of tool selection and execution ordering.
However, these agent-based IR methods lack insight-driven tool invocation and powerful degradation perception capabilities.
In contrast, the proposed PaAgent is equipped with a tool portrait bank that stores extensive historical interaction insights, enabling it to directly invoke the appropriate IR tool.
Moreover, our SORL strategy empowers PaAgent with fine-grained degradation perception capabilities by constructing an optimization objective jointly guided by semantic insights and quality scores.

\begin{figure*}[ht]
    \centering
    \includegraphics[width=0.9\linewidth]{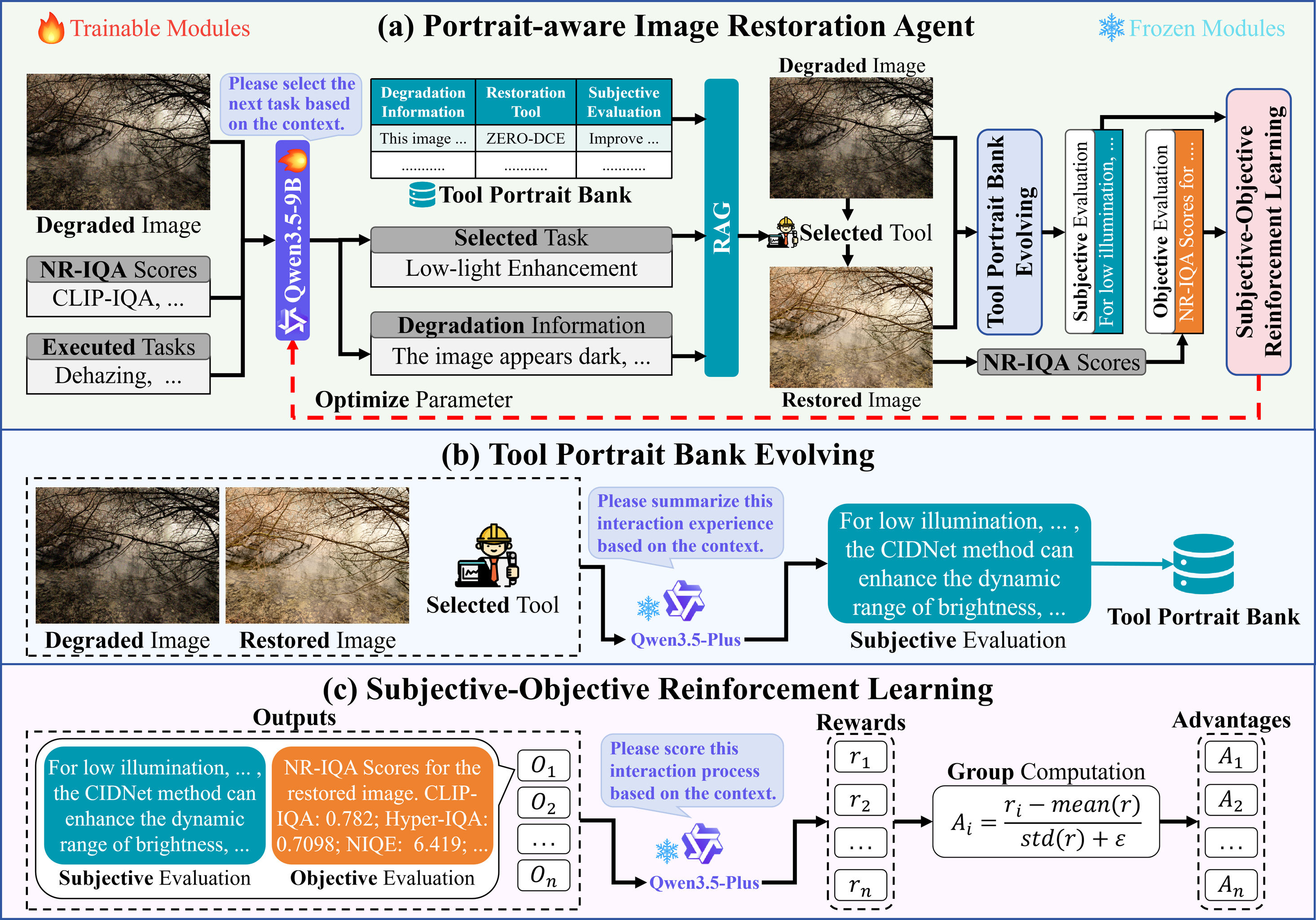}
    \caption{Overview of the proposed PaAgent architecture. 
    (a) illustrates the entire workflow of PaAgent, which leverages Qwen3.5-9B \cite{qwen3.5} for degradation perception and task recommendation, followed by an RAG \cite{RAG} module that queries the constructed tool portrait bank for optimal tool invocation. 
    (b) depicts the evolution of the tool portrait bank, where interaction insights are summarized by Qwen3.5-Plus \cite{qwen3.5} and stored for future utilization. 
    (c) shows the SORL strategy, which integrates MLLM's insights and NR-IQA scores \cite{CLIP-IQA, Hyper-IQA, NIQE, CPBD, BRISQUE, MUSIQ, LIQE} via Qwen3.5-Plus \cite{qwen3.5} to generate reward signals, thereby driving the GRPO \cite{DeepSeekMath} algorithm to fine-tune Qwen3.5-9B. \cite{qwen3.5}}
    \label{Framework}
\end{figure*}

\section{Proposed Method}
In this section, we first present the overall architecture of the proposed PaAgent,
and we then detail the proposed tool portrait bank and subjective-objective reinforcement learning strategy.

\begin{figure}[ht]
    \centering
    \includegraphics[width=1\linewidth]{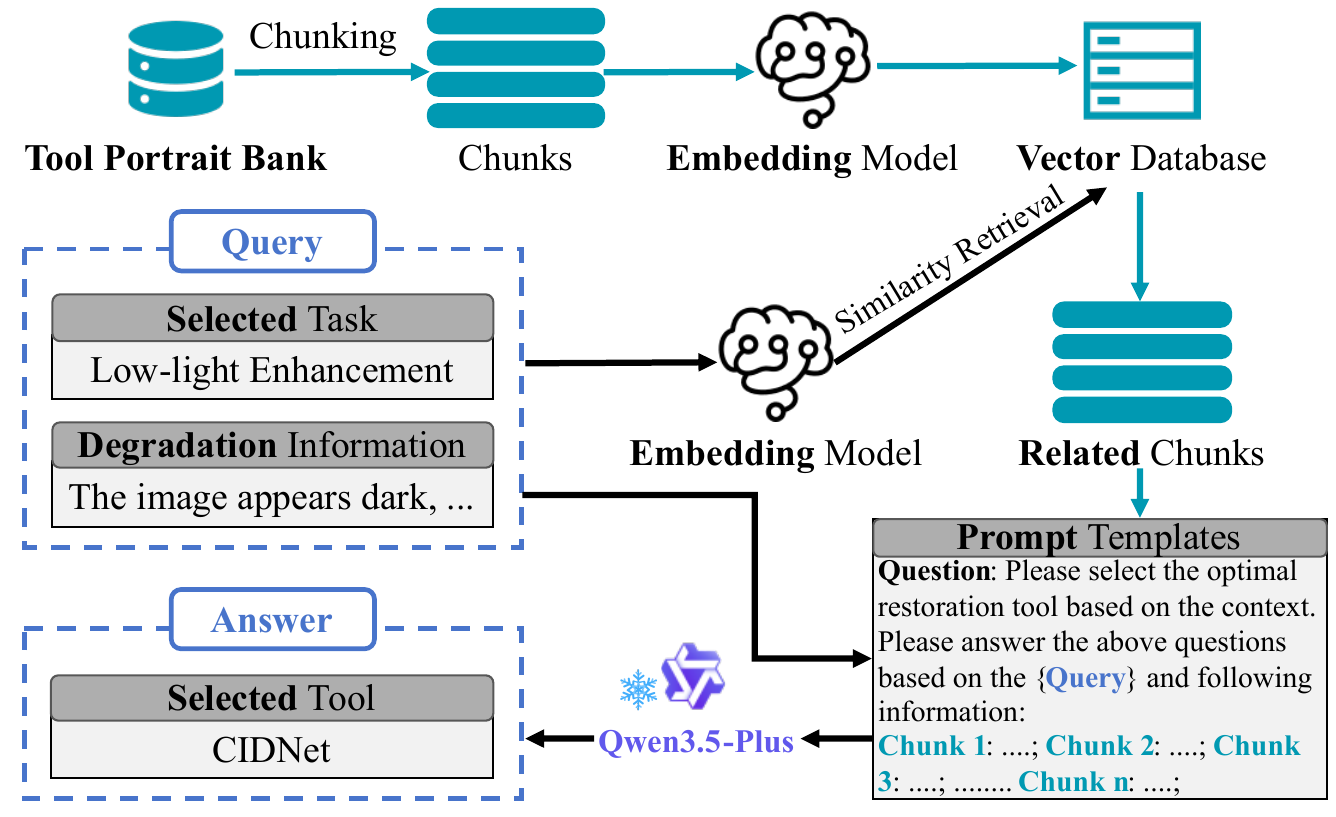}
    \caption{The illustration of the RAG \cite{RAG} process. It mainly consists of two phases: (1) offline knowledge base construction (upper stream), where the tool portrait bank is chunked, embedded via an embedding model, and stored in a vector database; and (2) online retrieval and reasoning (lower stream), where the query is embedded to retrieve relevant chunks through similarity search. The retrieved chunks are then integrated into prompt templates and fed into the QWen3.5-Plus \cite{qwen3.5} to select the IR tool. }
    \label{RAG}
\end{figure}

\subsection{Overall Architecture}
The overall architecture of our proposed Portrait-aware Agent (PaAgent) is shown in Fig. \ref{Framework}. 
Given a degraded image, we first leverage the Qwen3.5-9B \cite{qwen3.5} model to conduct comprehensive perception, which additionally incorporates NR-IQA scores \cite{CLIP-IQA, Hyper-IQA, NIQE, CPBD, BRISQUE, MUSIQ, LIQE} and previously executed tasks.
This perception stage yields degradation information and a selected restoration task, both of which serve as queries for the Retrieval-Augmented Generation (RAG) module \cite{RAG}.
As illustrated in Fig. \ref{RAG}, these queries are first encoded via an embedding model to perform similarity retrieval against the vector database, which stores chunked interaction insights from the tool portrait bank. 
The top-$k$ relevant chunks retrieved from the vector database are dynamically concatenated with the query context into the prompt template.
These enriched prompts are subsequently fed into Qwen3.5-Plus \cite{qwen3.5} for context-aware reasoning, enabling the inference of the most appropriate IR tool.
Ultimately, the selected IR tool is adopted to obtain restored images.

\subsection{Tool Portrait Bank}
Agent-based IR methods have demonstrated impressive capabilities in handling complex degradations;  however, they fail to leverage historical interaction insights, which leads to excessive tool trials and rollback operations.
To address this limitation, we develop a tool portrait bank designed to store past interaction insights that map diverse degradation scenarios to their corresponding restoration solutions.
As illustrated in Fig. \ref{Framework} (a), each entry in the portrait bank is organized as a triplet comprising degradation information, the applied restoration tool, and a subjective evaluation of the restoration quality.
To support efficient retrieval within the RAG \cite{RAG} framework, these structured triplets undergo a preprocessing pipeline as depicted in Fig. \ref{RAG}.
These interaction insights are first segmented into semantic chunks via a chunking strategy, which are subsequently encoded by an embedding model into dense vector representations and organized within a vector database. 
This vectorized organization enables semantic similarity-based retrieval of relevant chunks given a degradation query. 
Consequently, the search space for candidate IR tools is effectively narrowed, allowing our PaAgent to make informed decisions and avoid redundant explorations during the restoration process.
Furthermore, our constructed tool portrait bank is equipped with an evolution mechanism to ensure the continuous accumulation of interaction insights. 
As shown in Fig. \ref{Framework} (b), once a restoration task is performed, the triple consisting of the degraded image, the restored image, and the suggested tool is processed by the Qwen3.5-Plus \cite{qwen3.5} model to summarize the interaction insight and provide a subjective evaluation.
The new insight is then integrated into the tool portrait bank, facilitating a self-evolving knowledge base that progressively enhances PaAgent's proficiency in handling complex restoration tasks.

\subsection{Subjective-Objective Reinforcement Learning}
Training agent-based IR methods that rely on numerical scores makes it difficult to capture the subtle impact of different degradation patterns on image quality, thereby limiting their ability to accurately describe degradation information and make task-specific decisions.
To this end, we employ the GRPO \cite{DeepSeekMath} algorithm, a transformative advancement in RL distinguished by its ability to refine model policies without a complex value network, to enhance the accuracy of degradation perception.
By utilizing the mean reward of a sampled group as a baseline, the GRPO \cite{DeepSeekMath} algorithm effectively identifies superior reasoning patterns and penalizes suboptimal outputs, achieving a more efficient policy alignment process. 
Specifically, we propose a Subjective-Objective Reinforcement Learning (SORL) strategy, as shown in Fig. \ref{Framework} (c).
Given a restored image, its subjective evaluation is first obtained by utilizing an MLLM (i.e., Qwen3.5-Plus \cite{qwen3.5}) to summarize the triplet consisting of the degraded image, the selected IR tool, and the restored image itself.
Then, inspired by the \cite{Q-Agent, HFLS-Weather}, multiple NR-IQA metrics, including CLIP-IQA \cite{CLIP-IQA}, Hyper-IQA \cite{Hyper-IQA}, NIQE \cite{NIQE}, CPBD \cite{CPBD}, BRISQUE \cite{BRISQUE}, MUSIQ \cite{MUSIQ}, and LIQE \cite{LIQE}, are used to compute its objective quality scores and form an objective evaluation.
These subjective and objective evaluations are then integrated to constitute the $i$-th output $O_i$ within a sampled group. 
As illustrated in Fig. \ref{Framework}(c), for each restoration task, PaAgent generates a group of $n$ such outputs $\{O_1, O_2, \dots, O_n\}$.
These outputs $\{O_1, O_2, \dots, O_n\}$ are fed into the reward generator (i.e., Qwen3.5-Plus \cite{qwen3.5}), which performs dual-perspective analysis and assigns a binary reward $r_i \in \{0, 1\}$ to each $O_i$.
A reward of $r_i = 1$ indicates that the selected restoration task effectively addressed the identified degradation and produced a result consistent with both subjective visual preferences and high objective quality scores.
Conversely, a reward of $r_i = 0$ signifies that the chosen restoration task either failed to resolve the image defect or yielded a perceptually unsatisfactory outcome.
Based on these binary reward signals $\{r_1, r_2, \dots, r_n\}$, we first calculate the normalized advantage $A_i$ for each sample $i$:
\begin{equation}
    A_i = \frac{r_i - mean(r)}{std(r) + \epsilon},
    \label{eq:reward_advantage}
\end{equation}
where $mean(\cdot)$ denotes the mean operation, $std(\cdot)$ represents the standard deviation, and $\epsilon$ is a small positive constant added for numerical stability. 
Next, we leverage the normalized advantages $\{A_1, A_2, \dots, A_n\}$ within the GRPO \cite{DeepSeekMath} algorithm to update the current policy model $\pi_\theta$ (i.e., Qwen3.5-9B~\cite{qwen3.5}), where $\theta$ denotes the trainable parameters of the model. 
The core update mechanism follows a policy gradient scheme that maximizes the expected advantage for actions taken by the policy.
Formally, we first define the probability ratio between the current policy $\pi_\theta$ and the old policy $\pi_{\theta_{old}}$ as:
\begin{equation}
\rho_i(\theta) = \frac{\pi_\theta(T_i \mid I_{deg})}{\pi_{\theta_{old}}(T_i \mid I_{deg})},
\end{equation}
where $\pi_\theta(T_i \mid I_{deg})$ denotes the probability of selecting restoration task $T_i$ given a degraded input image $I_{deg}$. 
Thereafter, the clipped surrogate loss $\mathcal{L}_{CLIP}(\theta)$ is formulated as:
\begin{equation}
    \mathcal{L}_{CLIP}(\theta) = \frac{1}{n} \sum_{i=1}^{n} \min\left[ \rho_i(\theta) A_i, \text{clip}(\rho_i(\theta), 1 - \epsilon, 1 + \epsilon) A_i \right],
\end{equation}
where the $\text{clip}(\cdot)$ operation constrains the ratio to $[1-\epsilon, 1+\epsilon]$ to prevent overly large policy steps, and the $\min(\cdot)$ operator ensures pessimistic updates. 
Positive advantages $A_i$ reinforce the selection of task $T_i$, whereas negative advantages discourage similar strategies.
In practice, to prevent excessive updates and maintain training stability, we adopt a clipped surrogate objective augmented with a KL-divergence regularization term. 
Therefore, the final loss function $\mathcal{L}_{final}(\theta)$ is obtained by incorporating a KL-divergence penalty:
\begin{equation}
    \mathcal{L}_{final}(\theta) = -\mathcal{L}_{CLIP}(\theta) + \beta \cdot \mathbb{D}_{KL}\left[ \pi_{\theta} \| \pi_{ref} \right],
\end{equation}
where the hyperparameter $\beta$ is the regularization coefficient and the KL-divergence term $\mathbb{D}_{KL}\left[ \pi_\theta \| \pi_{ref} \right]$ \cite{DeepSeekMath} penalizes deviations from the reference policy $\pi_{ref}$ (e.g., the initial pre-trained model). 
To ensure computational stability, we employ the following unbiased estimator~\cite{DeepSeekMath}:
\begin{equation}
    \mathbb{D}_{KL}[\pi_\theta \| \pi_{ref}] = \frac{\pi_{ref}(T_i|I_{deg})}{\pi_\theta(T_i|I_{deg})} - \log \frac{\pi_{ref}(T_i|I_{deg})}{\pi_\theta(T_i|I_{deg})} - 1,
\end{equation}
which guarantees non-negative values and exhibits lower variance than standard estimators. 
Consequently, the GRPO \cite{DeepSeekMath} algorithm enables the policy model $\pi_\theta$ to learn effectively from these intra-group advantages rather than absolute reward scales, thereby enhancing training stability while maintaining distributional consistency with the reference model $\pi_{ref}$.

\section{Experiments}

\subsection{Experiment Settings}
\noindent
\textbf{Implementation Details.} 
We adopt LoRA \cite{Lora}, a parameter-efficient fine-tuning technique, to adapt the projection layers within all self-attention modules of both the vision encoder and the LLM.
For our experimental setup, we configure the LoRA rank to 16 and employ the SWIFT framework \cite{SWIFT} to facilitate the training pipeline.
The PaAgent is trained for 10 epochs on 2 NVIDIA A100 GPUs with a batch size of 4, using the Adam optimizer with a learning rate of $10^{-6}$.
For the IR toolkit of PaAgent, we integrate a diverse collection of pre-trained models: 
denoising (Restormer \cite{Restormer} and MPRNet \cite{MPRNet}), 
dehazing (OKNet \cite{OKNet}, FocalNet \cite{FocalNet}, DehazeFormer \cite{DehazeFormer}, and Dehamer \cite{Dehamer}), 
deraining (Restormer \cite{Restormer} and MPRNet \cite{MPRNet}), 
deblurring (Restormer \cite{Restormer}, MPRNet \cite{MPRNet}, and DRBNet \cite{DRBNet}), 
desnowing (InvDSNet \cite{InvDSNet}, OKNet \cite{OKNet}, and FocalNet \cite{FocalNet}), 
low-light enhancement (ZERO-DCE \cite{Zero-DCE}, ZERO-DCE++ \cite{Zero-DCE++}, LLFlow \cite{LLFlow}, and CIDNet \cite{CIDNet}), 
and composite degradation enhancement (LEDNet \cite{LEDNet} and MoCE-IR \cite{MoCE-IR}).

\noindent
\textbf{Compared Methods.} 
We compare our PaAgent method against eleven IR approaches, including 
ten AiO IR methods (PromptIR \cite{PromptIR}, GridFormer \cite{GridFormer}, RAM \cite{RAM}, NDR-Restore \cite{NDR-Restore}, DGSolver \cite{DGSolver}, AWRaCLe \cite{AWRaCLe}, DFPIR \cite{DFPIR}, DA-RCOT \cite{DA-RCOT}, CPLIR \cite{CPLIR}, and AdaIR \cite{AdaIR}), 
and one agent-based IR method (AgenticIR \cite{AgenticIR}).  

\begin{figure*}[!ht]
    \Large
    \centering
    \resizebox{1\linewidth}{!}{
        \begin{tabular}{c@{ }c@{ }c@{ }c@{ }c@{ }c@{ }c@{ }}
            \includegraphics[height=3cm,width=4cm]{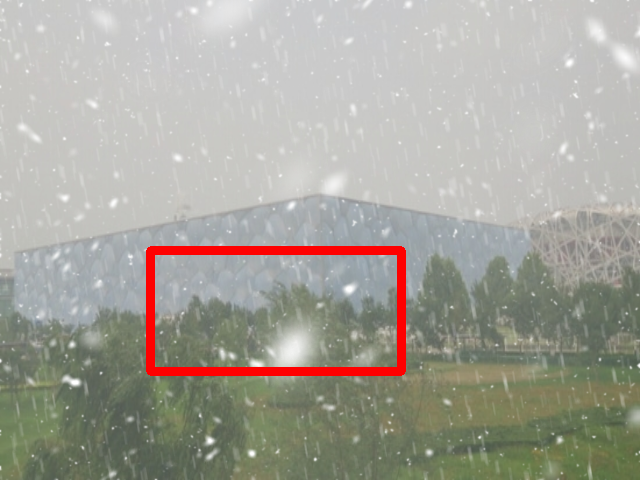} &
            \includegraphics[height=3cm,width=4cm]{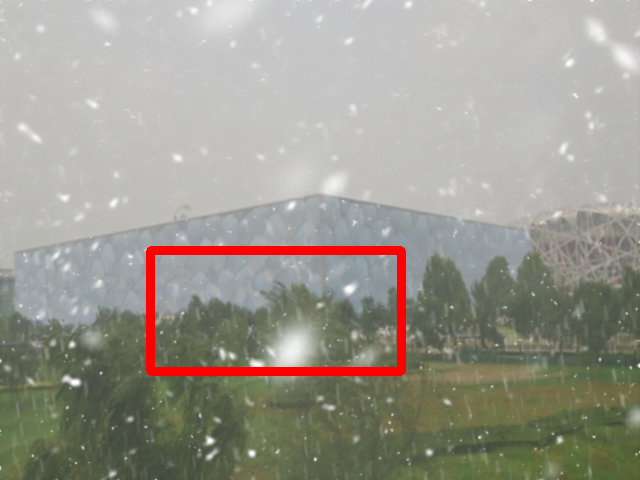} &
            \includegraphics[height=3cm,width=4cm]{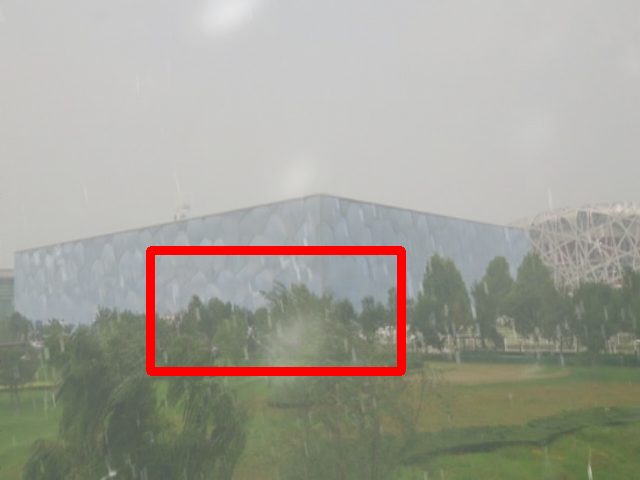} &
            \includegraphics[height=3cm,width=4cm]{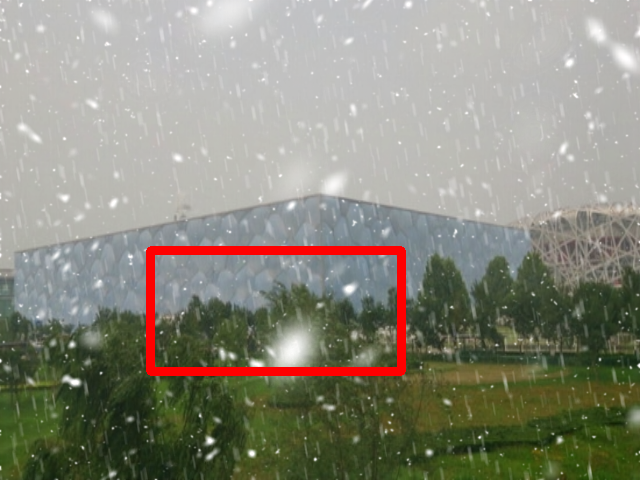} &
            \includegraphics[height=3cm,width=4cm]{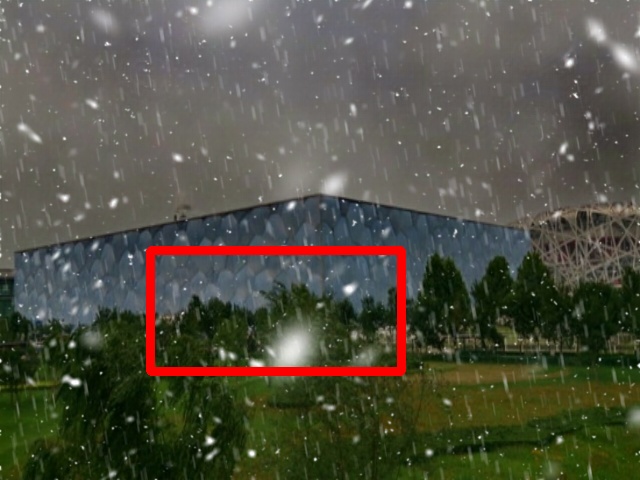} &
            \includegraphics[height=3cm,width=4cm]{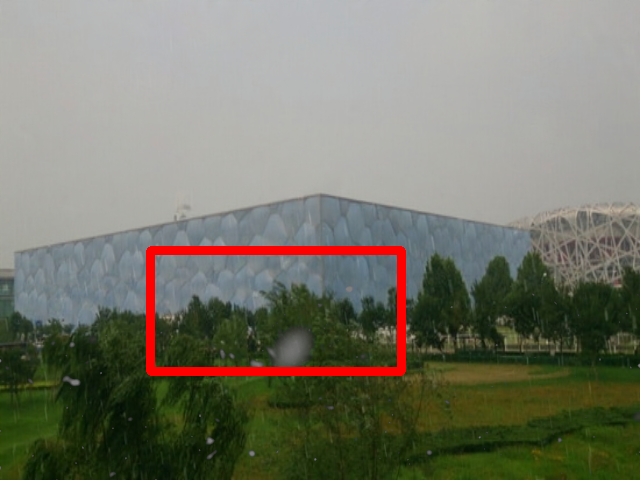} &
            \includegraphics[height=3cm,width=4cm]{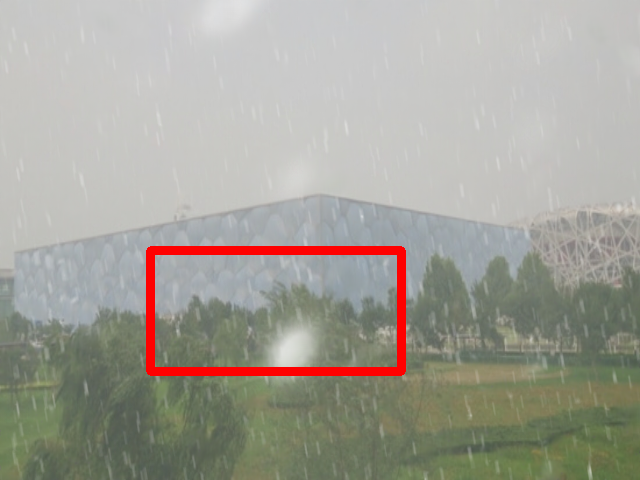} \\

            \includegraphics[height=2cm,width=4cm]{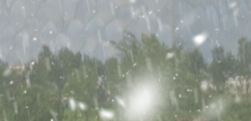} &
            \includegraphics[height=2cm,width=4cm]{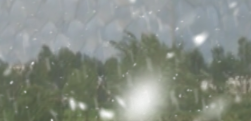} &
            \includegraphics[height=2cm,width=4cm]{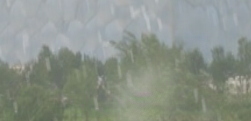} &
            \includegraphics[height=2cm,width=4cm]{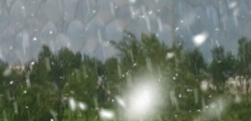} &
            \includegraphics[height=2cm,width=4cm]{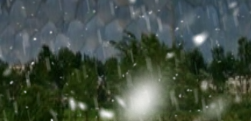} &
            \includegraphics[height=2cm,width=4cm]{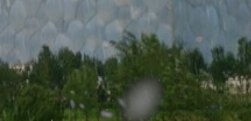} &
            \includegraphics[height=2cm,width=4cm]{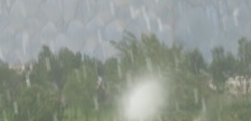} \\

            \includegraphics[height=3cm,width=4cm]{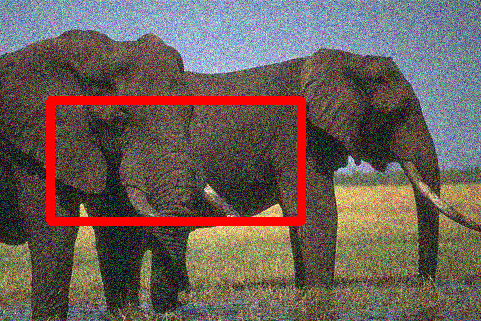} &
            \includegraphics[height=3cm,width=4cm]{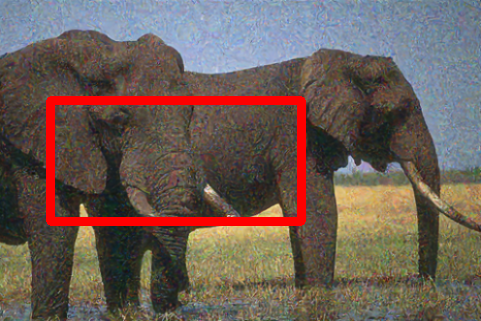} &
            \includegraphics[height=3cm,width=4cm]{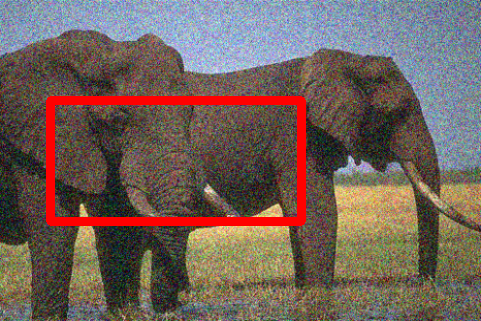} &
            \includegraphics[height=3cm,width=4cm]{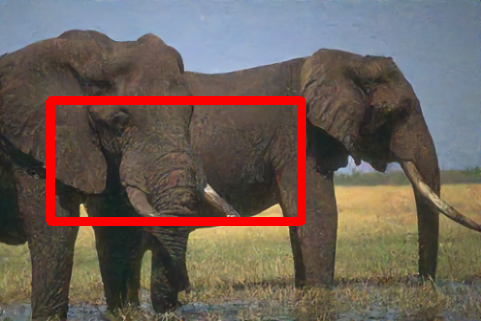} &
            \includegraphics[height=3cm,width=4cm]{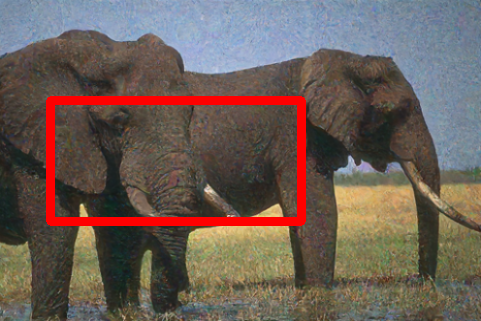} &
            \includegraphics[height=3cm,width=4cm]{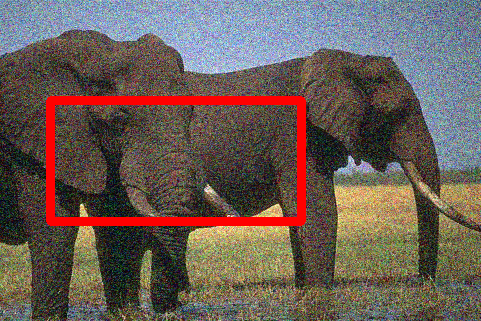} &
            \includegraphics[height=3cm,width=4cm]{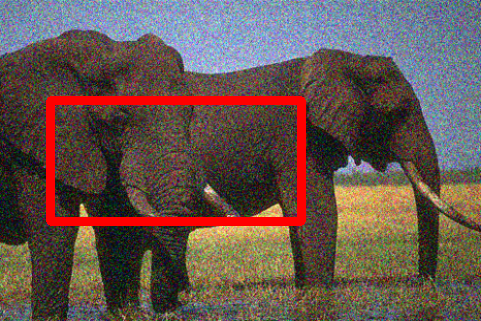} \\

            \includegraphics[height=2cm,width=4cm]{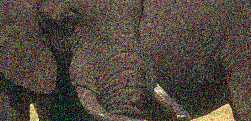} &
            \includegraphics[height=2cm,width=4cm]{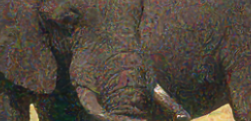} &
            \includegraphics[height=2cm,width=4cm]{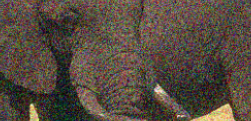} &
            \includegraphics[height=2cm,width=4cm]{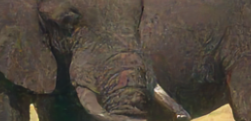} &
            \includegraphics[height=2cm,width=4cm]{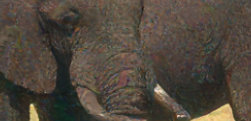} &
            \includegraphics[height=2cm,width=4cm]{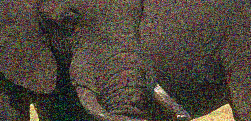} &
            \includegraphics[height=2cm,width=4cm]{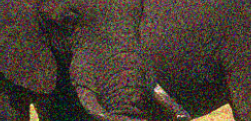} \\

            (a) Input & 
            (b) PromptIR \cite{PromptIR} &
            (c) GridFormer \cite{GridFormer} & 
            (d) RAM \cite{RAM} & 
            (e) NDR-Restore \cite{NDR-Restore} & 
            (f) DGSolver \cite{DGSolver} & 
            (g) AWRaCLe \cite{AWRaCLe} \\

            \includegraphics[height=3cm,width=4cm]{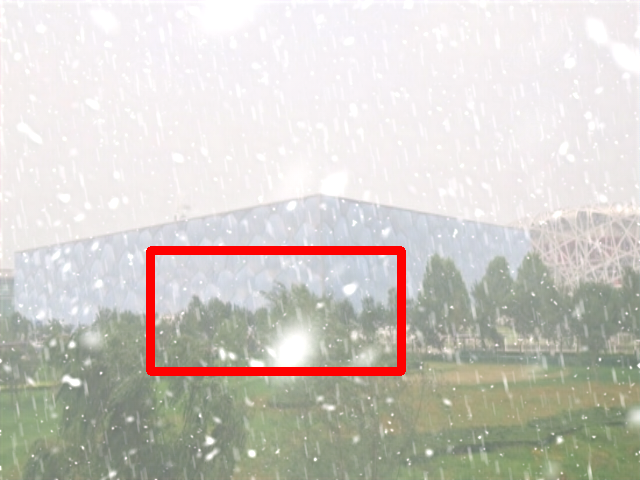} &
            \includegraphics[height=3cm,width=4cm]{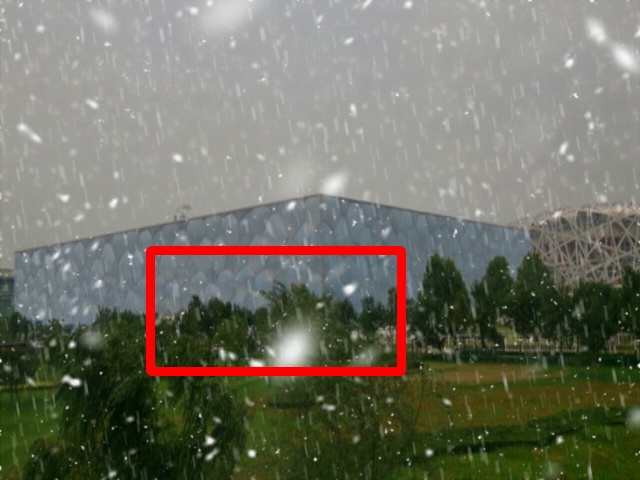} &
            \includegraphics[height=3cm,width=4cm]{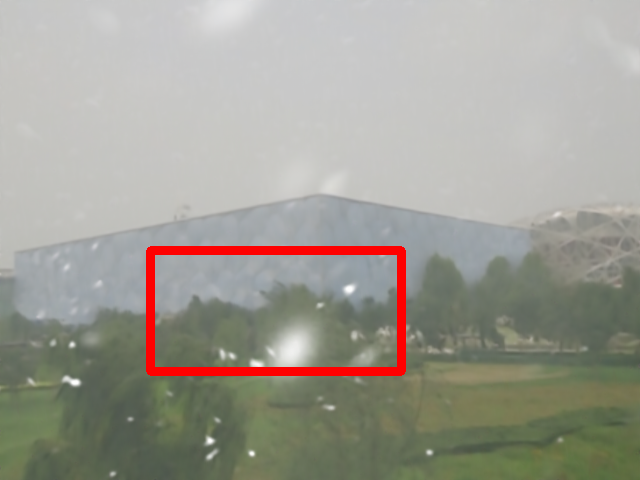} &
            \includegraphics[height=3cm,width=4cm]{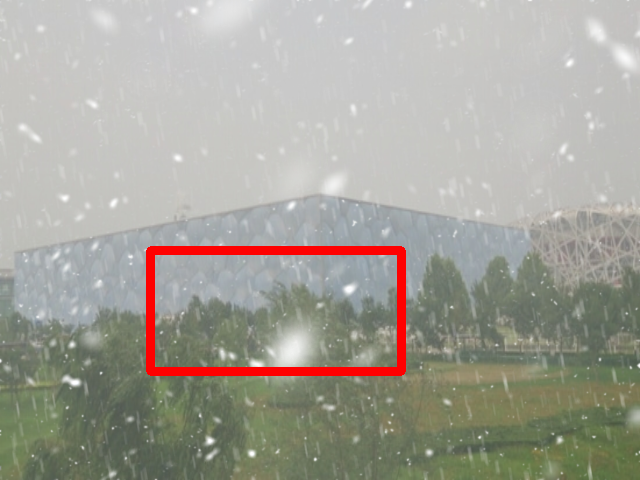} &
            \includegraphics[height=3cm,width=4cm]{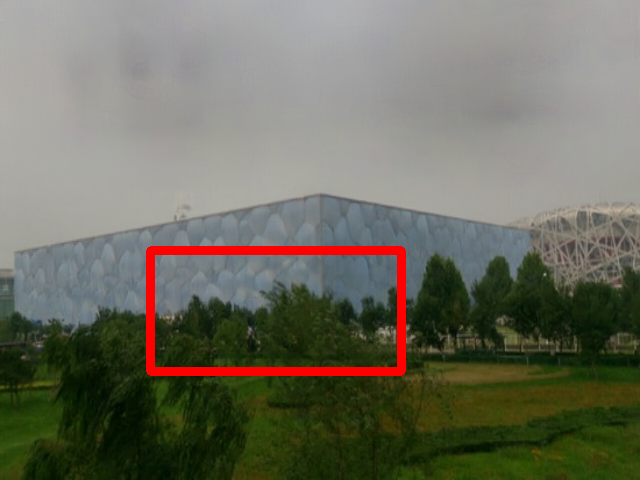} &
            \includegraphics[height=3cm,width=4cm]{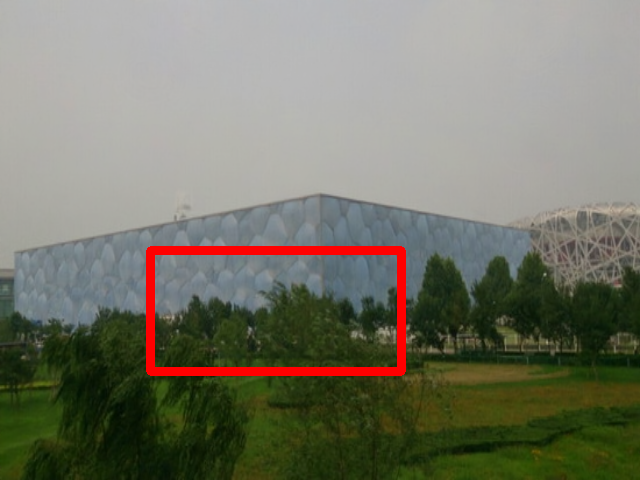} &
            \includegraphics[height=3cm,width=4cm]{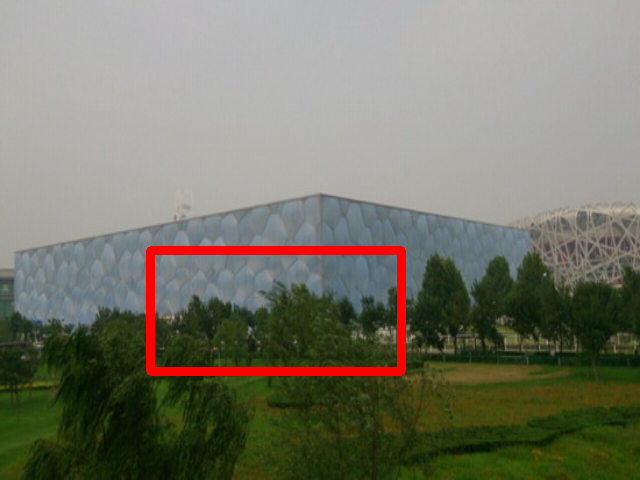} \\

            \includegraphics[height=2cm,width=4cm]{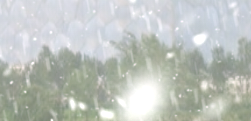} &
            \includegraphics[height=2cm,width=4cm]{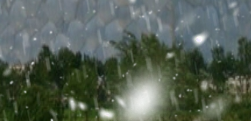} &
            \includegraphics[height=2cm,width=4cm]{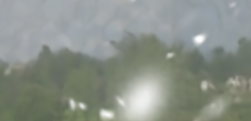} &
            \includegraphics[height=2cm,width=4cm]{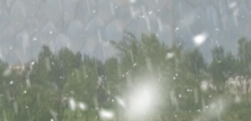} &
            \includegraphics[height=2cm,width=4cm]{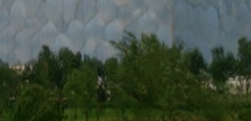} &
            \includegraphics[height=2cm,width=4cm]{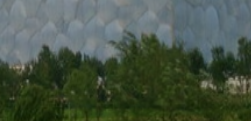} &
            \includegraphics[height=2cm,width=4cm]{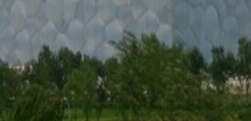} \\

            \includegraphics[height=3cm,width=4cm]{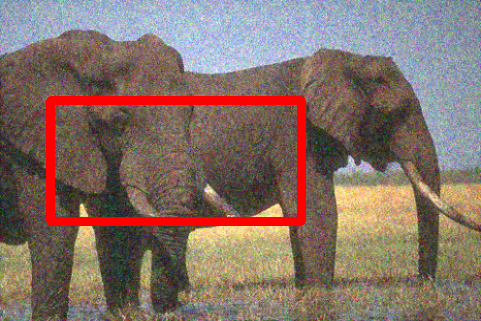} &
            \includegraphics[height=3cm,width=4cm]{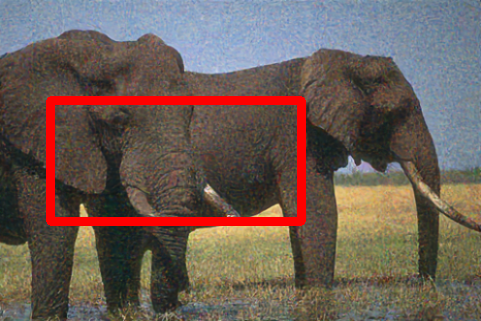} &
            \includegraphics[height=3cm,width=4cm]{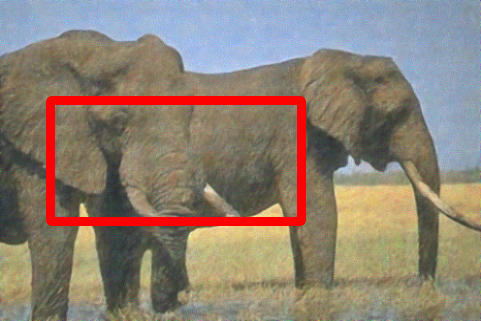} &
            \includegraphics[height=3cm,width=4cm]{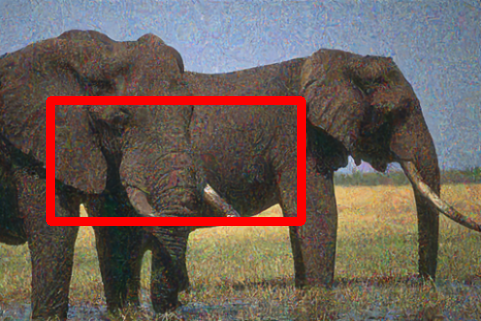} &
            \includegraphics[height=3cm,width=4cm]{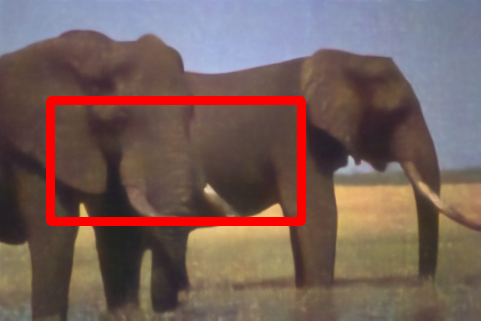} &
            \includegraphics[height=3cm,width=4cm]{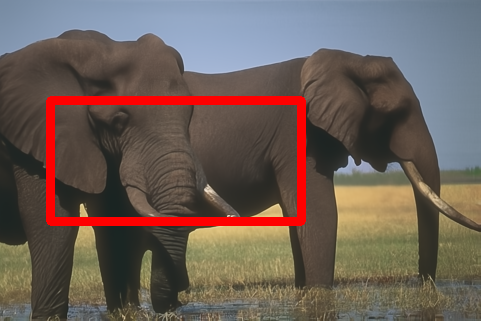} &
            \includegraphics[height=3cm,width=4cm]{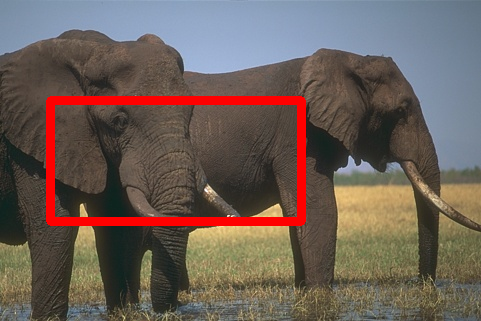} \\

            \includegraphics[height=2cm,width=4cm]{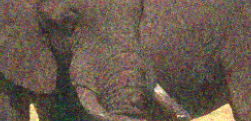} &
            \includegraphics[height=2cm,width=4cm]{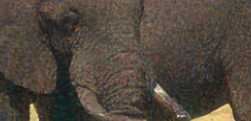} &
            \includegraphics[height=2cm,width=4cm]{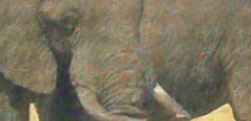} &
            \includegraphics[height=2cm,width=4cm]{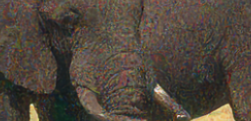} &
            \includegraphics[height=2cm,width=4cm]{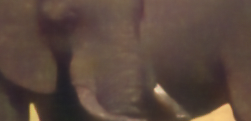} &
            \includegraphics[height=2cm,width=4cm]{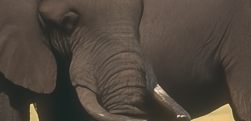} &
            \includegraphics[height=2cm,width=4cm]{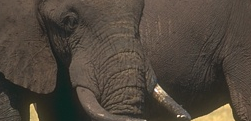} \\

            (h) DFPIR \cite{DFPIR} & 
            (i) DA-RCOT \cite{DA-RCOT} &
            (j) CPLIR \cite{CPLIR} & 
            (k) AdaIR \cite{AdaIR} & 
            (l) AgenticIR \cite{AgenticIR} & 
            (m) PaAgent & 
            (n) Reference \\ 
        \end{tabular}
    }
    \caption{Visual comparison of different methods on snow (CSD \cite{CSD}) and noise (BSD68 \cite{BSD68}) images. Our PaAgent yields better visual results on different degraded images, with outputs closer to the reference images.}
    \label{Qual_snow_noise}
\end{figure*}

\noindent
\textbf{Benchmark Datasets.} 
The training set consists of 15,013 images, which include 13,013 mixed degraded images from the CDD-11 training subset \cite{OneRestore}, along with 2,000 motion blurred images from the GoPro training subset \cite{GoPro}.
The CDD-11 \cite{OneRestore} dataset contains 11 categories of image degradations including single degradations (low-light, haze, rain, snow) and their combinations (low+haze, low+rain, low+snow, haze+rain, haze+snow, low+haze+rain, low+haze+snow).
The test set has 16,230 image pairs, including 
2,222 motion blurred image pairs from the GoPro testing subset \cite{GoPro},
1,000 haze image pairs from the SOTS \cite{RESIDE} dataset,
408 noise image pairs from the BSD68 \cite{BSD68} dataset,
4,300 rainy image pairs from the Rain13K \cite{MPRNet} dataset,
2,000 snow image pairs from the CSD \cite{CSD} dataset,
500 low-light image pairs from the LOL-V1 \cite{LOL-v1} dataset,
3,600 low-light and motion blurred image pairs from the LOL\_Blur \cite{LEDNet} dataset,
and 2,200 mixed degraded images from the CDD-11 testing subset \cite{OneRestore}.

\noindent
\textbf{Evaluation Metrics.} 
We adopt the peak signal-to-noise ratio (PSNR) and the structural similarity index measure (SSIM) \cite{PSNR_SSIM} to measure the performance of different methods.
Higher scores of PSNR and SSIM \cite{PSNR_SSIM} denote that the enhanced results are closer to the reference images in terms of image content and structure.
PSNR and SSIM \cite{PSNR_SSIM} metrics are produced on the Python platform.

\begin{figure*}[!ht]
    \Large
    \centering
    \resizebox{1\linewidth}{!}{
        \begin{tabular}{c@{ }c@{ }c@{ }c@{ }c@{ }c@{ }c@{ }}
            \includegraphics[height=3cm,width=4cm]{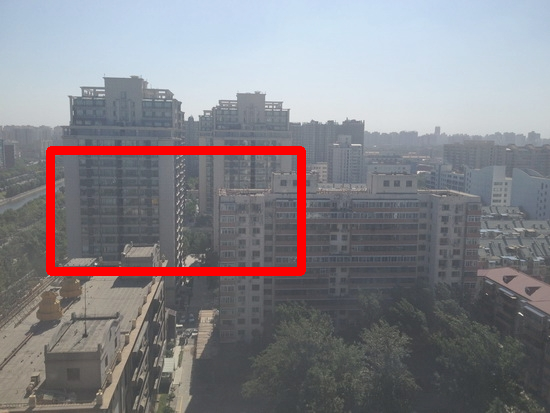} &
            \includegraphics[height=3cm,width=4cm]{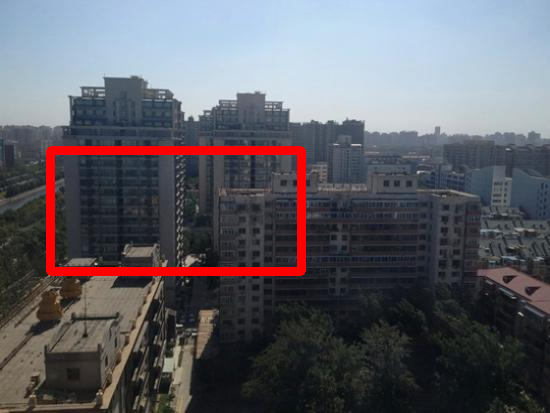} &
            \includegraphics[height=3cm,width=4cm]{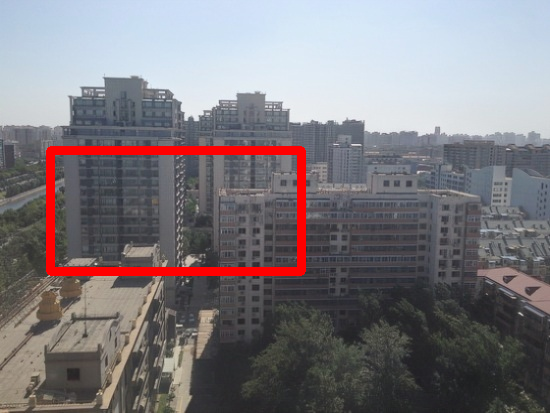} &
            \includegraphics[height=3cm,width=4cm]{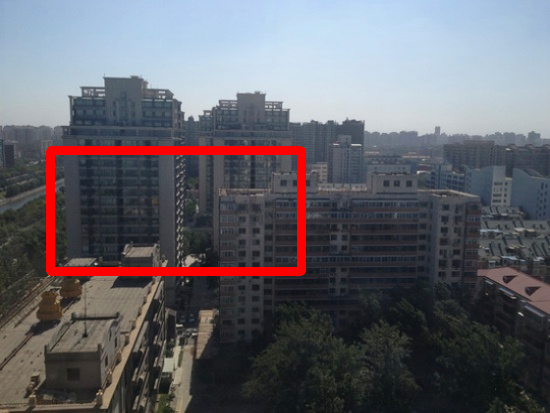} &
            \includegraphics[height=3cm,width=4cm]{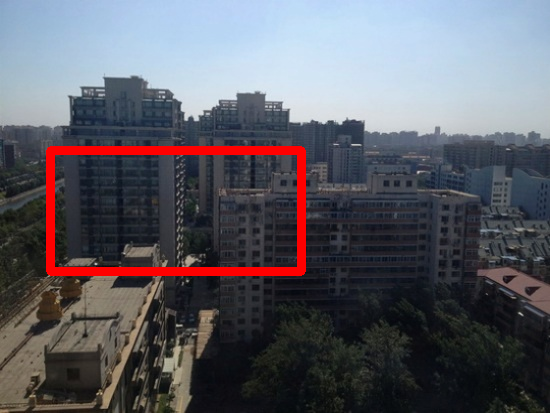} &
            \includegraphics[height=3cm,width=4cm]{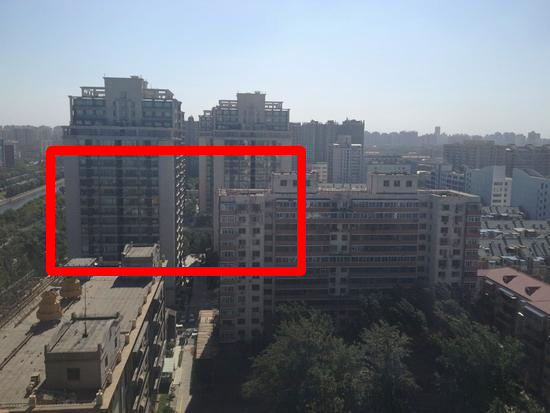} &
            \includegraphics[height=3cm,width=4cm]{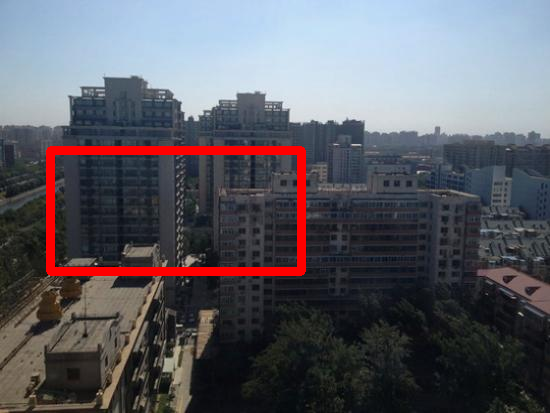} \\

            \includegraphics[height=2cm,width=4cm]{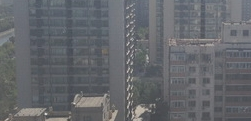} &
            \includegraphics[height=2cm,width=4cm]{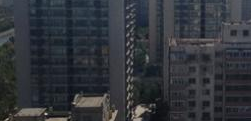} &
            \includegraphics[height=2cm,width=4cm]{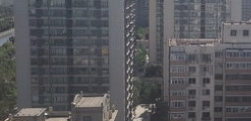} &
            \includegraphics[height=2cm,width=4cm]{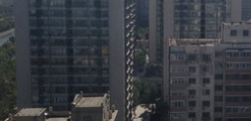} &
            \includegraphics[height=2cm,width=4cm]{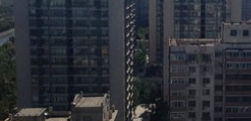} &
            \includegraphics[height=2cm,width=4cm]{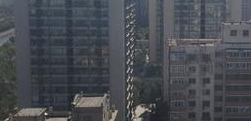} &
            \includegraphics[height=2cm,width=4cm]{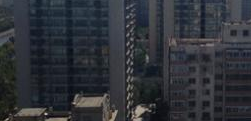} \\

            \includegraphics[height=3cm,width=4cm]{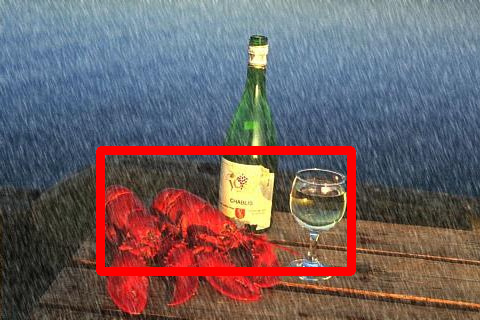} &
            \includegraphics[height=3cm,width=4cm]{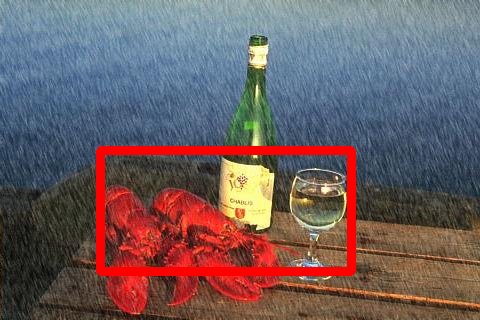} &
            \includegraphics[height=3cm,width=4cm]{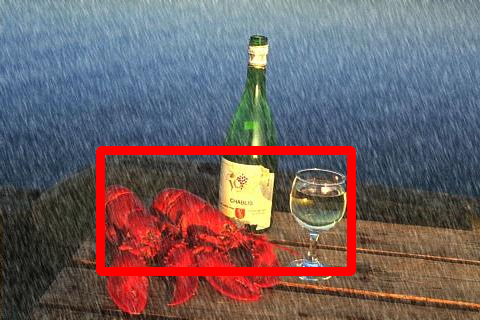} &
            \includegraphics[height=3cm,width=4cm]{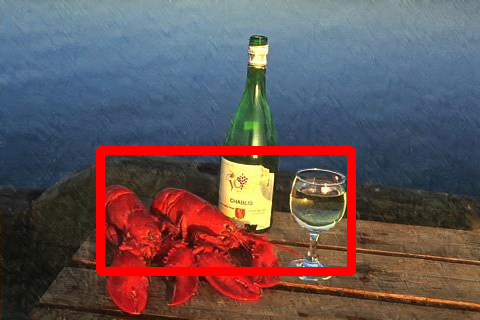} &
            \includegraphics[height=3cm,width=4cm]{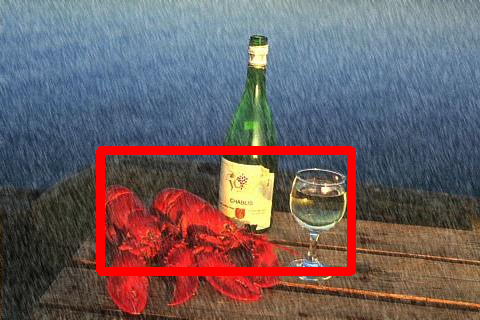} &
            \includegraphics[height=3cm,width=4cm]{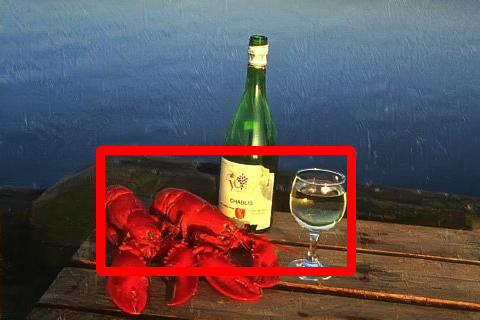} &
            \includegraphics[height=3cm,width=4cm]{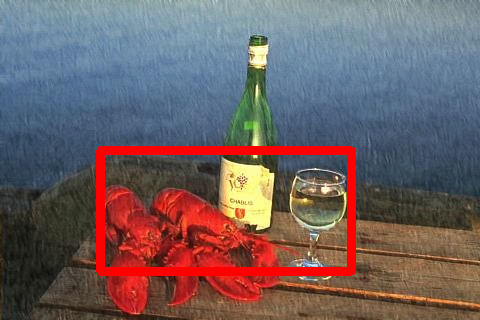} \\

            \includegraphics[height=2cm,width=4cm]{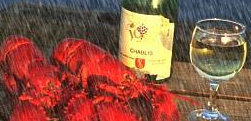} &
            \includegraphics[height=2cm,width=4cm]{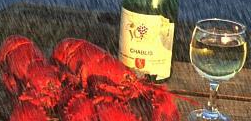} &
            \includegraphics[height=2cm,width=4cm]{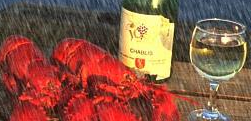} &
            \includegraphics[height=2cm,width=4cm]{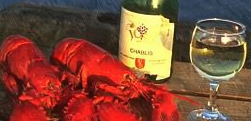} &
            \includegraphics[height=2cm,width=4cm]{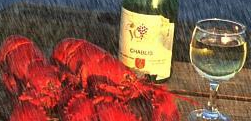} &
            \includegraphics[height=2cm,width=4cm]{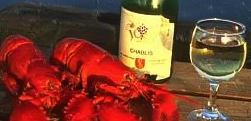} &
            \includegraphics[height=2cm,width=4cm]{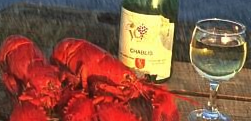} \\

            (a) Input & 
            (b) PromptIR \cite{PromptIR} &
            (c) GridFormer \cite{GridFormer} & 
            (d) RAM \cite{RAM} & 
            (e) NDR-Restore \cite{NDR-Restore} & 
            (f) DGSolver \cite{DGSolver} & 
            (g) AWRaCLe \cite{AWRaCLe} \\

            \includegraphics[height=3cm,width=4cm]{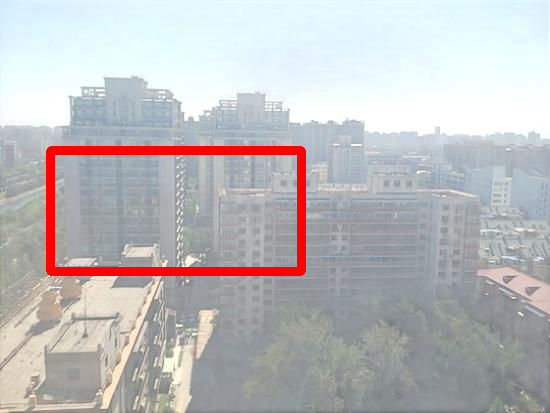} &
            \includegraphics[height=3cm,width=4cm]{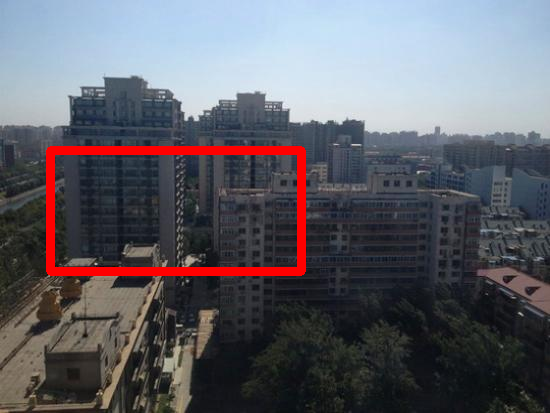} &
            \includegraphics[height=3cm,width=4cm]{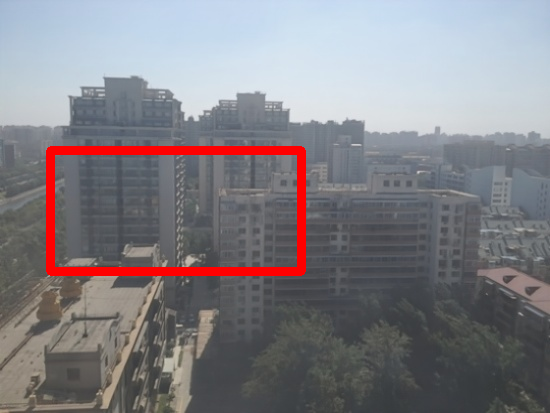} &
            \includegraphics[height=3cm,width=4cm]{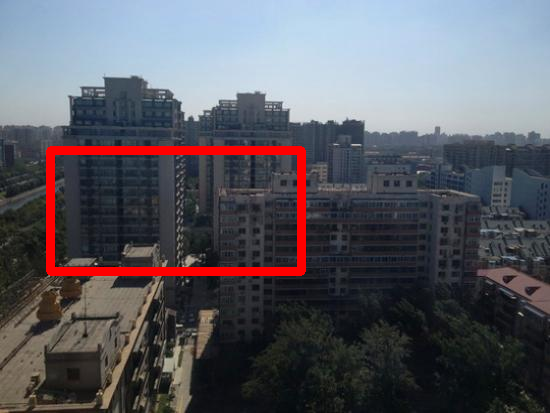} &
            \includegraphics[height=3cm,width=4cm]{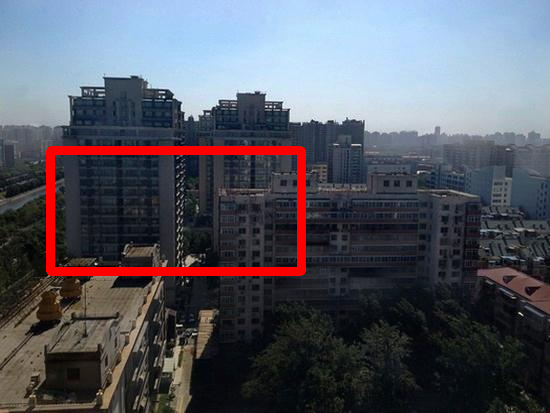} &
            \includegraphics[height=3cm,width=4cm]{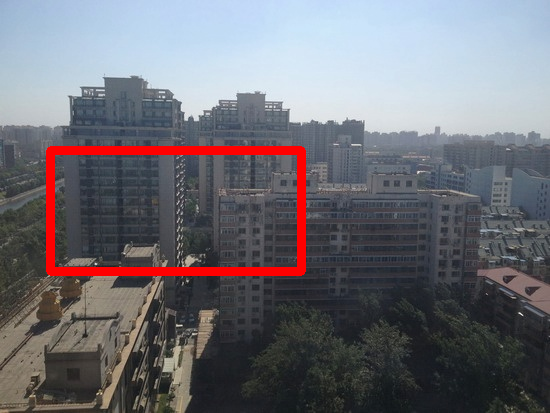} &
            \includegraphics[height=3cm,width=4cm]{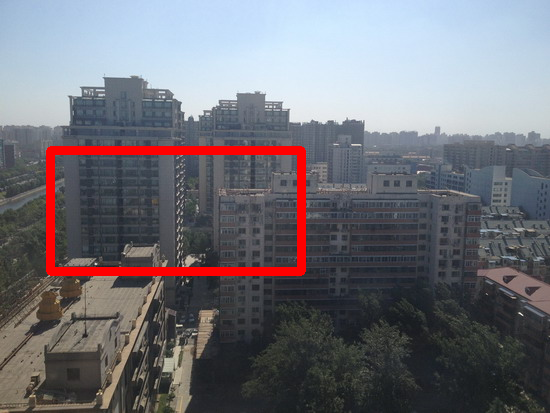} \\

            \includegraphics[height=2cm,width=4cm]{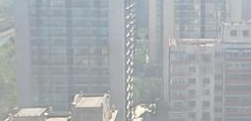} &
            \includegraphics[height=2cm,width=4cm]{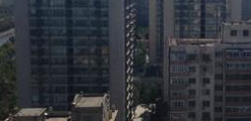} &
            \includegraphics[height=2cm,width=4cm]{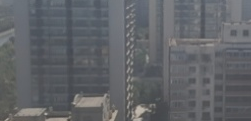} &
            \includegraphics[height=2cm,width=4cm]{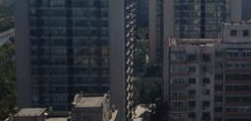} &
            \includegraphics[height=2cm,width=4cm]{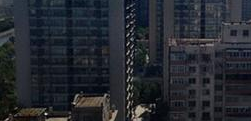} &
            \includegraphics[height=2cm,width=4cm]{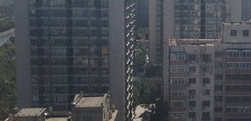} &
            \includegraphics[height=2cm,width=4cm]{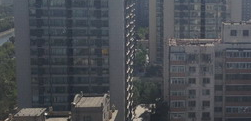} \\

            \includegraphics[height=3cm,width=4cm]{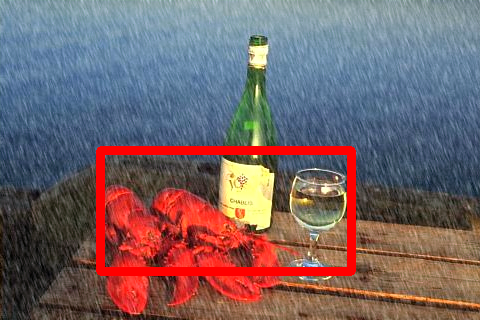} &
            \includegraphics[height=3cm,width=4cm]{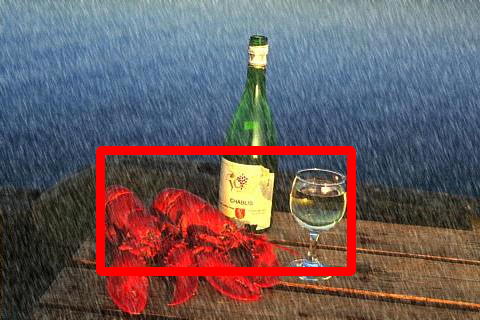} &
            \includegraphics[height=3cm,width=4cm]{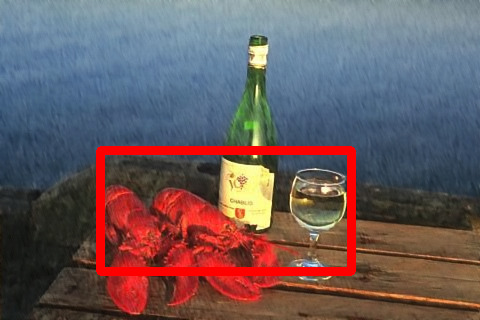} &
            \includegraphics[height=3cm,width=4cm]{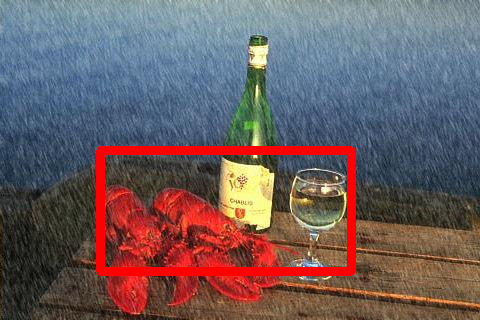} &
            \includegraphics[height=3cm,width=4cm]{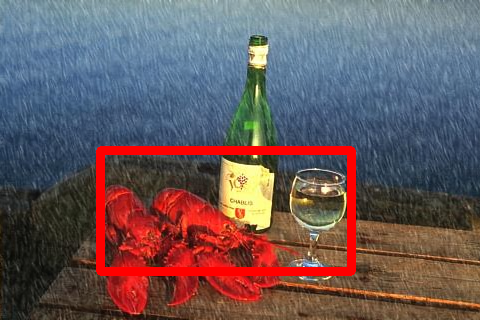} &
            \includegraphics[height=3cm,width=4cm]{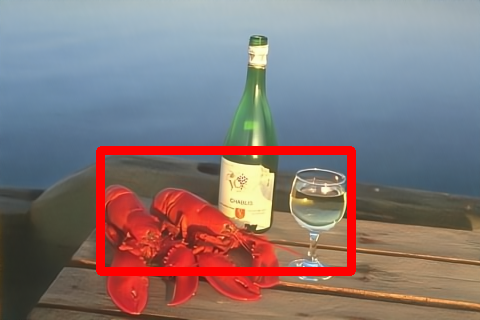} &
            \includegraphics[height=3cm,width=4cm]{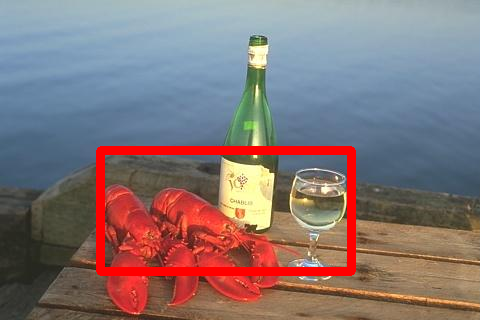} \\

            \includegraphics[height=2cm,width=4cm]{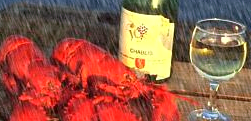} &
            \includegraphics[height=2cm,width=4cm]{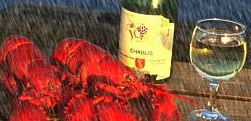} &
            \includegraphics[height=2cm,width=4cm]{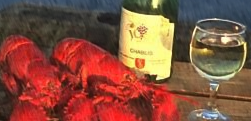} &
            \includegraphics[height=2cm,width=4cm]{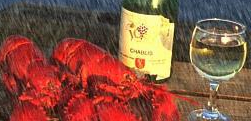} &
            \includegraphics[height=2cm,width=4cm]{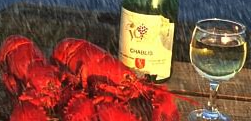} &
            \includegraphics[height=2cm,width=4cm]{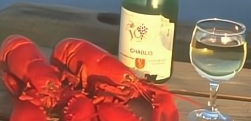} &
            \includegraphics[height=2cm,width=4cm]{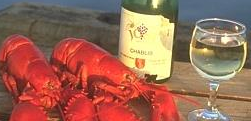} \\

            (h) DFPIR \cite{DFPIR} & 
            (i) DA-RCOT \cite{DA-RCOT} &
            (j) CPLIR \cite{CPLIR} & 
            (k) AdaIR \cite{AdaIR} & 
            (l) AgenticIR \cite{AgenticIR} & 
            (m) PaAgent & 
            (n) Reference \\ 
        \end{tabular}
    }
    \caption{Visual comparison of different methods on haze (SOTS \cite{RESIDE}) and rainy (Rain13K \cite{MPRNet}) images. Our PaAgent yields better visual results on different degraded images, with outputs closer to the reference images.}
    \label{Qual_haze_rain}
\end{figure*}

\begin{figure*}[!ht]
    \Large
    \centering
    \resizebox{1\linewidth}{!}{
        \begin{tabular}{c@{ }c@{ }c@{ }c@{ }c@{ }c@{ }c@{ }}
            \includegraphics[height=3cm,width=4cm]{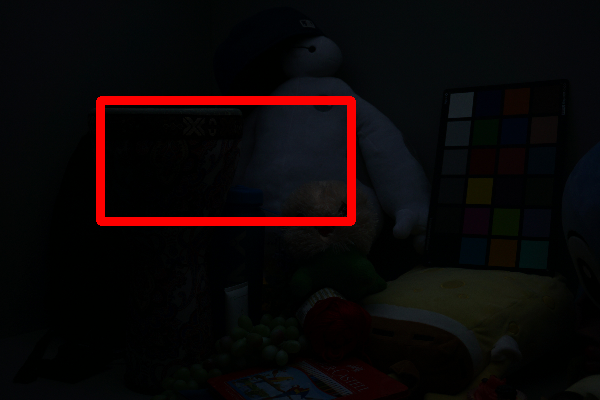} &
            \includegraphics[height=3cm,width=4cm]{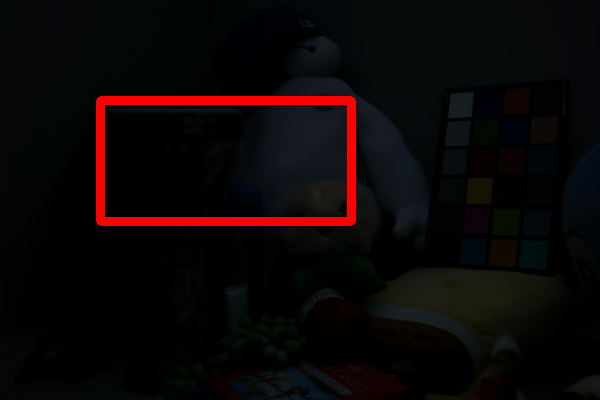} &
            \includegraphics[height=3cm,width=4cm]{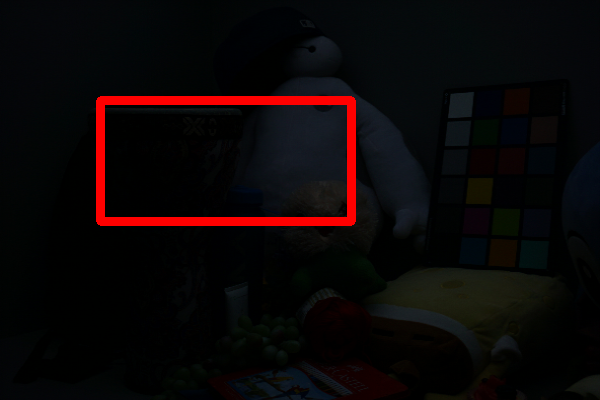} &
            \includegraphics[height=3cm,width=4cm]{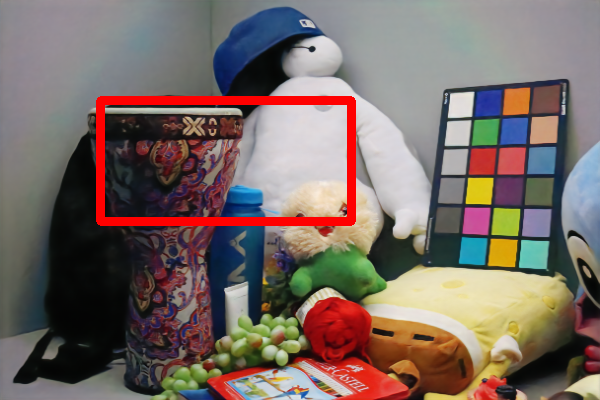} &
            \includegraphics[height=3cm,width=4cm]{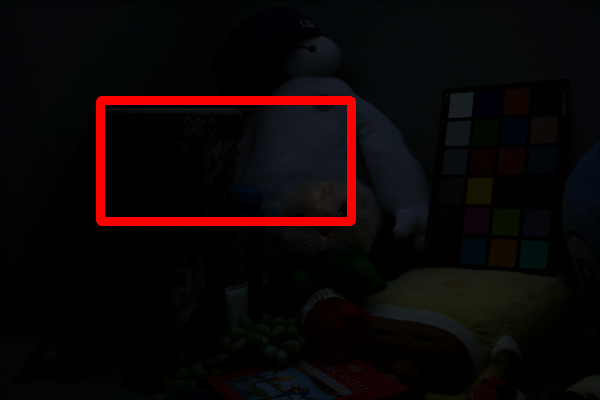} &
            \includegraphics[height=3cm,width=4cm]{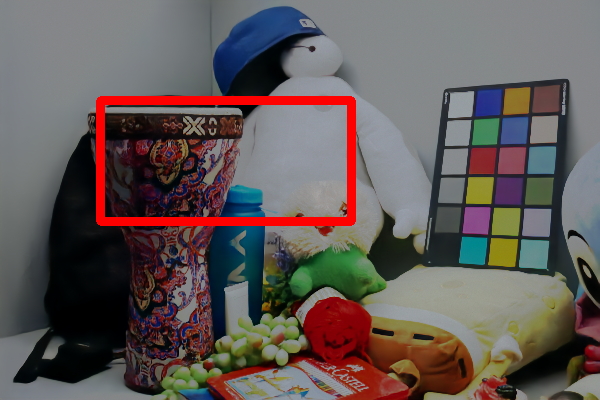} &
            \includegraphics[height=3cm,width=4cm]{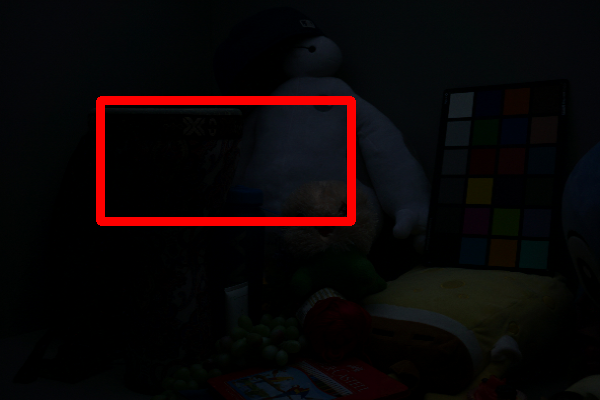} \\

            \includegraphics[height=2cm,width=4cm]{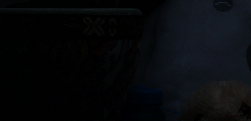} &
            \includegraphics[height=2cm,width=4cm]{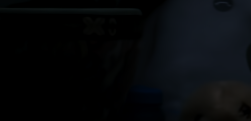} &
            \includegraphics[height=2cm,width=4cm]{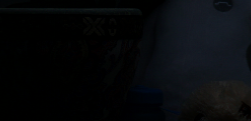} &
            \includegraphics[height=2cm,width=4cm]{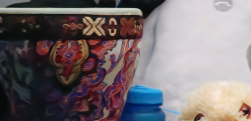} &
            \includegraphics[height=2cm,width=4cm]{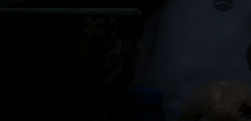} &
            \includegraphics[height=2cm,width=4cm]{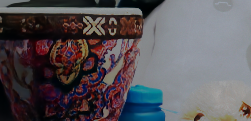} &
            \includegraphics[height=2cm,width=4cm]{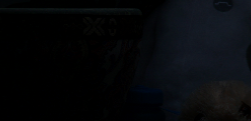} \\

            \includegraphics[height=3cm,width=4cm]{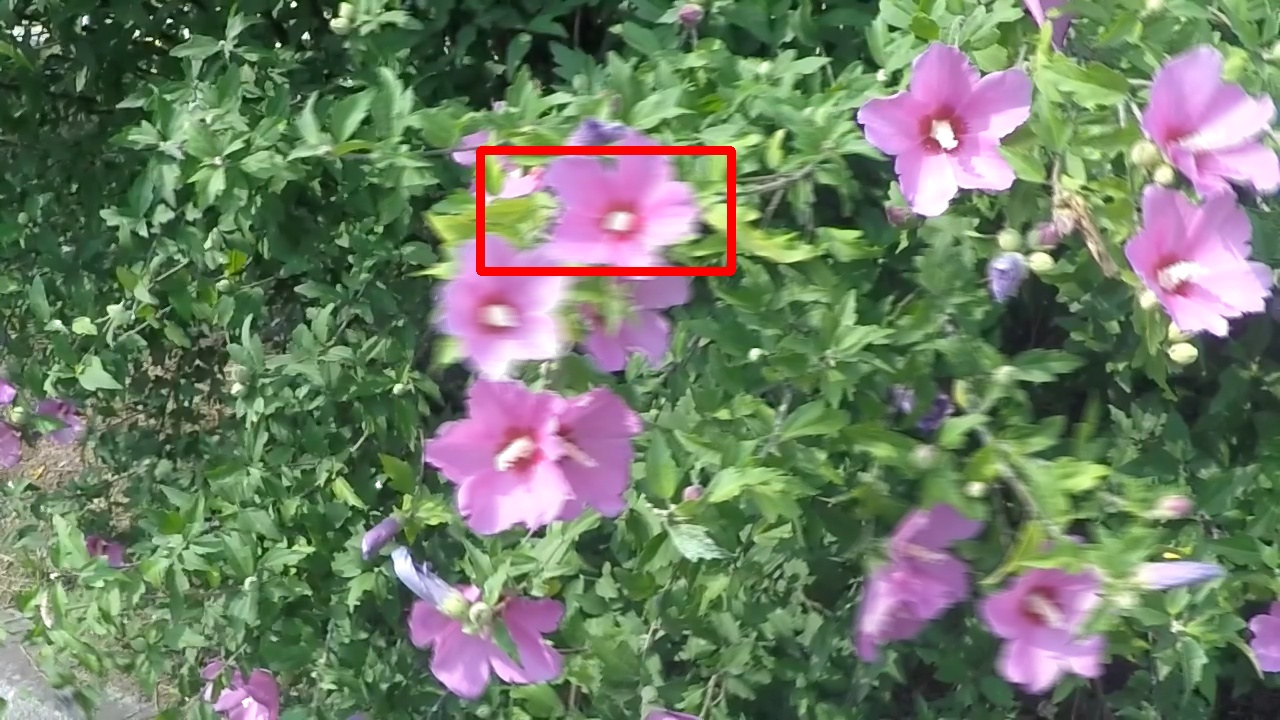} &
            \includegraphics[height=3cm,width=4cm]{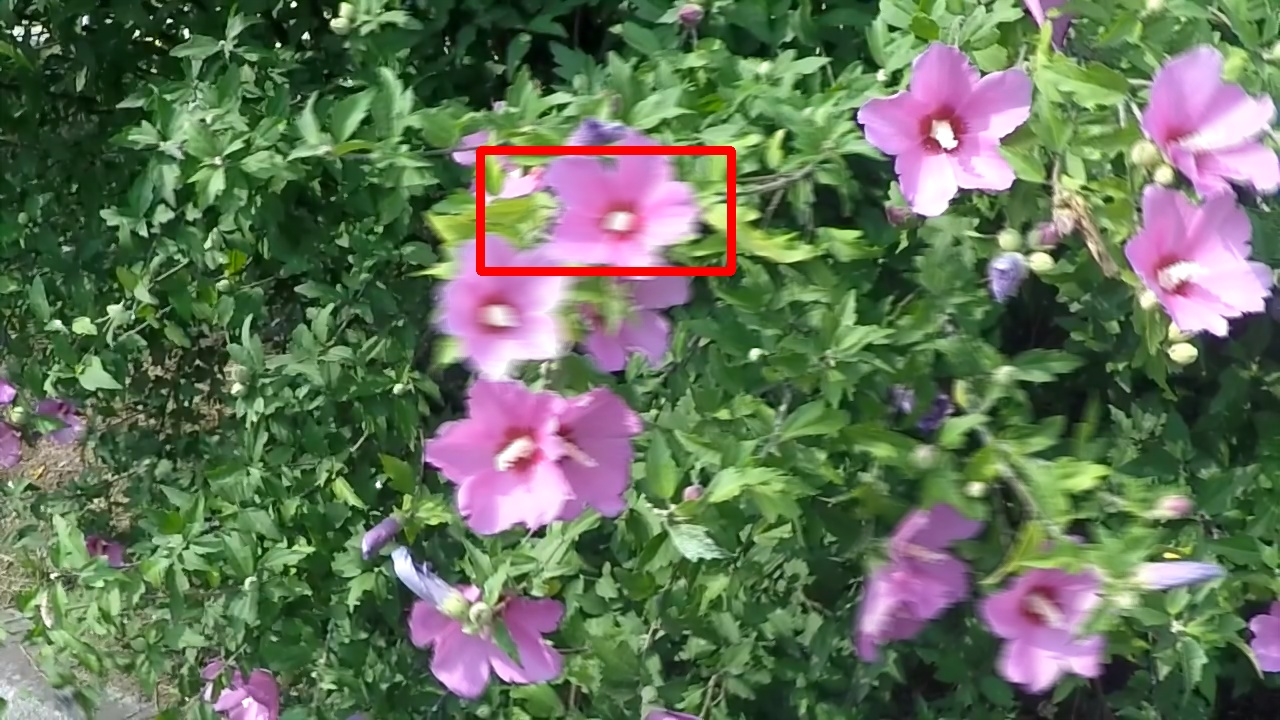} &
            \includegraphics[height=3cm,width=4cm]{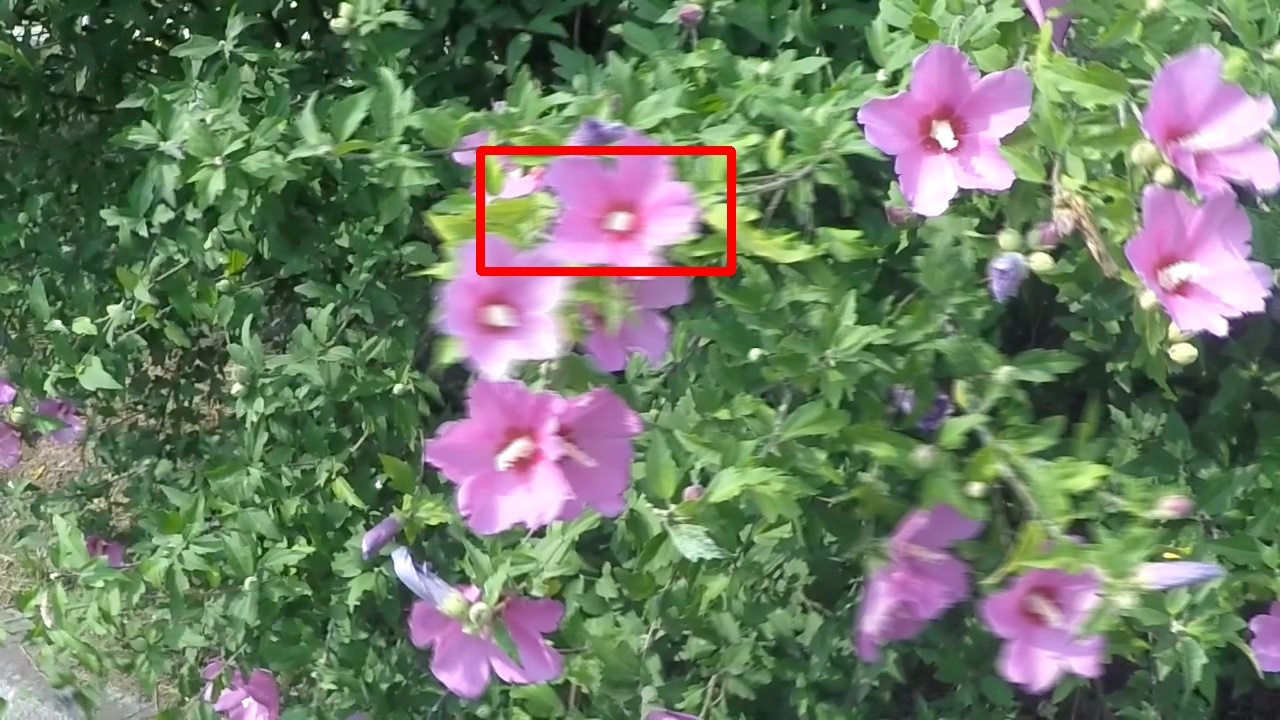} &
            \includegraphics[height=3cm,width=4cm]{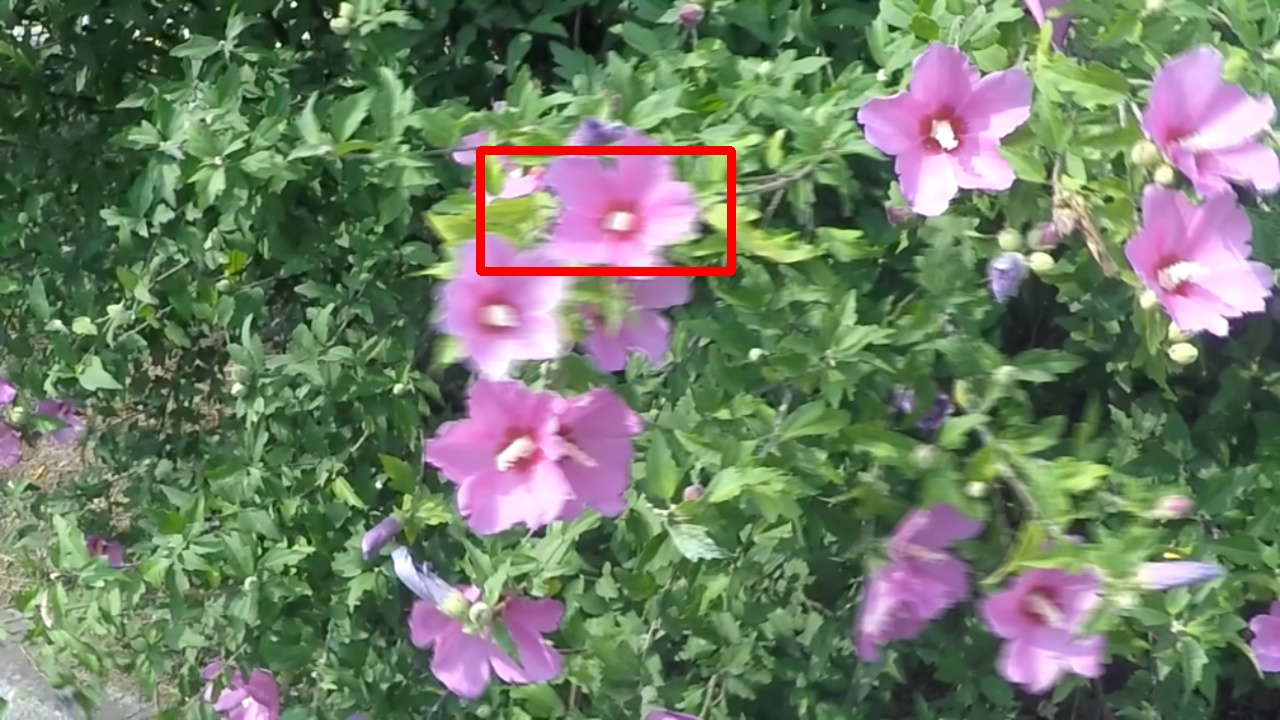} &
            \includegraphics[height=3cm,width=4cm]{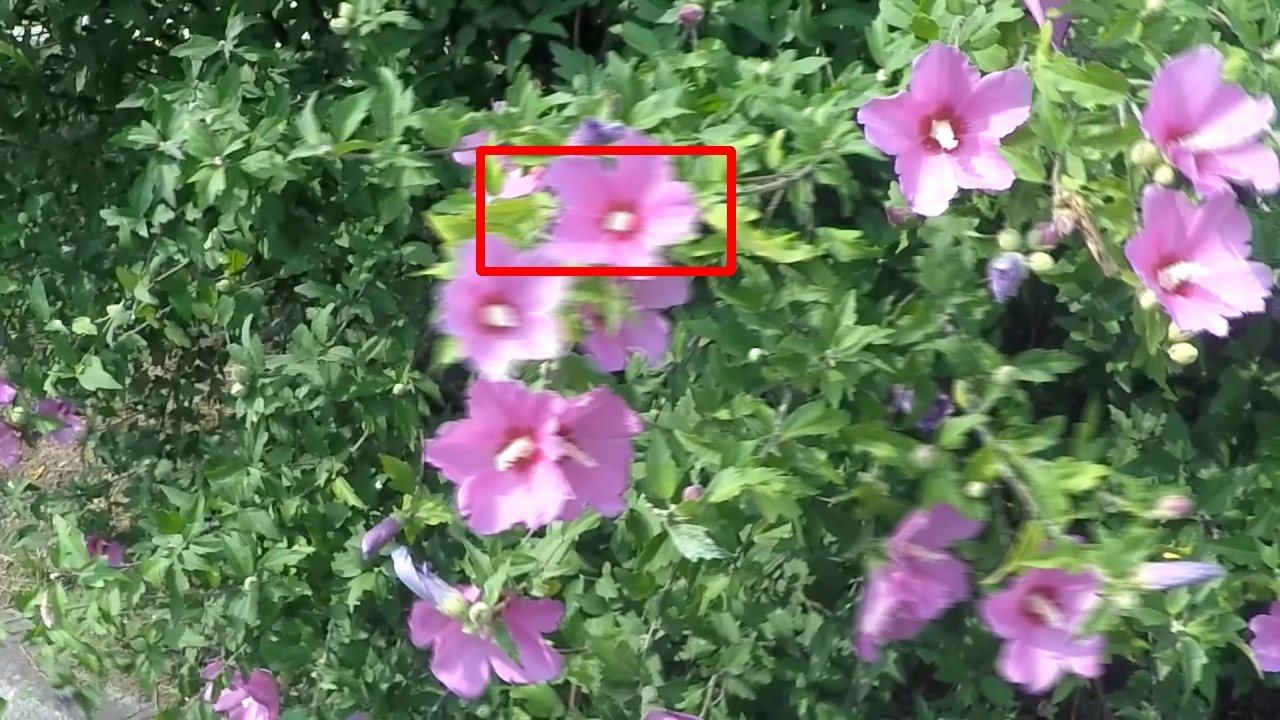} &
            \includegraphics[height=3cm,width=4cm]{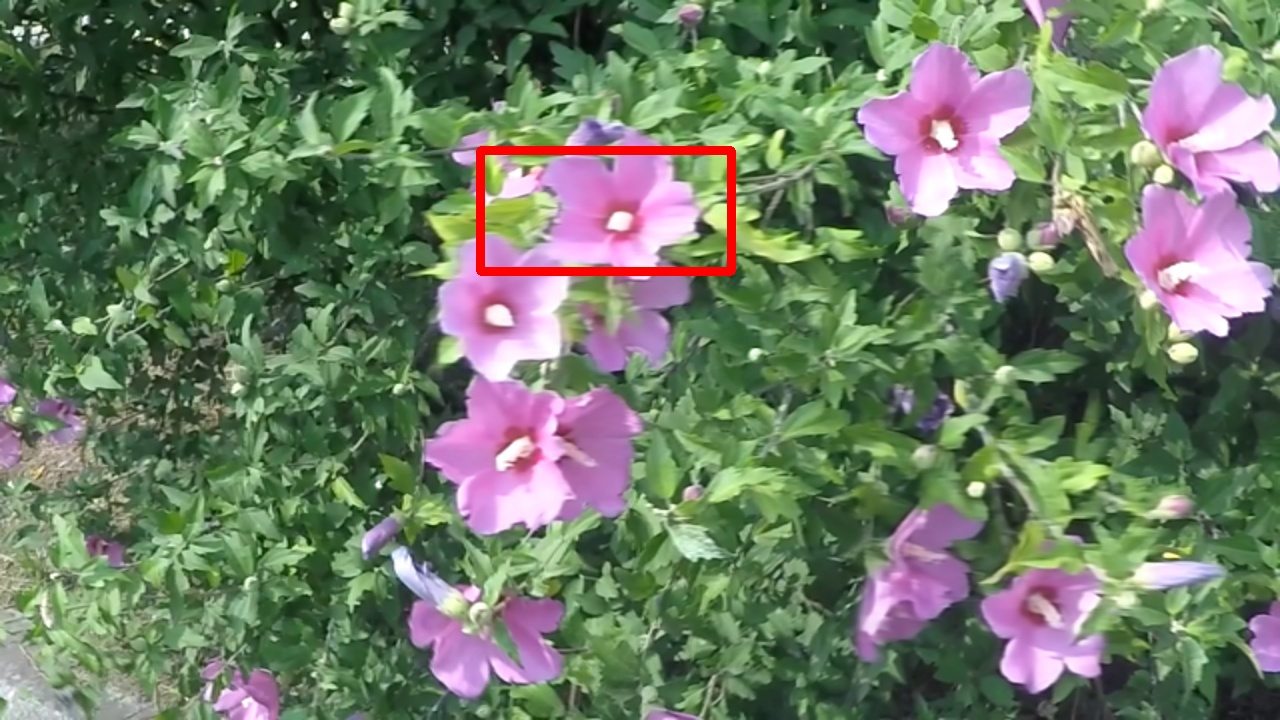} &
            \includegraphics[height=3cm,width=4cm]{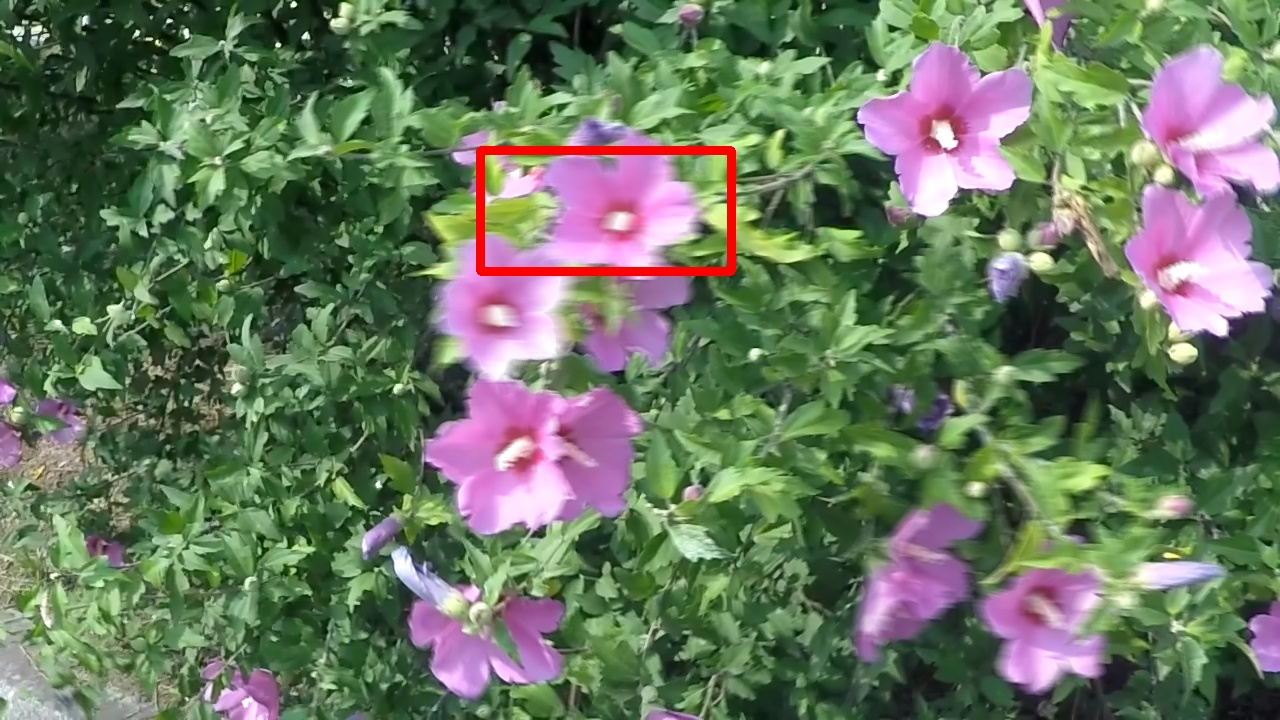} \\

            \includegraphics[height=2cm,width=4cm]{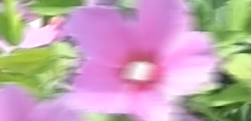} &
            \includegraphics[height=2cm,width=4cm]{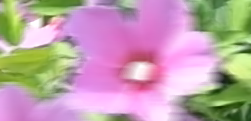} &
            \includegraphics[height=2cm,width=4cm]{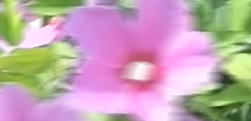} &
            \includegraphics[height=2cm,width=4cm]{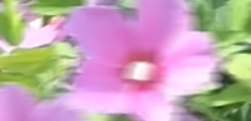} &
            \includegraphics[height=2cm,width=4cm]{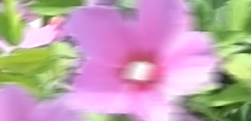} &
            \includegraphics[height=2cm,width=4cm]{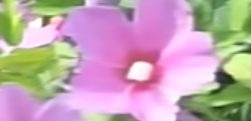} &
            \includegraphics[height=2cm,width=4cm]{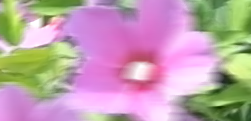} \\

            (a) Input & 
            (b) PromptIR \cite{PromptIR} &
            (c) GridFormer \cite{GridFormer} & 
            (d) RAM \cite{RAM} & 
            (e) NDR-Restore \cite{NDR-Restore} & 
            (f) DGSolver \cite{DGSolver} & 
            (g) AWRaCLe \cite{AWRaCLe} \\

            \includegraphics[height=3cm,width=4cm]{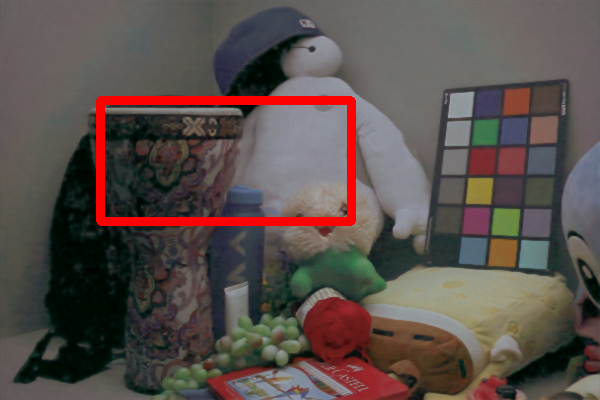} &
            \includegraphics[height=3cm,width=4cm]{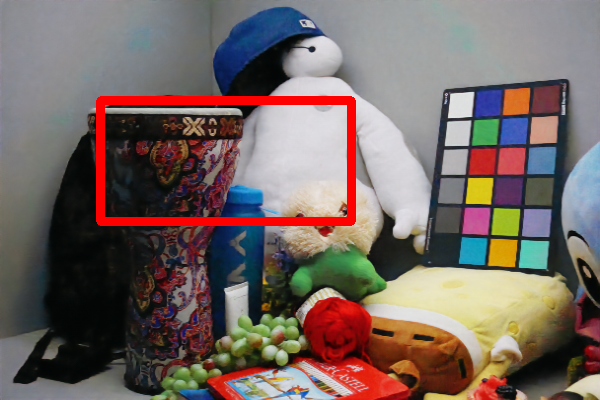} &
            \includegraphics[height=3cm,width=4cm]{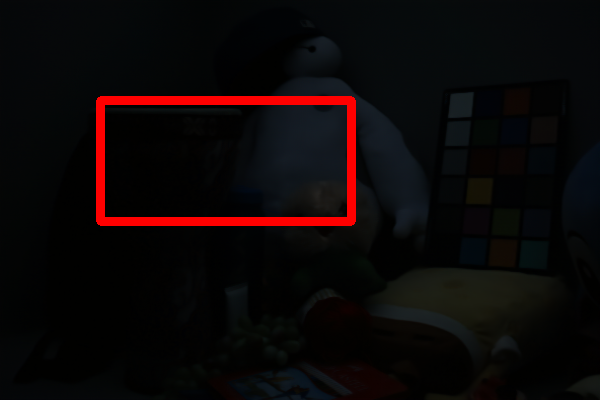} &
            \includegraphics[height=3cm,width=4cm]{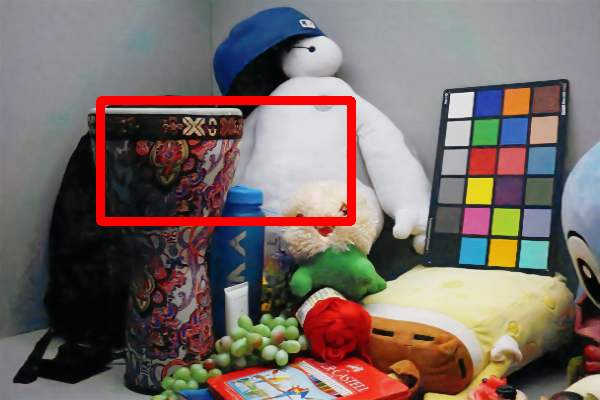} &
            \includegraphics[height=3cm,width=4cm]{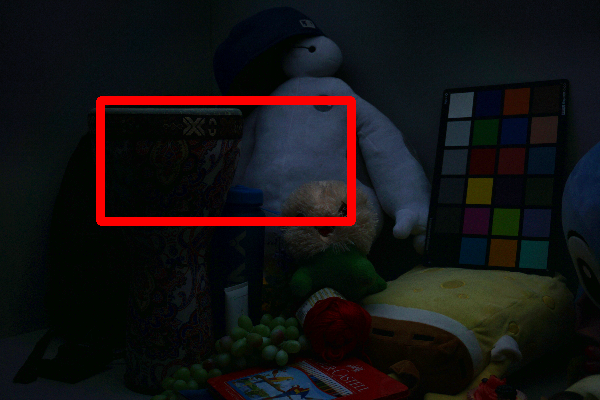} &
            \includegraphics[height=3cm,width=4cm]{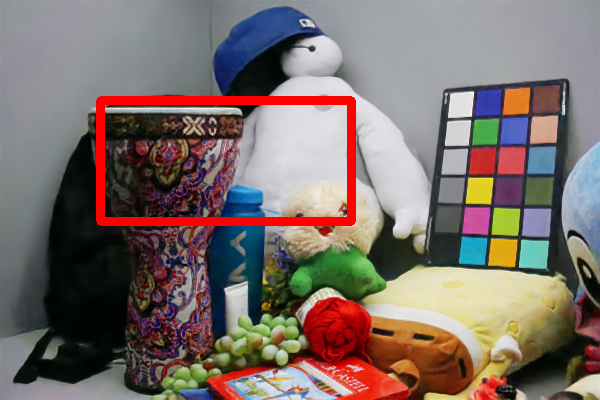} &
            \includegraphics[height=3cm,width=4cm]{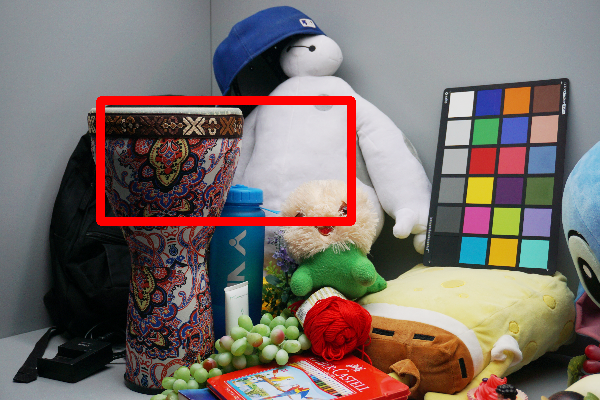} \\

            \includegraphics[height=2cm,width=4cm]{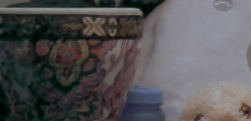} &
            \includegraphics[height=2cm,width=4cm]{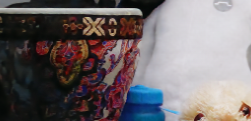} &
            \includegraphics[height=2cm,width=4cm]{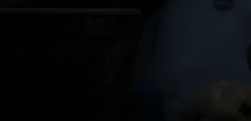} &
            \includegraphics[height=2cm,width=4cm]{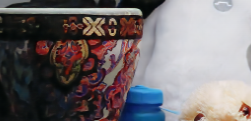} &
            \includegraphics[height=2cm,width=4cm]{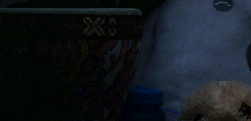} &
            \includegraphics[height=2cm,width=4cm]{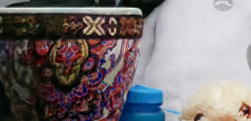} &
            \includegraphics[height=2cm,width=4cm]{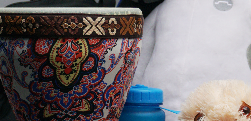} \\

            \includegraphics[height=3cm,width=4cm]{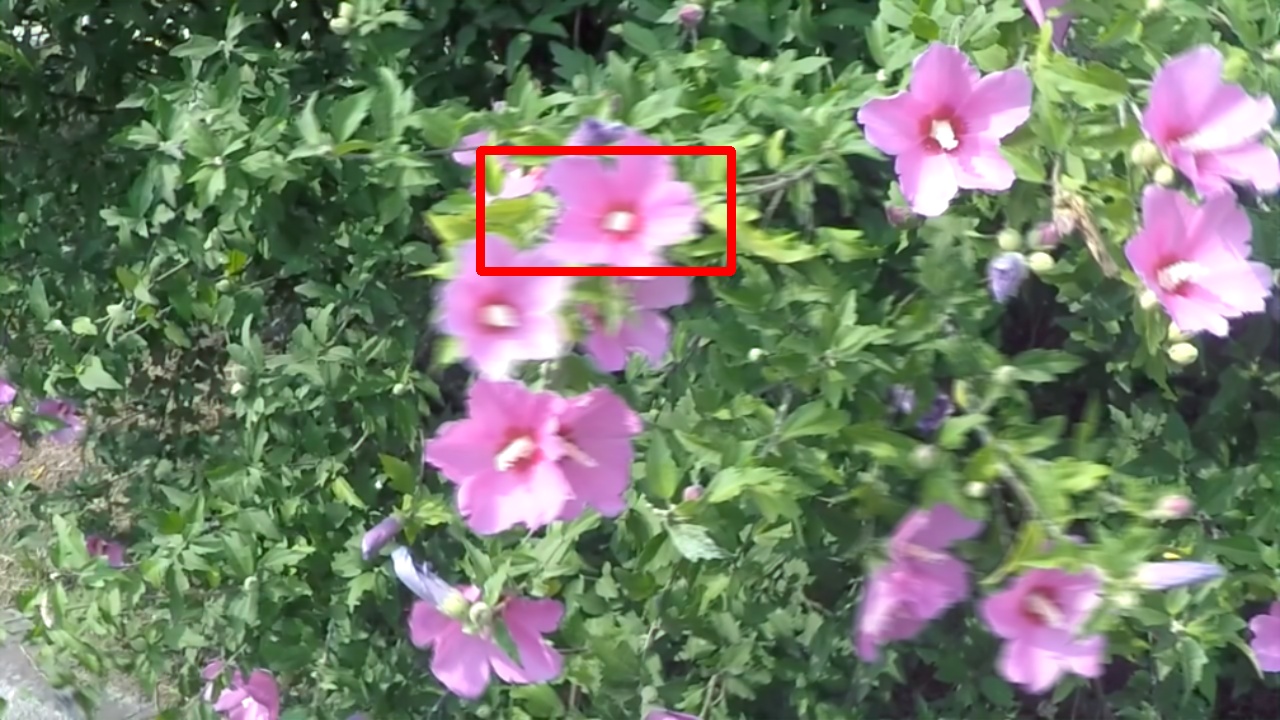} &
            \includegraphics[height=3cm,width=4cm]{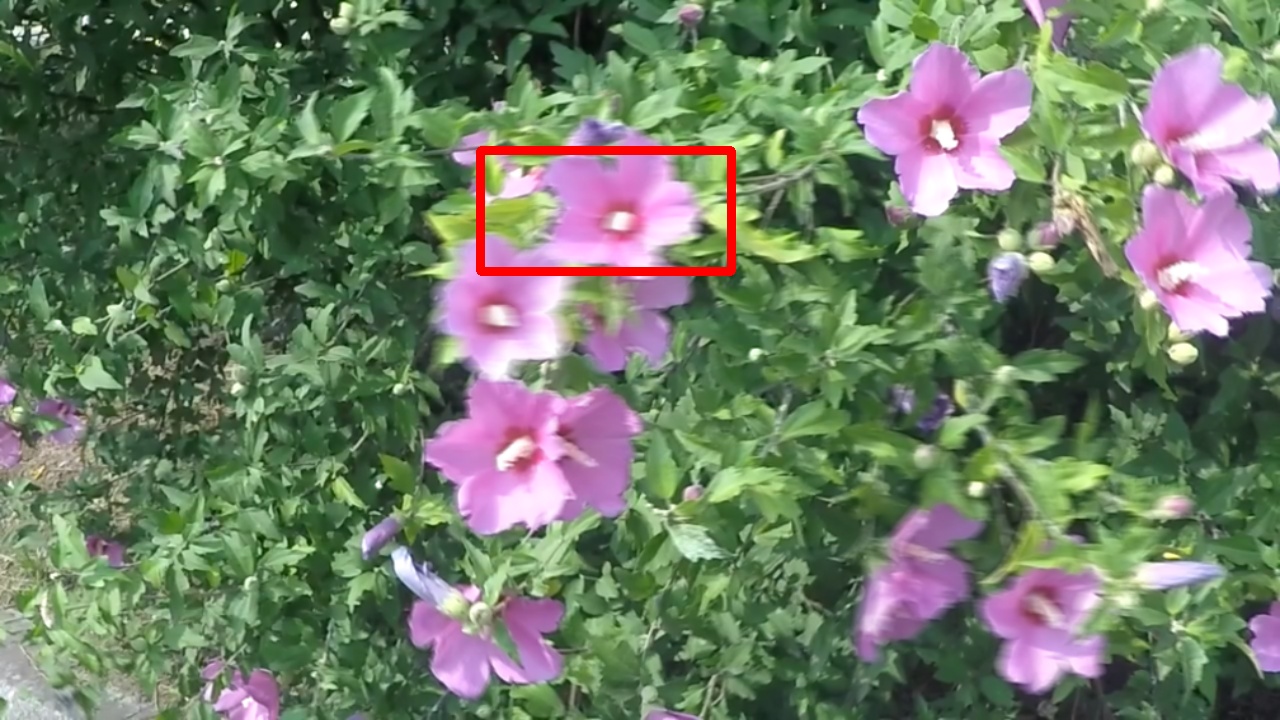} &
            \includegraphics[height=3cm,width=4cm]{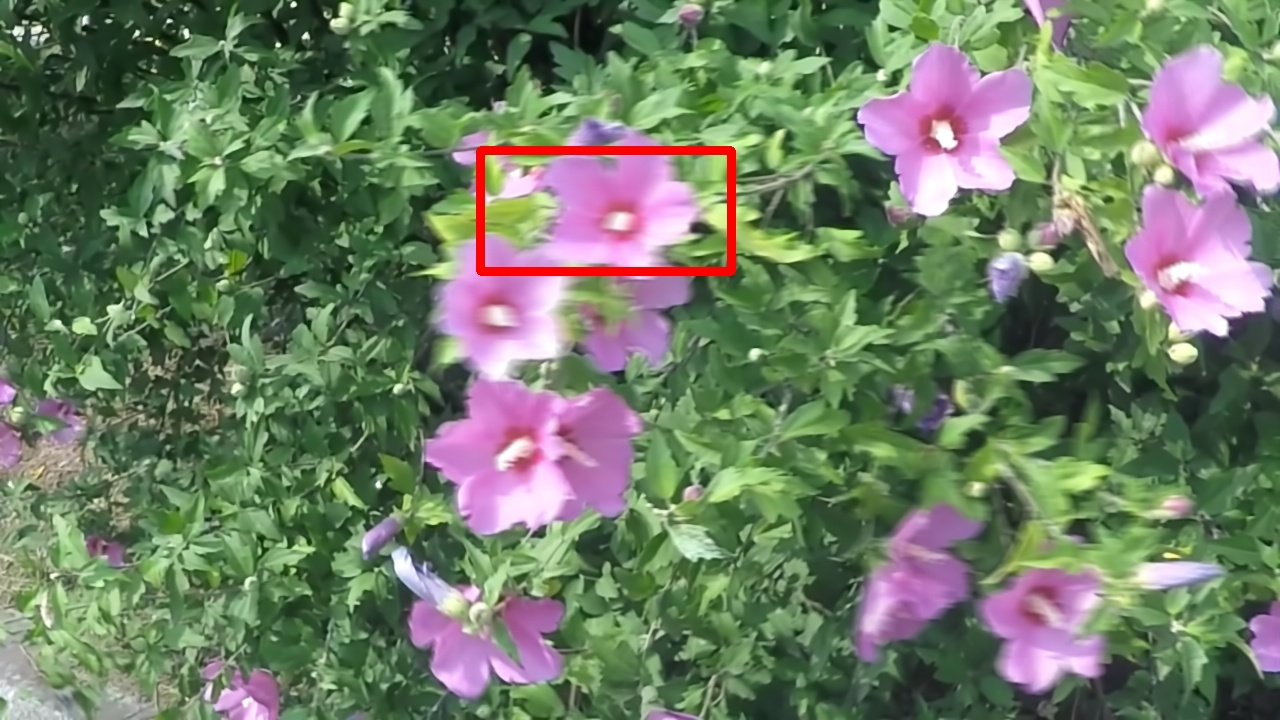} &
            \includegraphics[height=3cm,width=4cm]{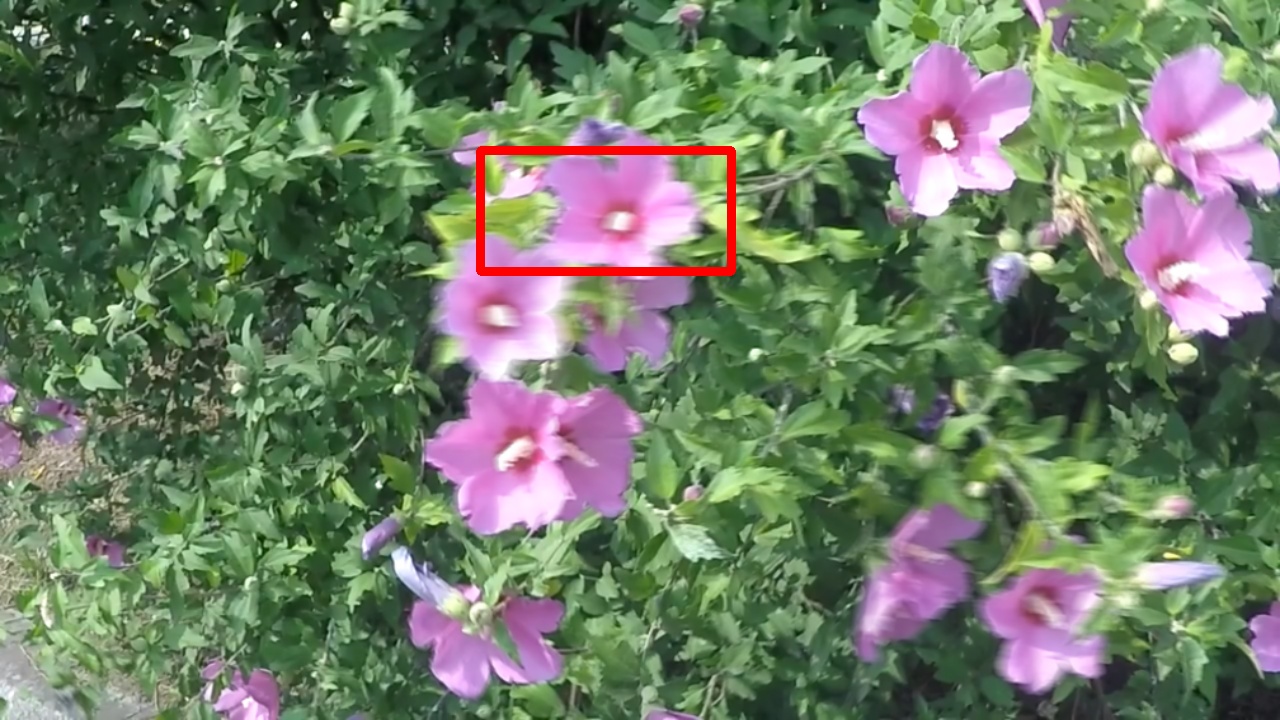} &
            \includegraphics[height=3cm,width=4cm]{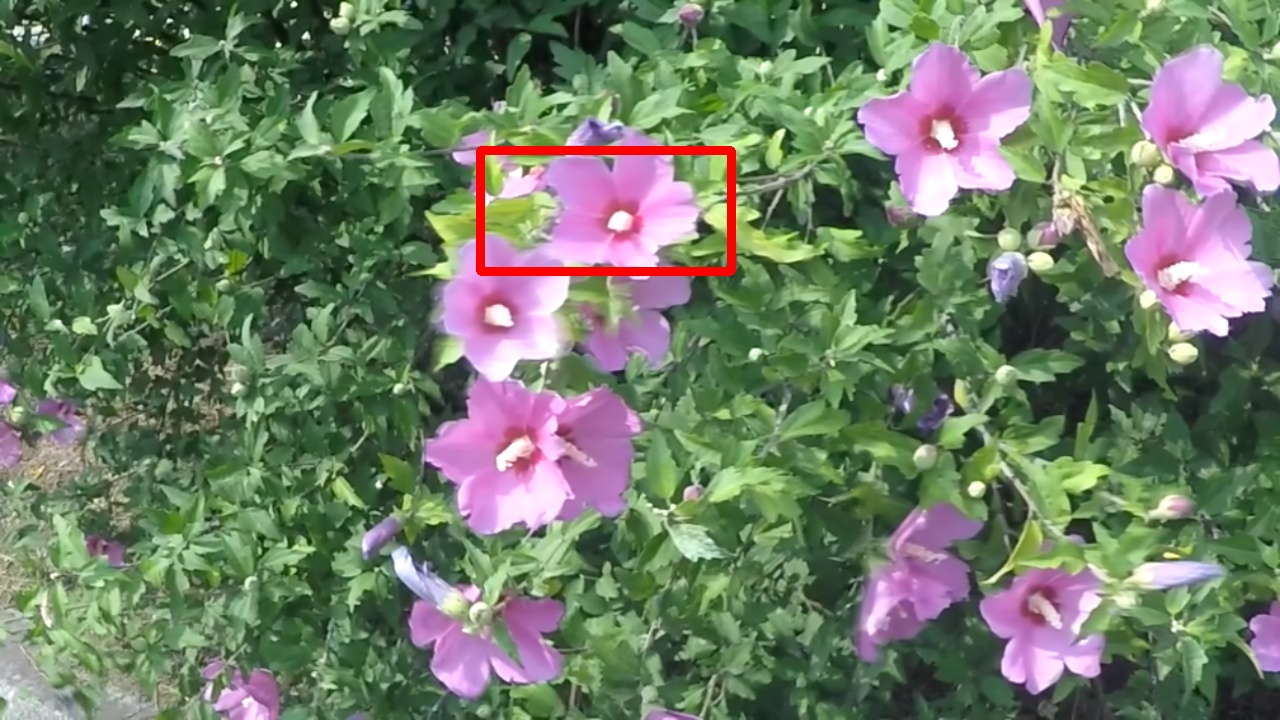} &
            \includegraphics[height=3cm,width=4cm]{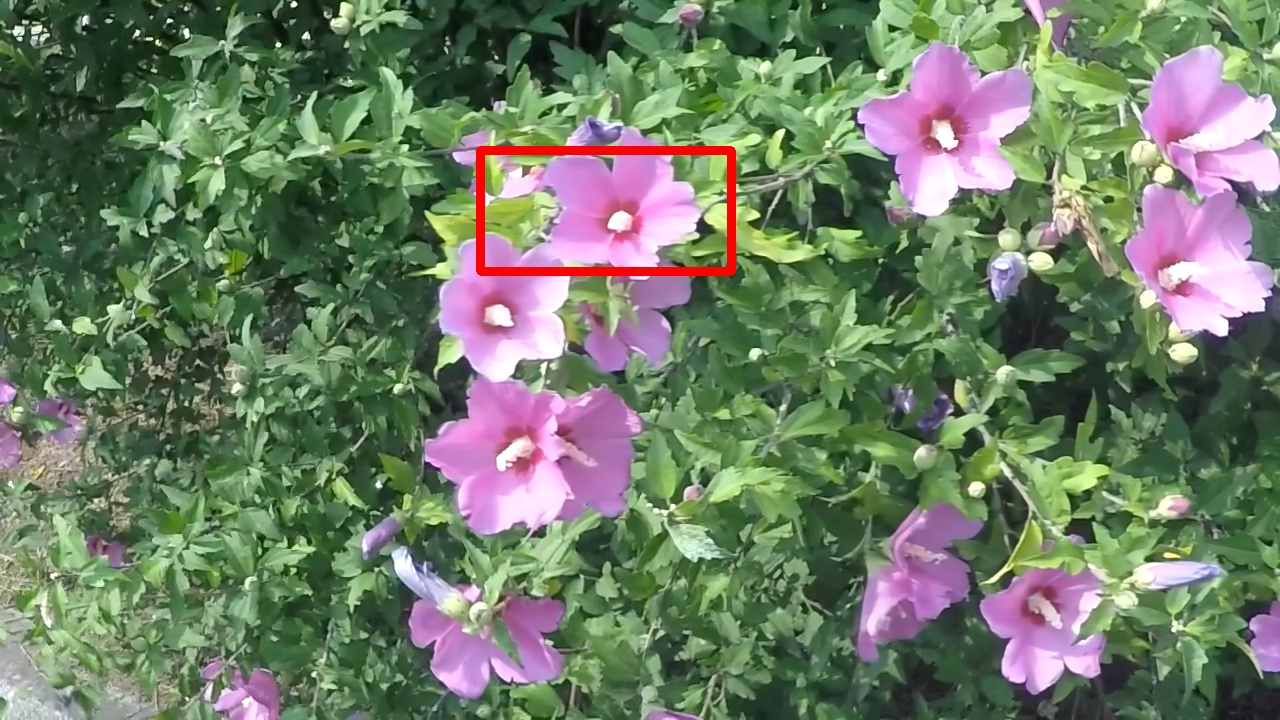} &
            \includegraphics[height=3cm,width=4cm]{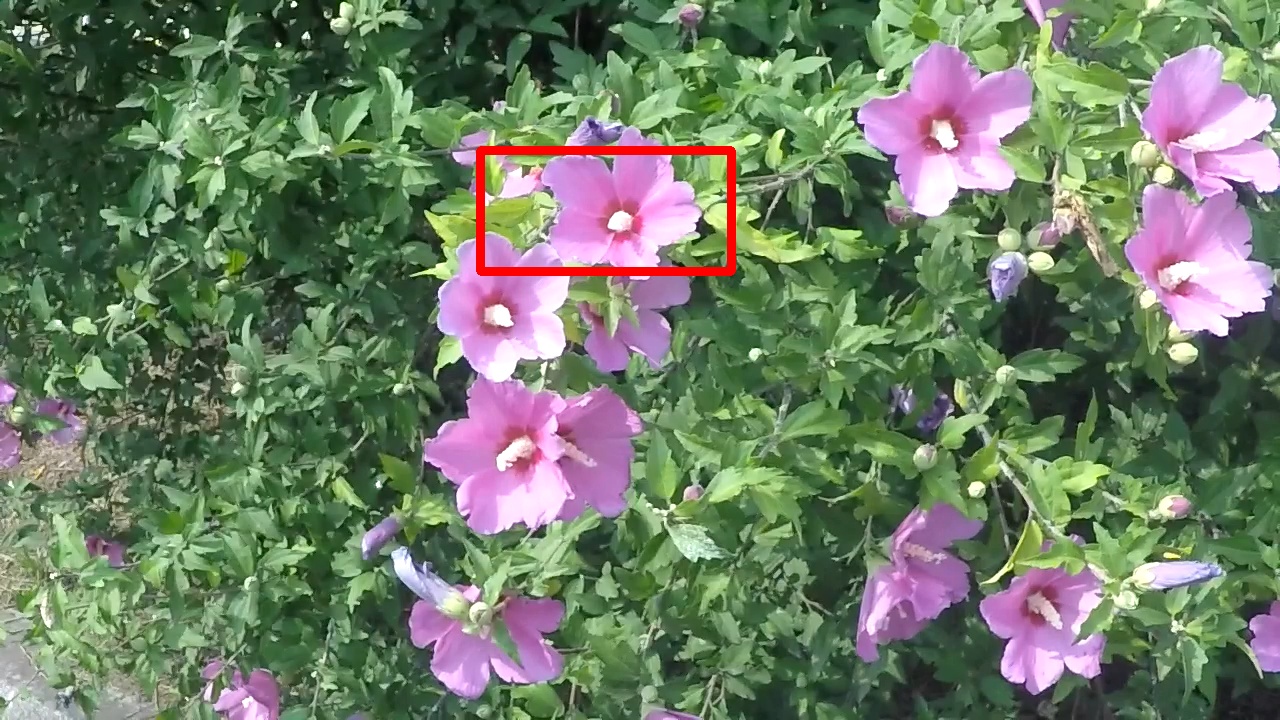} \\

            \includegraphics[height=2cm,width=4cm]{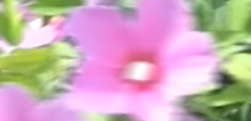} &
            \includegraphics[height=2cm,width=4cm]{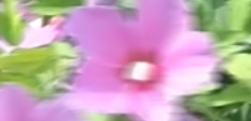} &
            \includegraphics[height=2cm,width=4cm]{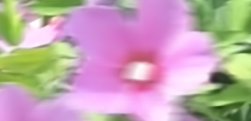} &
            \includegraphics[height=2cm,width=4cm]{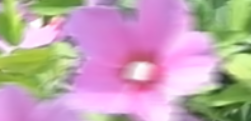} &
            \includegraphics[height=2cm,width=4cm]{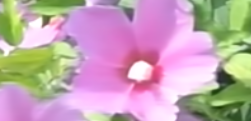} &
            \includegraphics[height=2cm,width=4cm]{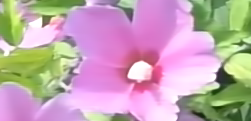} &
            \includegraphics[height=2cm,width=4cm]{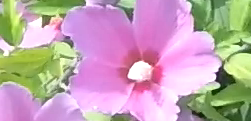} \\

            (h) DFPIR \cite{DFPIR} & 
            (i) DA-RCOT \cite{DA-RCOT} &
            (j) CPLIR \cite{CPLIR} & 
            (k) AdaIR \cite{AdaIR} & 
            (l) AgenticIR \cite{AgenticIR} & 
            (m) PaAgent & 
            (n) Reference \\ 
        \end{tabular}
    }
    \caption{Visual comparison of different methods on low-light (LOL-V1 \cite{LOL-v1}) and blur (GoPro \cite{GoPro}) images. Our PaAgent yields better visual results on different degraded images, with outputs closer to the reference images.}
    \label{Qual_low_blur}
\end{figure*}

\begin{figure*}[!ht]
    \Large
    \centering
    \resizebox{1\linewidth}{!}{
        \begin{tabular}{c@{ }c@{ }c@{ }c@{ }c@{ }c@{ }c@{ }}
            \includegraphics[height=3cm,width=4cm]{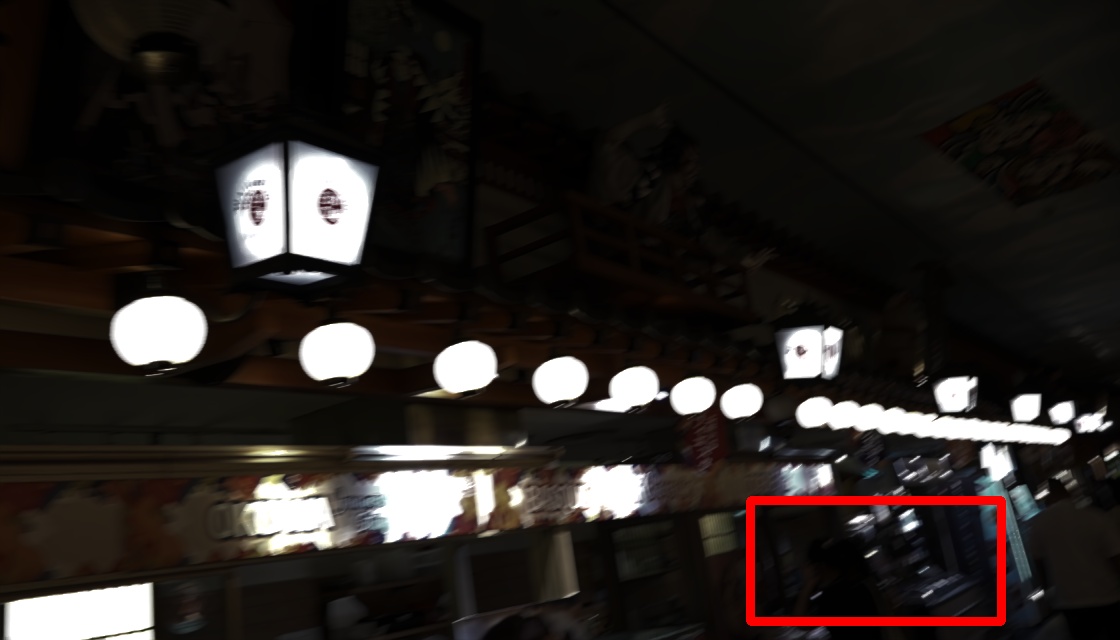} &
            \includegraphics[height=3cm,width=4cm]{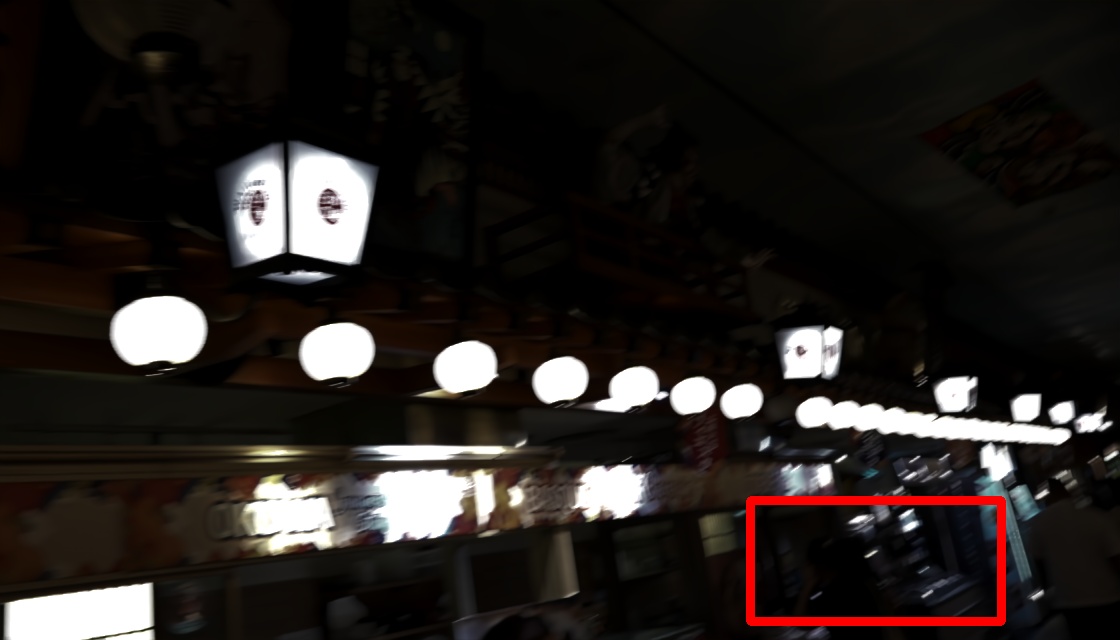} &
            \includegraphics[height=3cm,width=4cm]{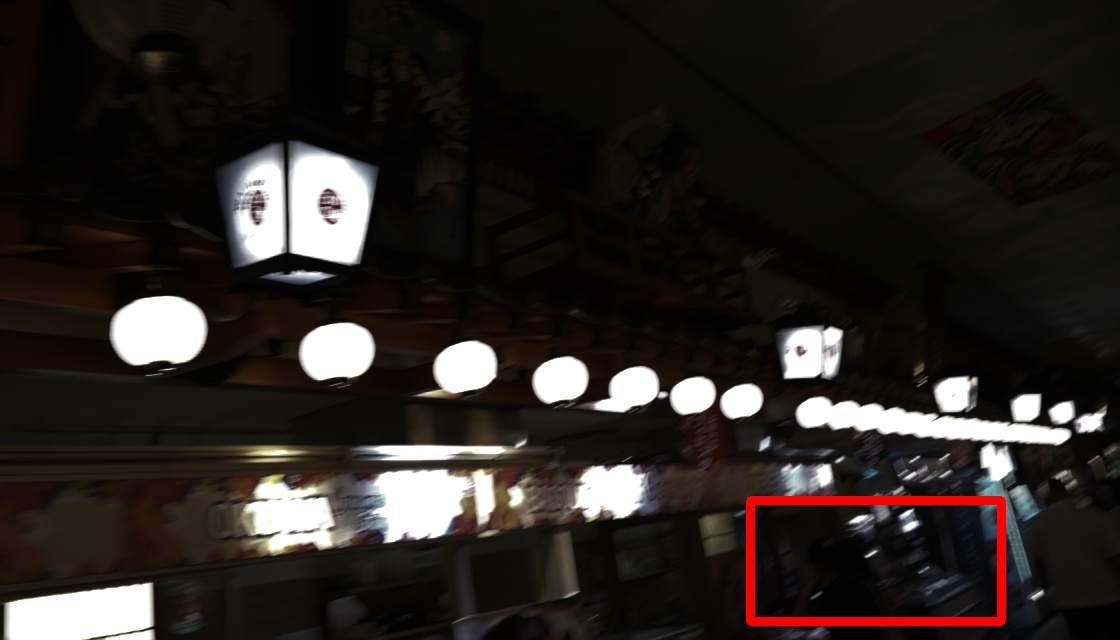} &
            \includegraphics[height=3cm,width=4cm]{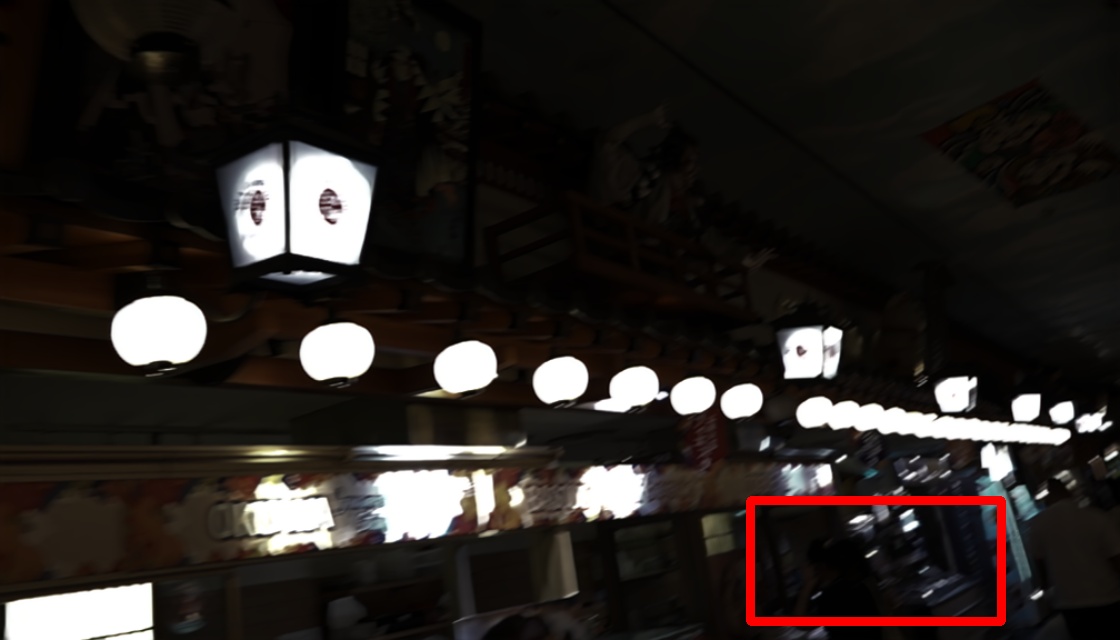} &
            \includegraphics[height=3cm,width=4cm]{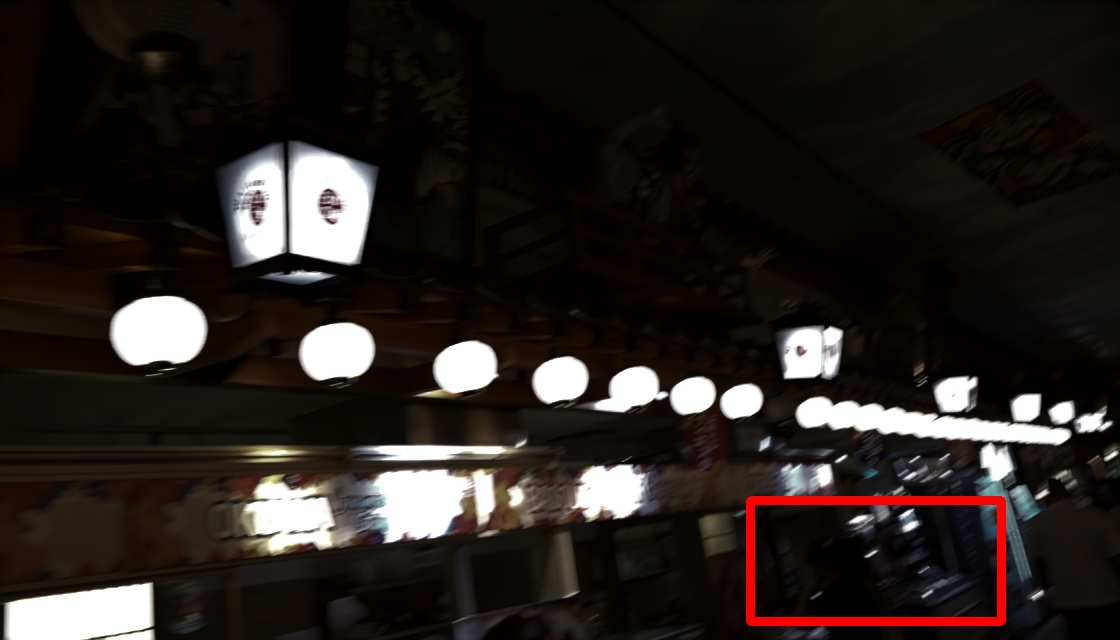} &
            \includegraphics[height=3cm,width=4cm]{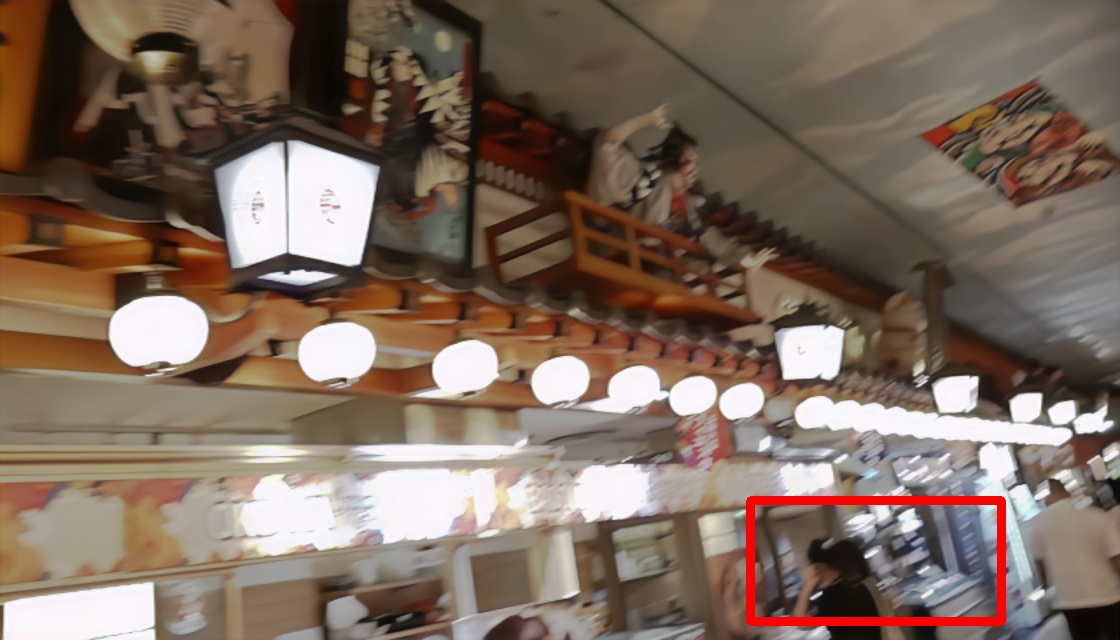} &
            \includegraphics[height=3cm,width=4cm]{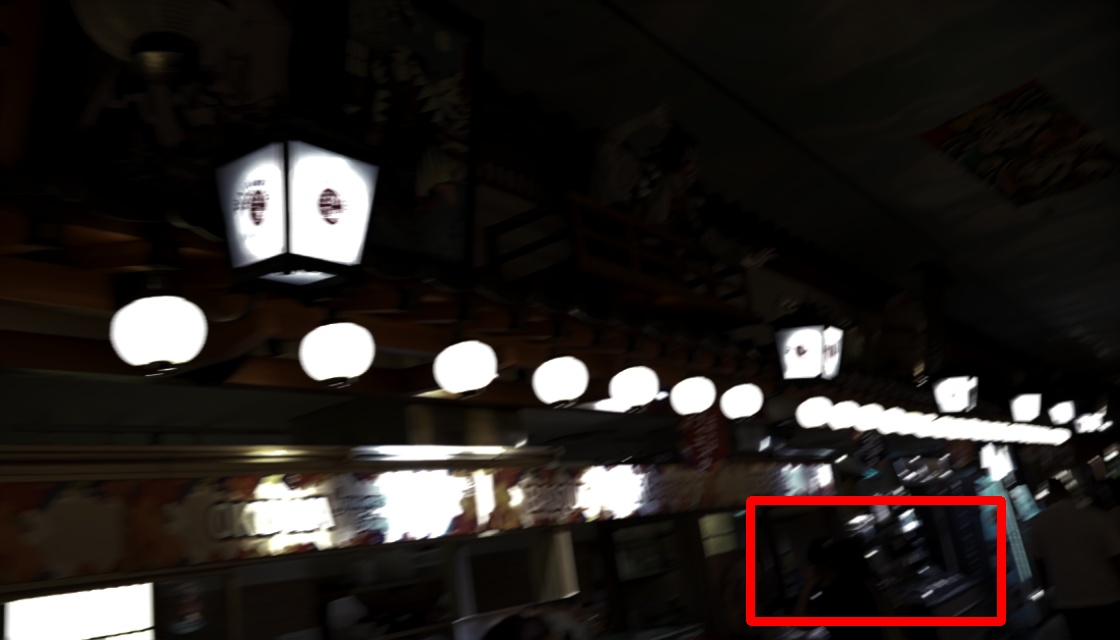} \\

            \includegraphics[height=2cm,width=4cm]{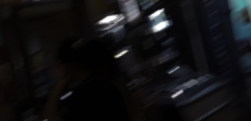} &
            \includegraphics[height=2cm,width=4cm]{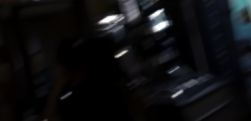} &
            \includegraphics[height=2cm,width=4cm]{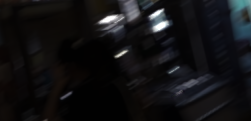} &
            \includegraphics[height=2cm,width=4cm]{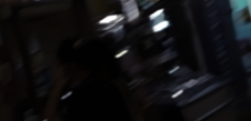} &
            \includegraphics[height=2cm,width=4cm]{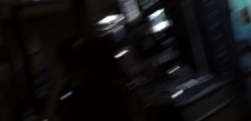} &
            \includegraphics[height=2cm,width=4cm]{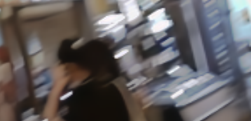} &
            \includegraphics[height=2cm,width=4cm]{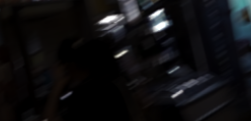} \\

            \includegraphics[height=3cm,width=4cm]{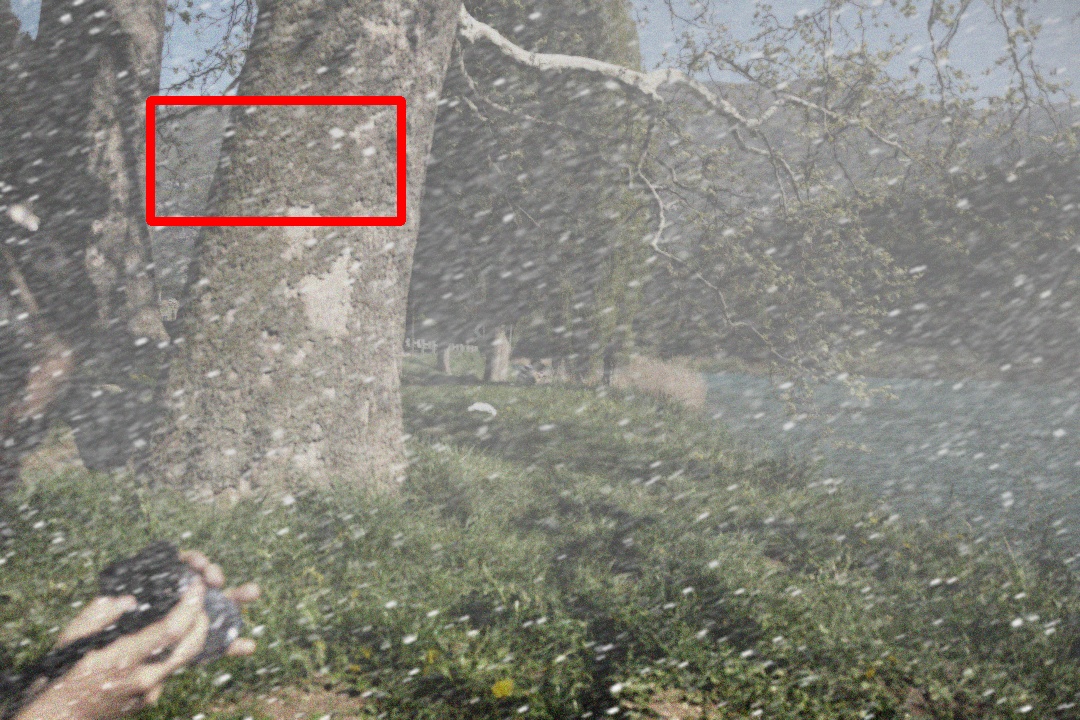} &
            \includegraphics[height=3cm,width=4cm]{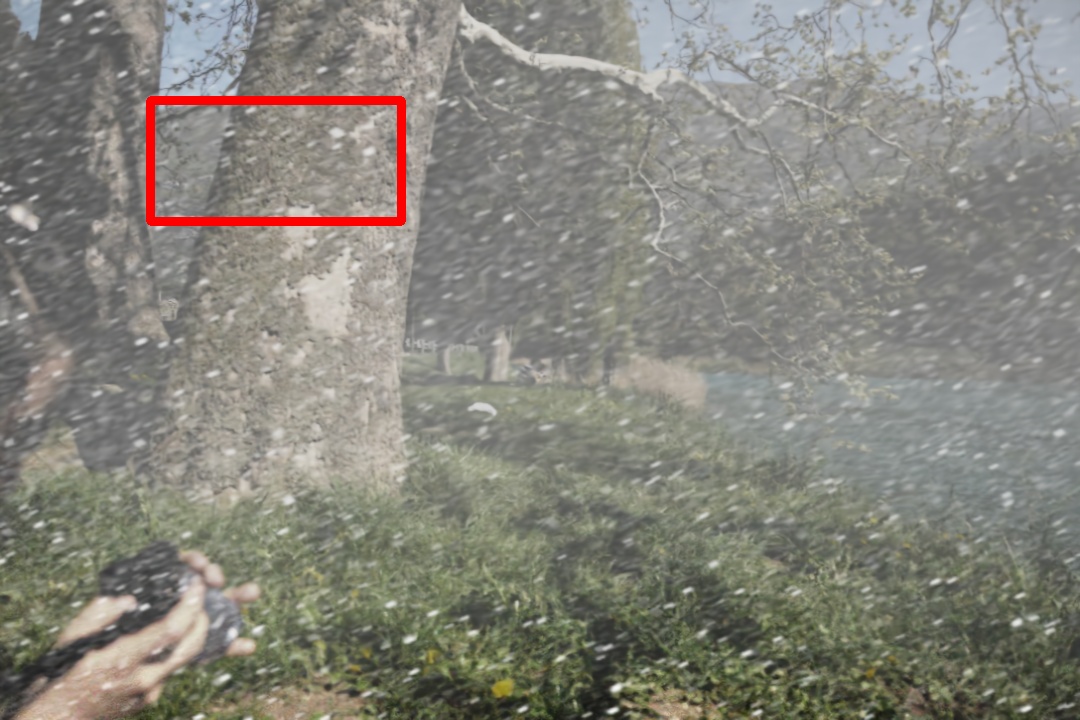} &
            \includegraphics[height=3cm,width=4cm]{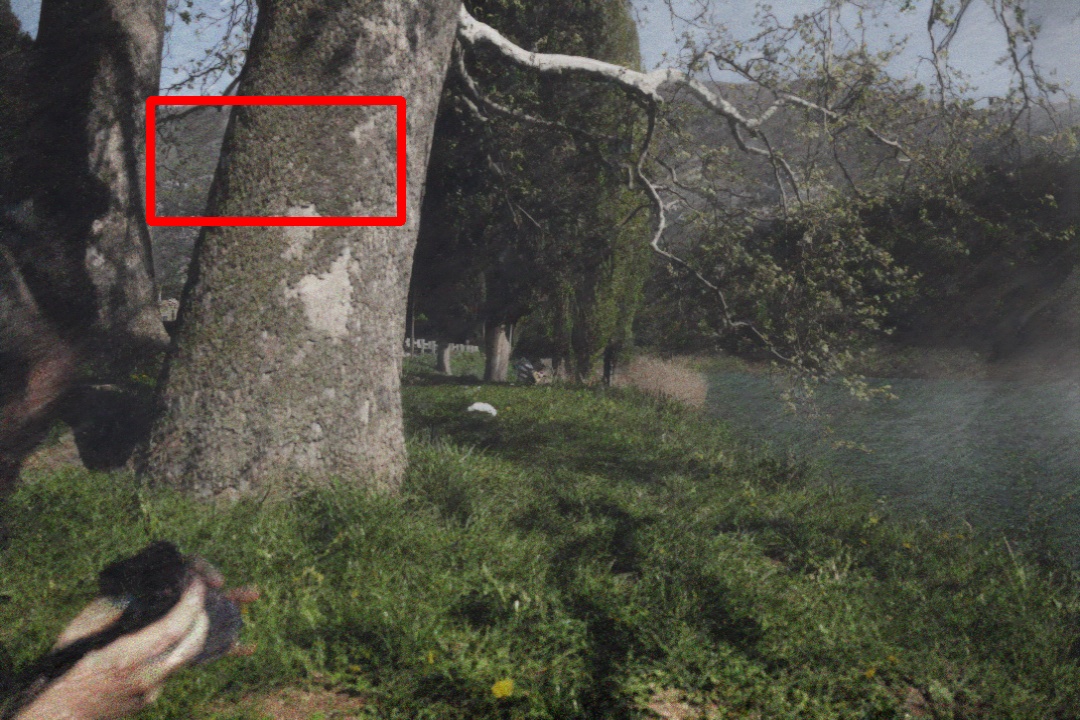} &
            \includegraphics[height=3cm,width=4cm]{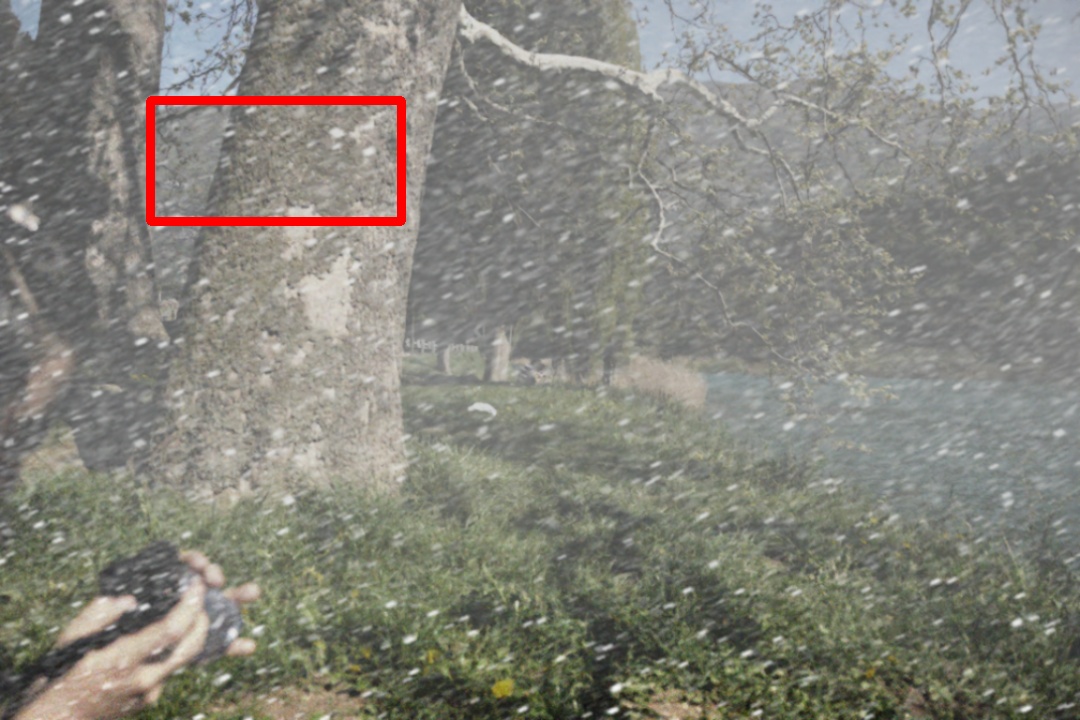} &
            \includegraphics[height=3cm,width=4cm]{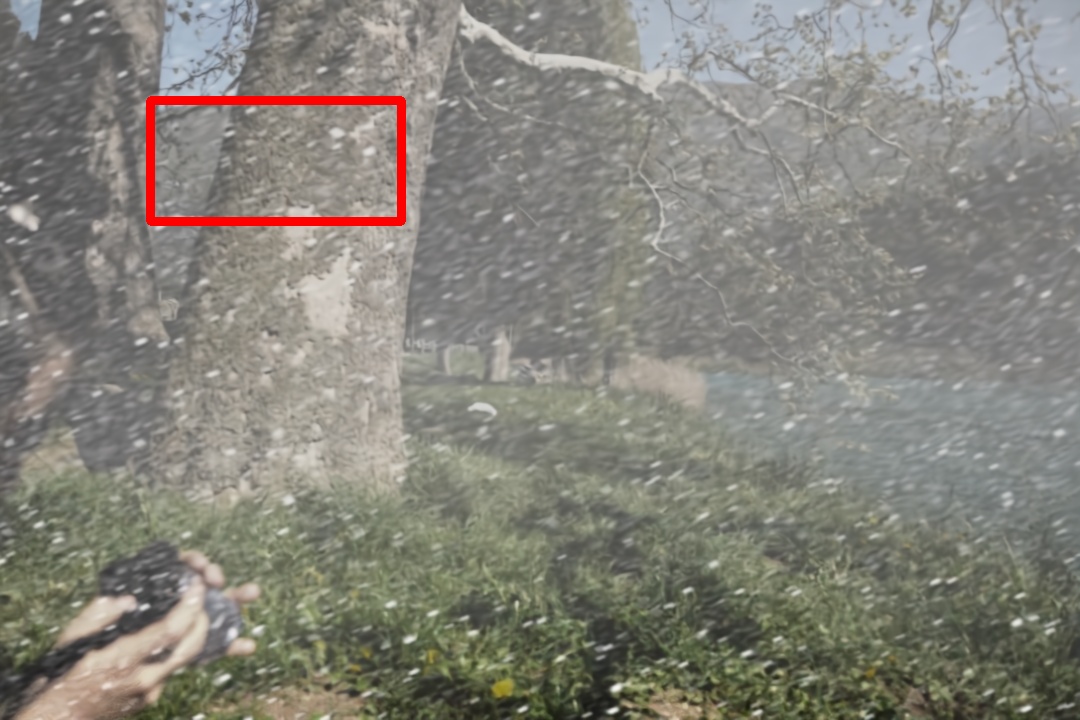} &
            \includegraphics[height=3cm,width=4cm]{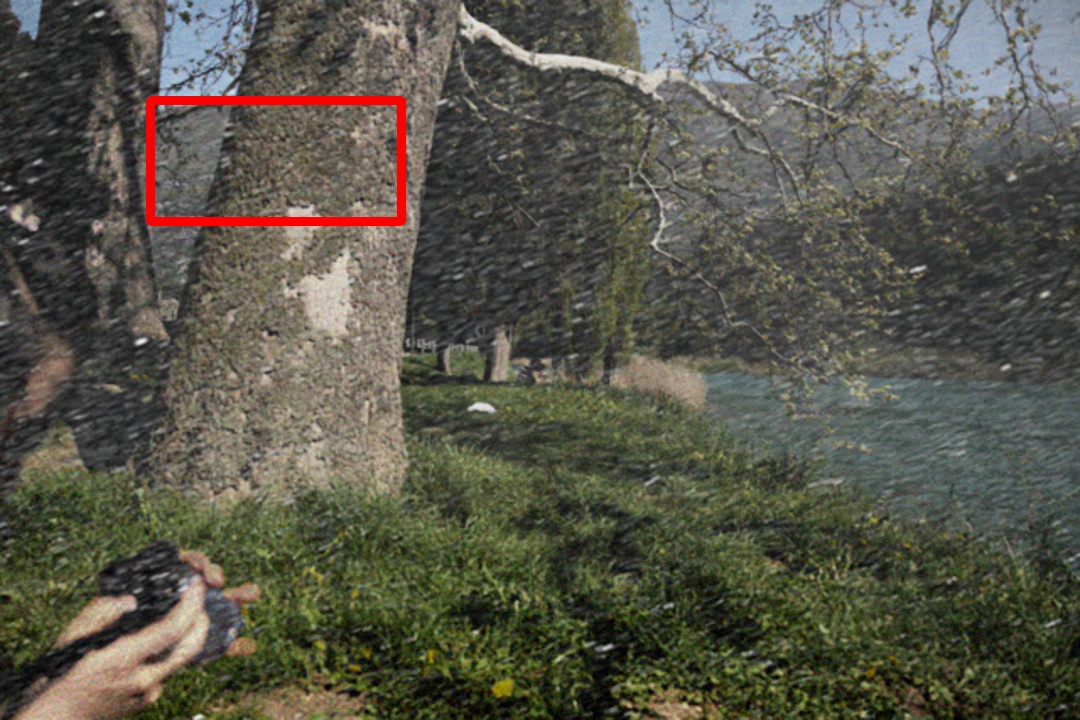} &
            \includegraphics[height=3cm,width=4cm]{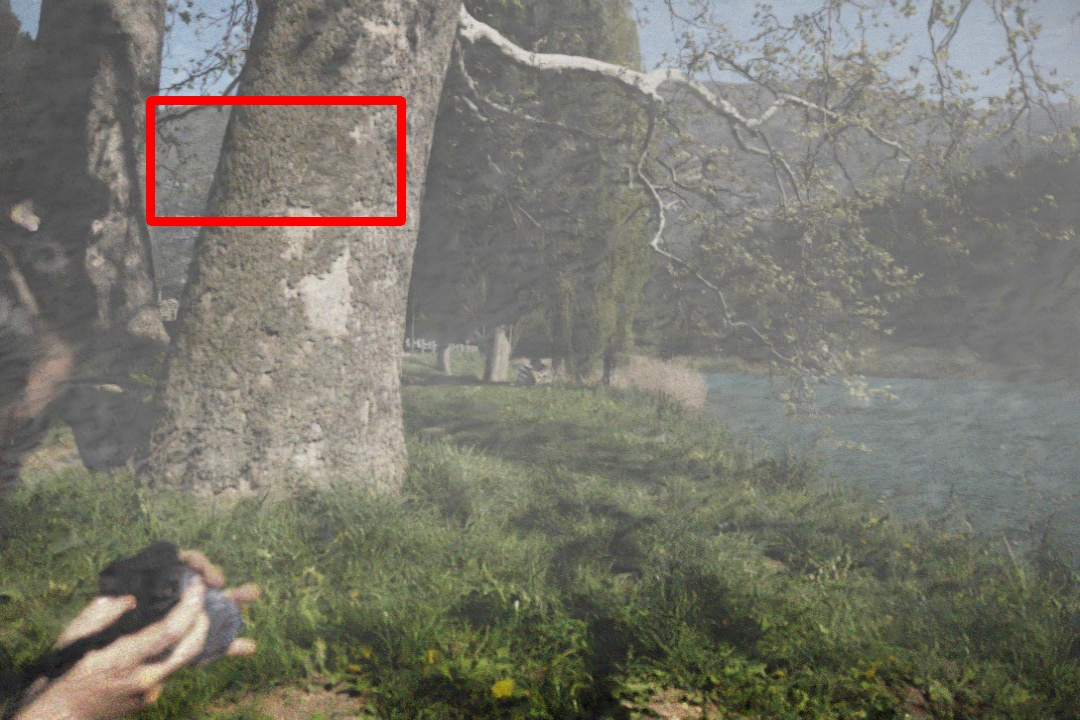} \\

            \includegraphics[height=2cm,width=4cm]{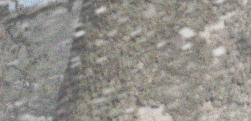} &
            \includegraphics[height=2cm,width=4cm]{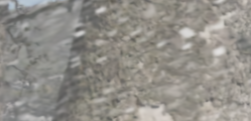} &
            \includegraphics[height=2cm,width=4cm]{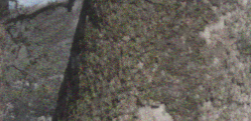} &
            \includegraphics[height=2cm,width=4cm]{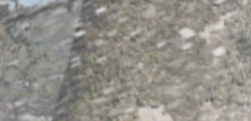} &
            \includegraphics[height=2cm,width=4cm]{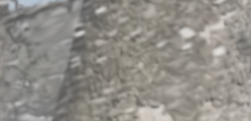} &
            \includegraphics[height=2cm,width=4cm]{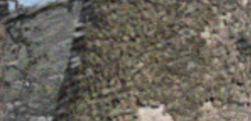} &
            \includegraphics[height=2cm,width=4cm]{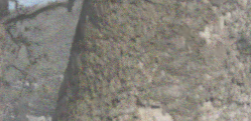} \\

            (a) Input & 
            (b) PromptIR \cite{PromptIR} &
            (c) GridFormer \cite{GridFormer} & 
            (d) RAM \cite{RAM} & 
            (e) NDR-Restore \cite{NDR-Restore} & 
            (f) DGSolver \cite{DGSolver} & 
            (g) AWRaCLe \cite{AWRaCLe} \\

            \includegraphics[height=3cm,width=4cm]{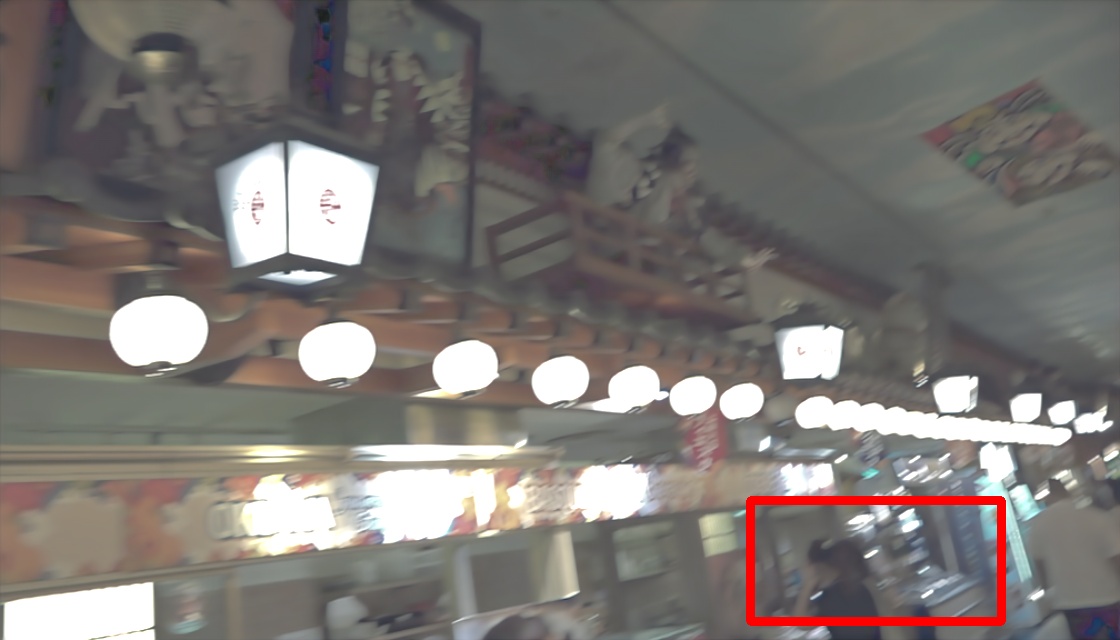} &
            \includegraphics[height=3cm,width=4cm]{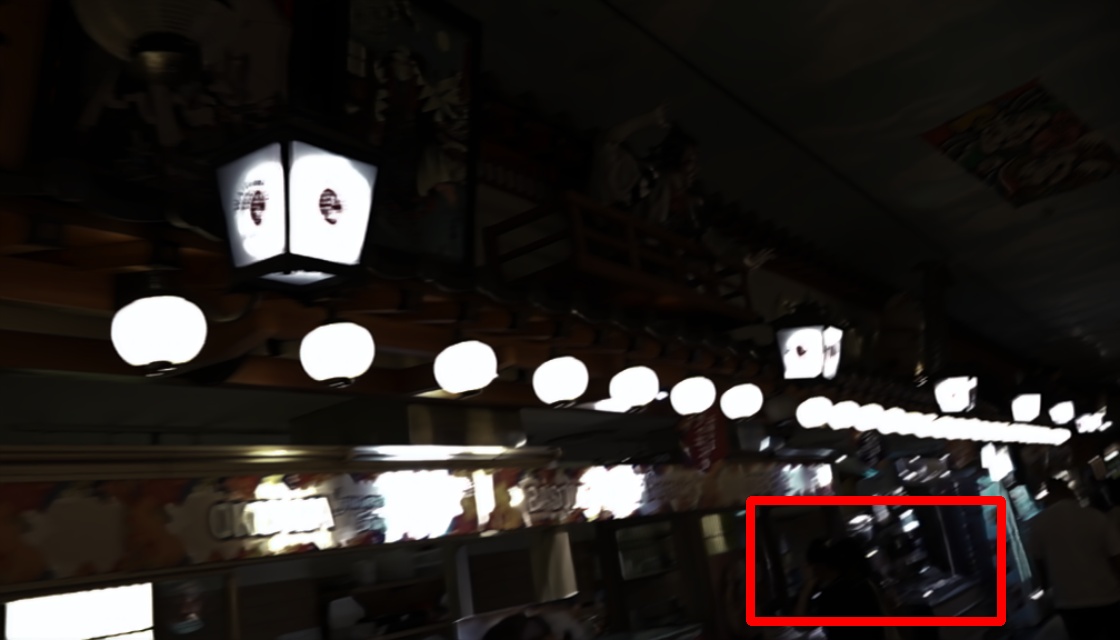} &
            \includegraphics[height=3cm,width=4cm]{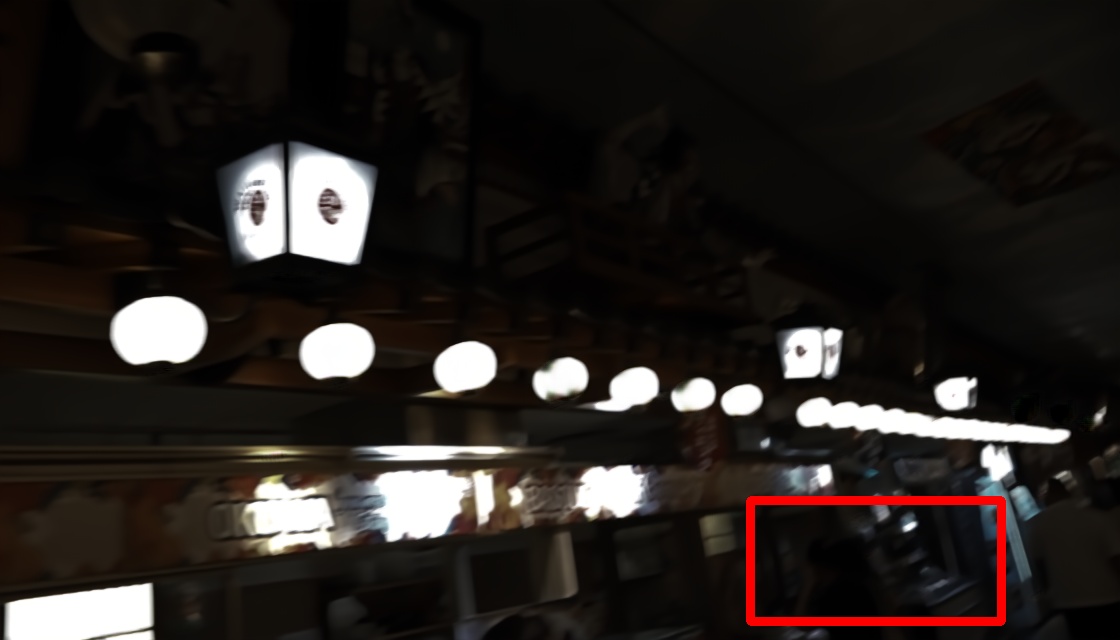} &
            \includegraphics[height=3cm,width=4cm]{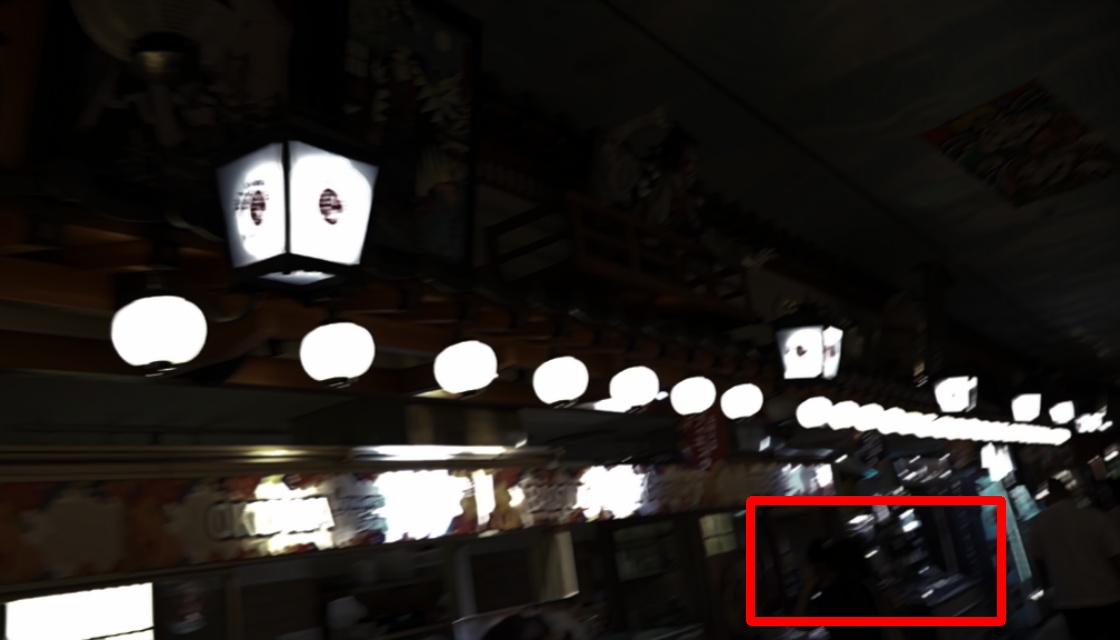} &
            \includegraphics[height=3cm,width=4cm]{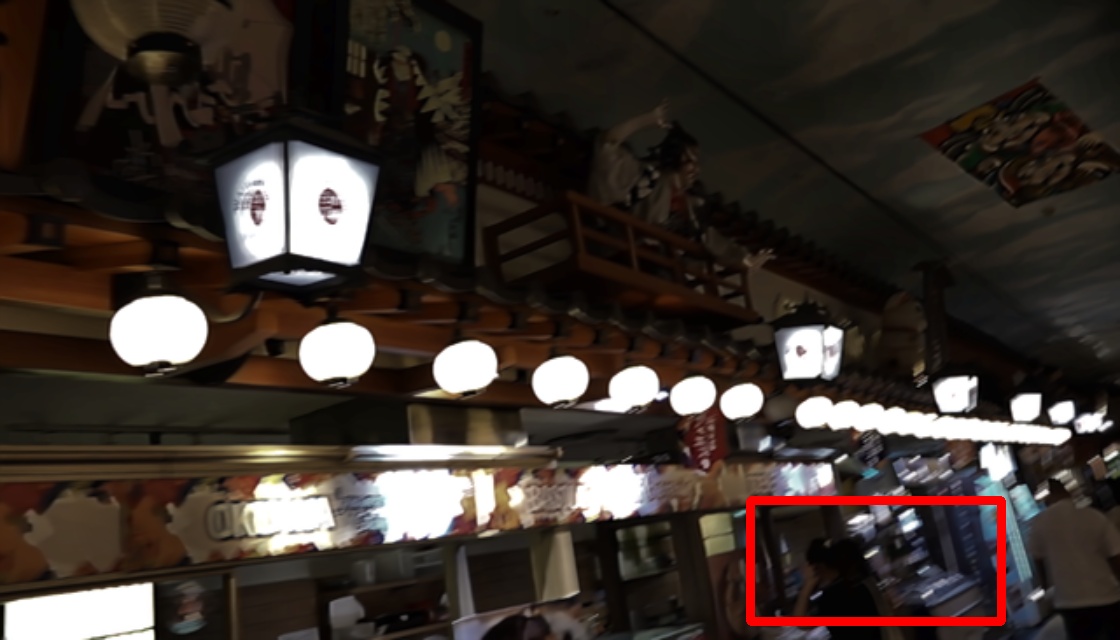} &
            \includegraphics[height=3cm,width=4cm]{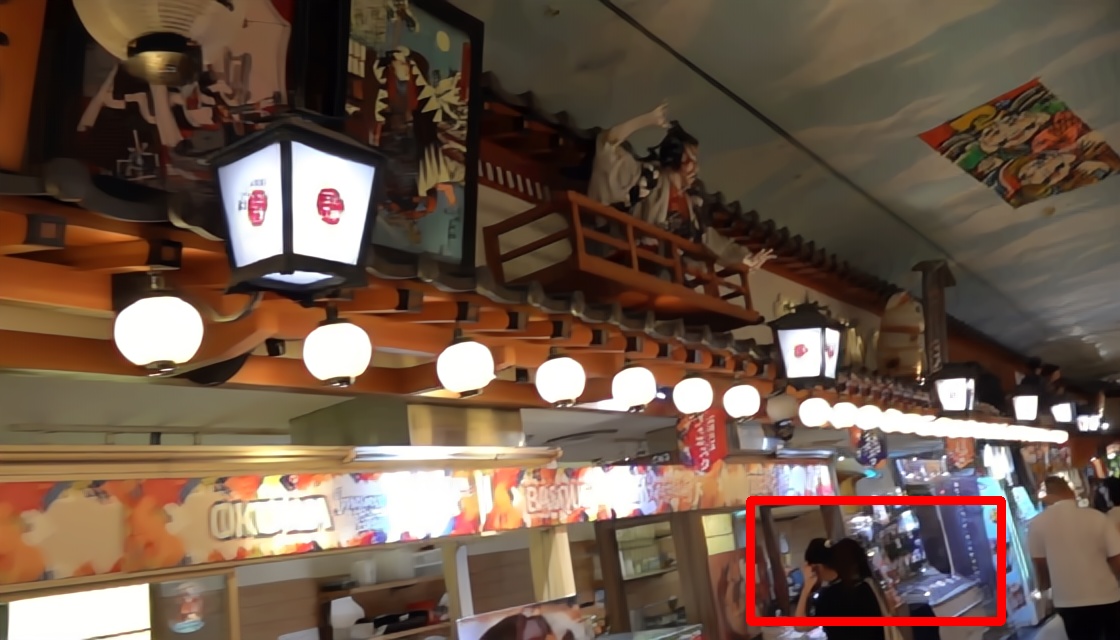} &
            \includegraphics[height=3cm,width=4cm]{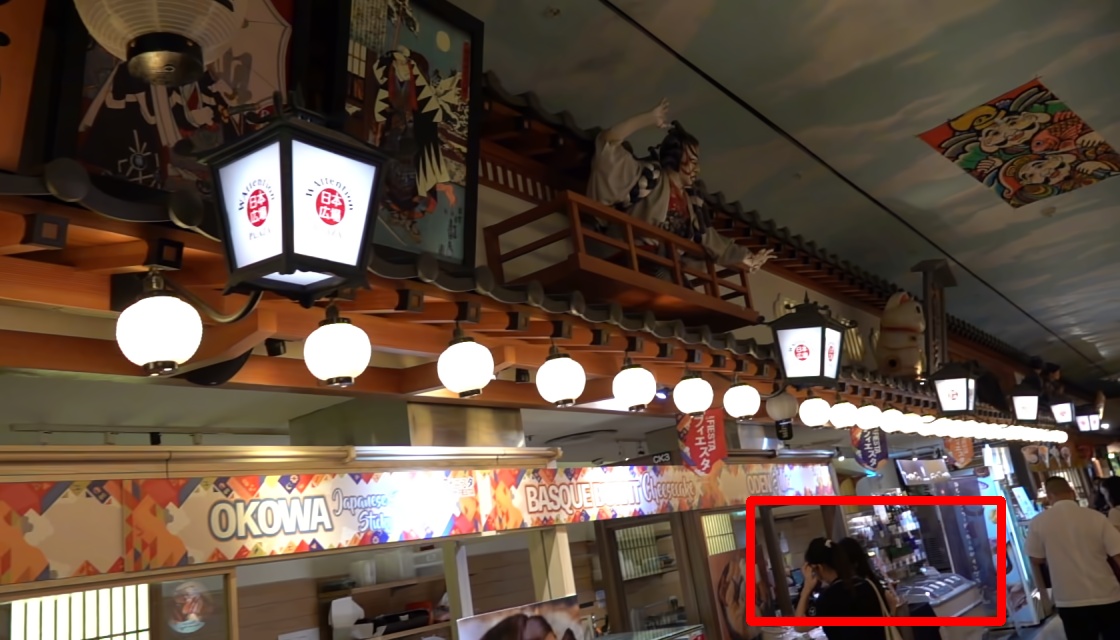} \\

            \includegraphics[height=2cm,width=4cm]{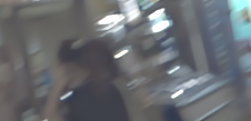} &
            \includegraphics[height=2cm,width=4cm]{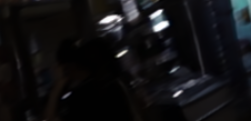} &
            \includegraphics[height=2cm,width=4cm]{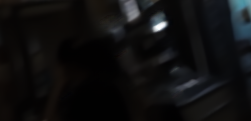} &
            \includegraphics[height=2cm,width=4cm]{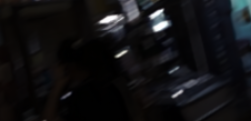} &
            \includegraphics[height=2cm,width=4cm]{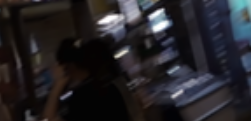} &
            \includegraphics[height=2cm,width=4cm]{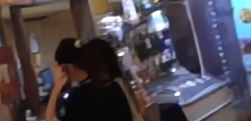} &
            \includegraphics[height=2cm,width=4cm]{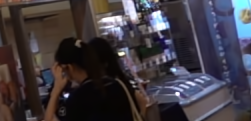} \\

            \includegraphics[height=3cm,width=4cm]{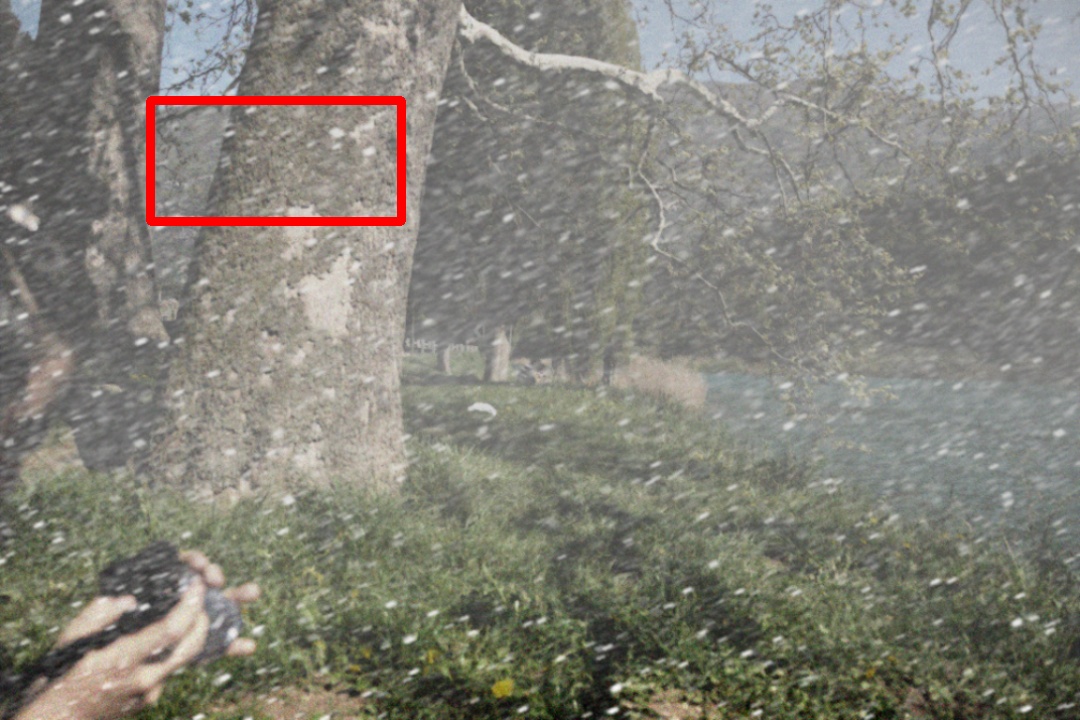} &
            \includegraphics[height=3cm,width=4cm]{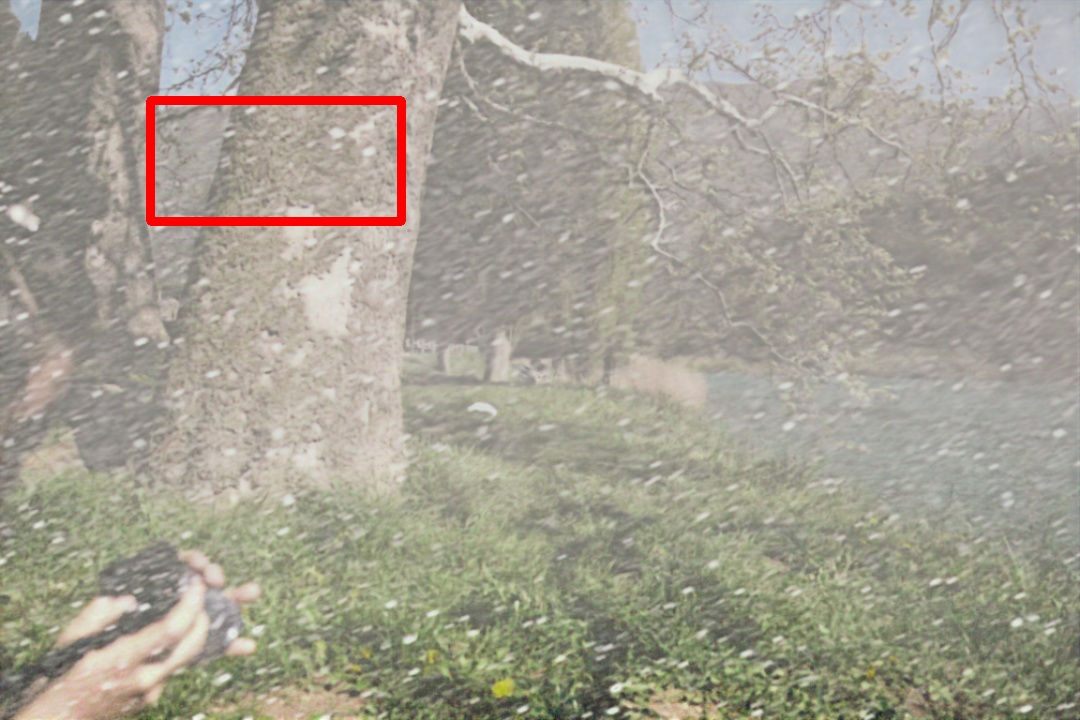} &
            \includegraphics[height=3cm,width=4cm]{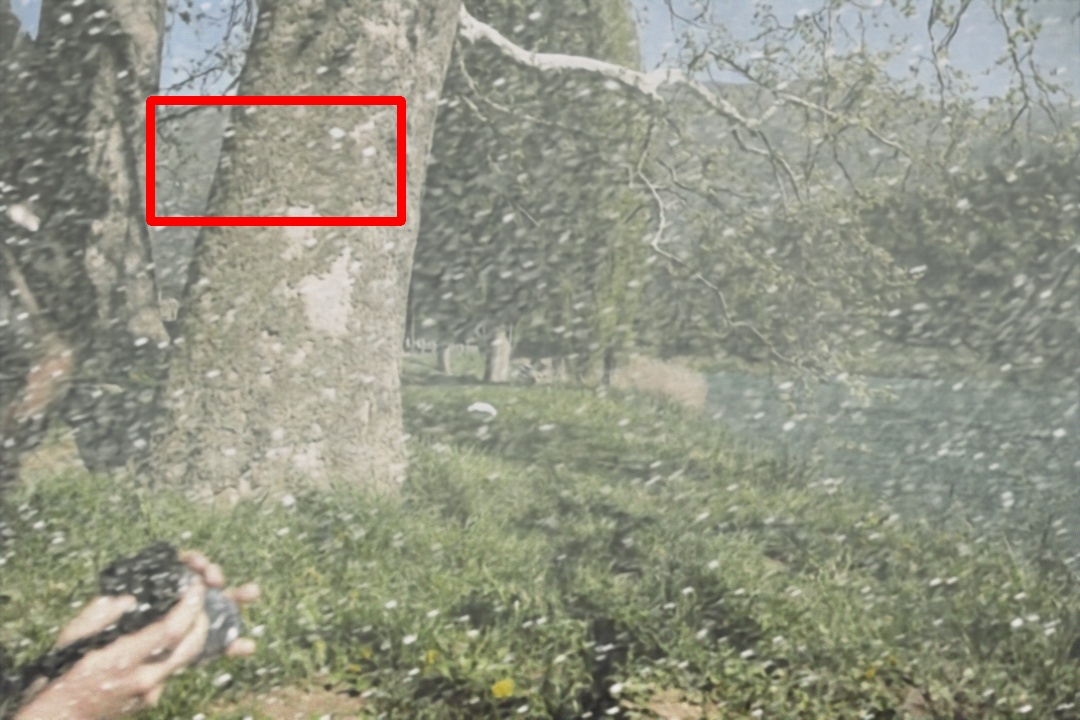} &
            \includegraphics[height=3cm,width=4cm]{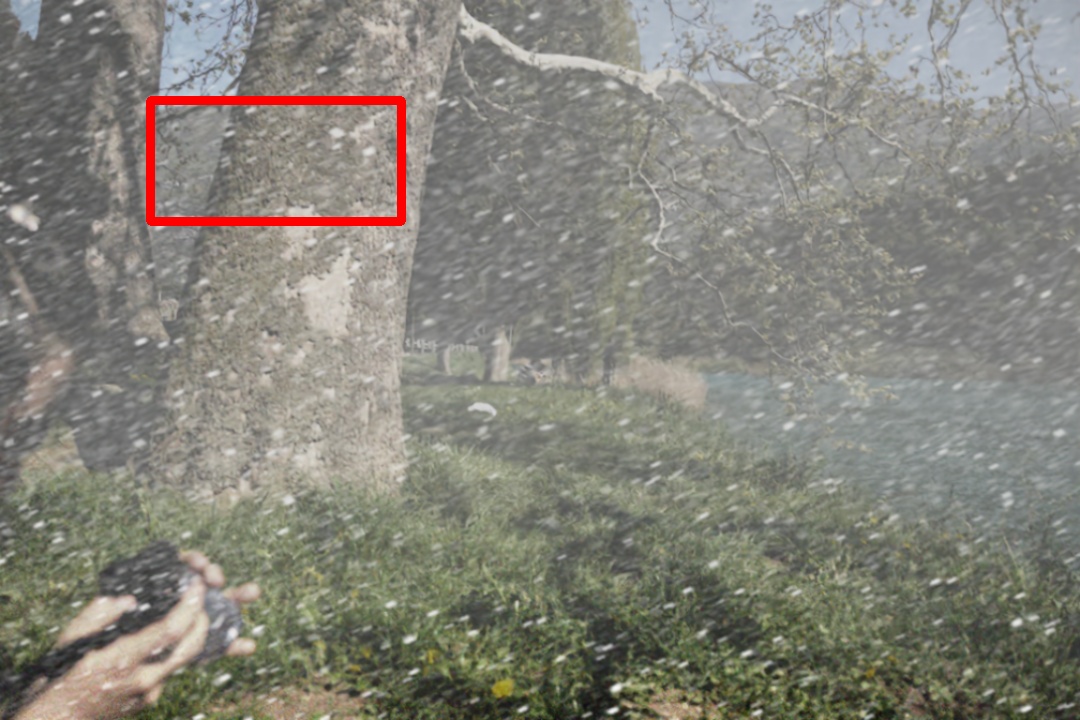} &
            \includegraphics[height=3cm,width=4cm]{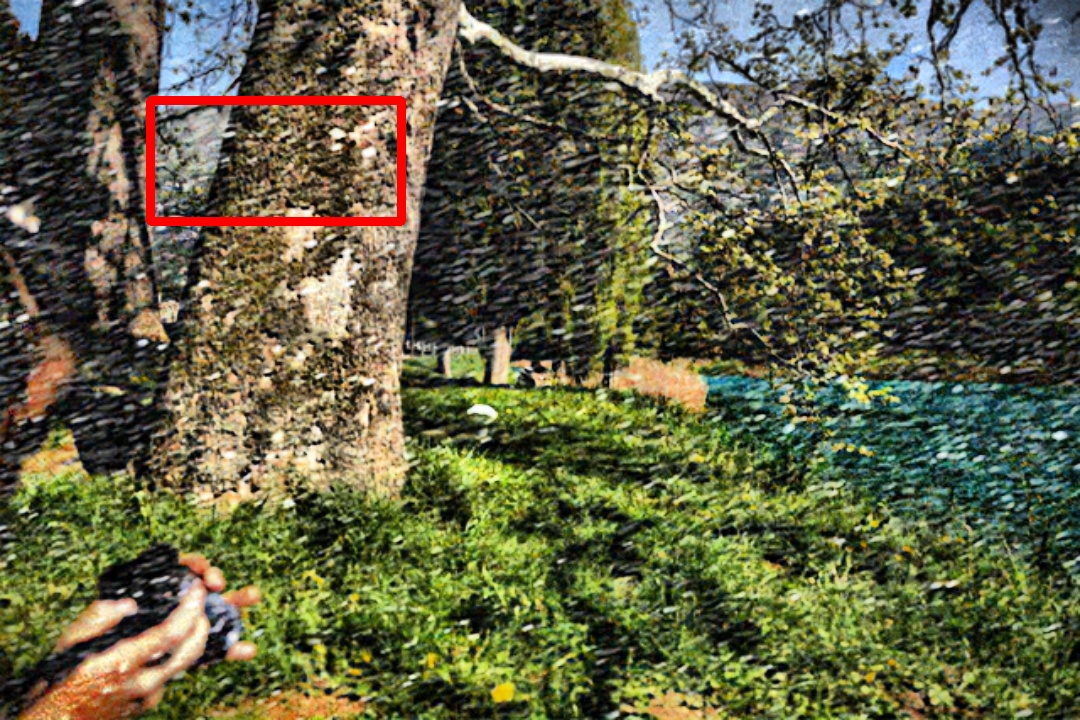} &
            \includegraphics[height=3cm,width=4cm]{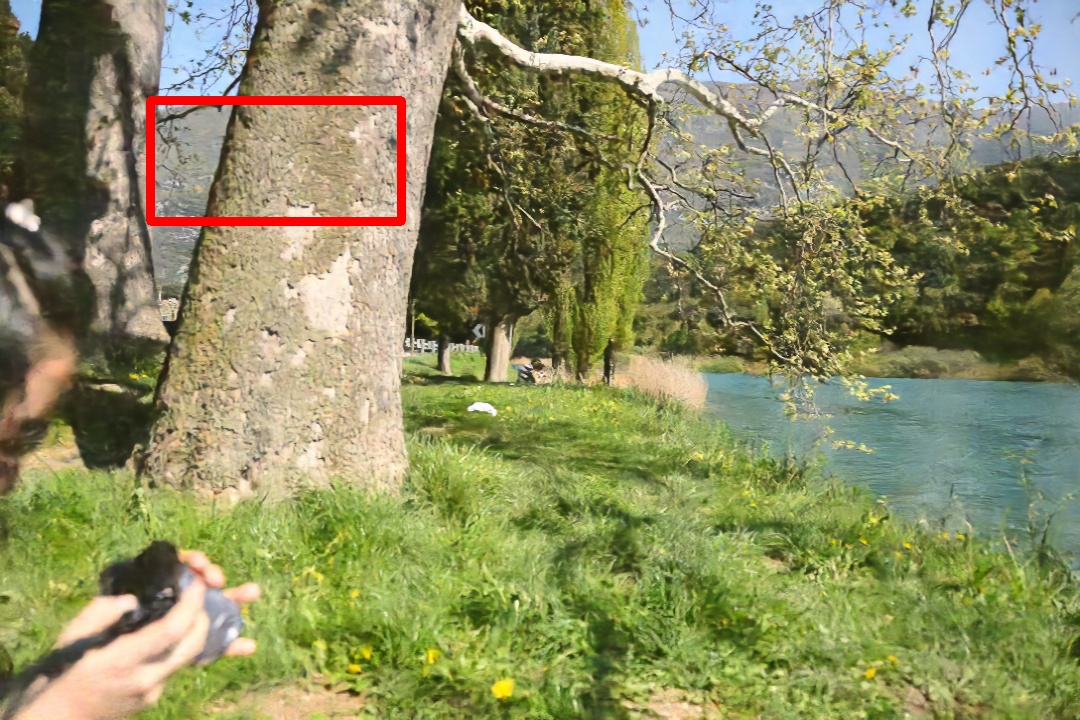} &
            \includegraphics[height=3cm,width=4cm]{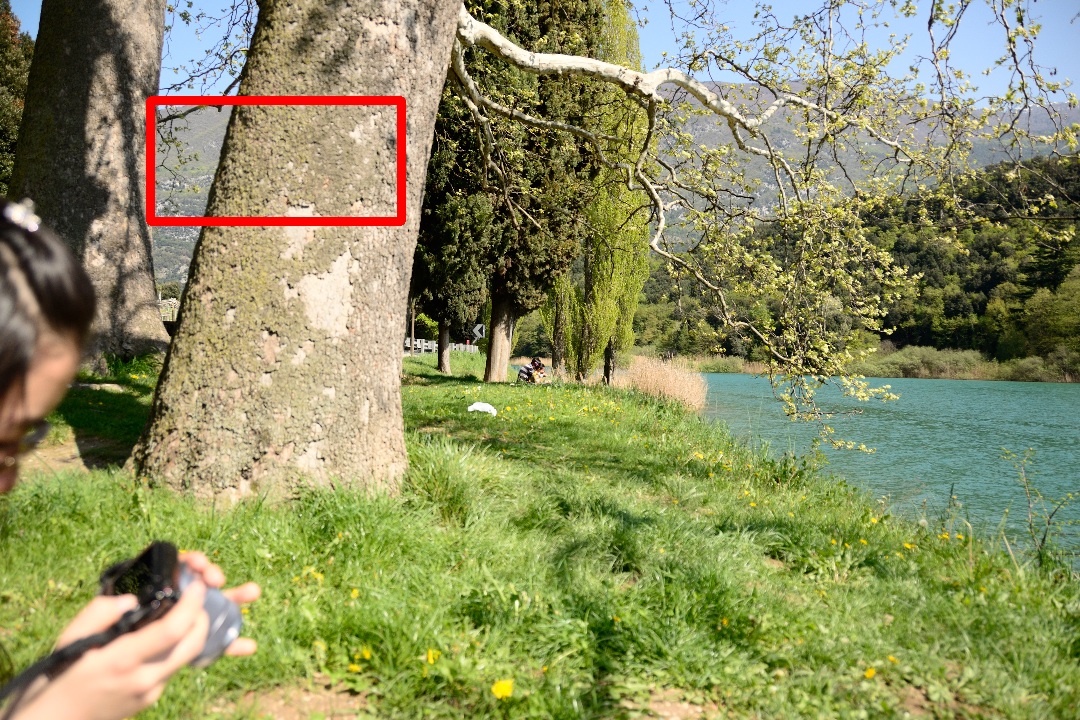} \\

            \includegraphics[height=2cm,width=4cm]{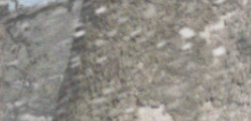} &
            \includegraphics[height=2cm,width=4cm]{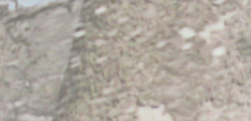} &
            \includegraphics[height=2cm,width=4cm]{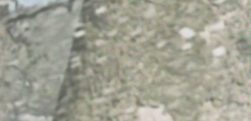} &
            \includegraphics[height=2cm,width=4cm]{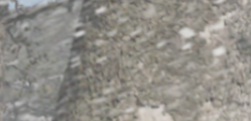} &
            \includegraphics[height=2cm,width=4cm]{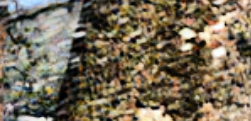} &
            \includegraphics[height=2cm,width=4cm]{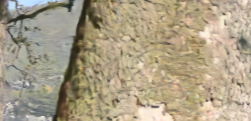} &
            \includegraphics[height=2cm,width=4cm]{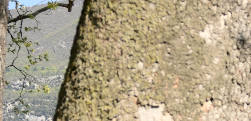} \\

            (h) DFPIR \cite{DFPIR} & 
            (i) DA-RCOT \cite{DA-RCOT} &
            (j) CPLIR \cite{CPLIR} & 
            (k) AdaIR \cite{AdaIR} & 
            (l) AgenticIR \cite{AgenticIR} & 
            (m) PaAgent & 
            (n) Reference \\ 
        \end{tabular}
    }
    \caption{Visual comparison of different methods on composite degradation (LOL\_Blur \cite{LEDNet} and CDD-11 \cite{OneRestore}) images. Our PaAgent yields better visual results on different degraded images, with outputs closer to the reference images.}
    \label{Qual_composit}
\end{figure*}

\subsection{Qualitative Evaluation}
We conduct the visual comparisons of different methods on various benchmark datasets \cite{CSD, BSD68, RESIDE, MPRNet, LOL-v1, GoPro, OneRestore, LEDNet}, as shown in Figs. \ref{Qual_snow_noise}, \ref{Qual_haze_rain}, \ref{Qual_low_blur}, and \ref{Qual_composit}.
Fig. \ref{Qual_snow_noise} presents the results on desnowing (CSD \cite{CSD}) and denoising (BSD68 \cite{BSD68}) tasks.
While DGSolver \cite{DGSolver} and AgenticIR \cite{AgenticIR} demonstrate robustness under dense snowfall, they compromise the fidelity of fine textural details.
RAM \cite{RAM} and NDR-Restore \cite{NDR-Restore} achieve visually plausible results yet suffer from residual noise and artifacts within complex textural regions.
By contrast, our PaAgent effectively suppresses snow streaks and noise across diverse scenarios while preserving fine structural details, such as architectural textures, animal fur, and foliage.
As illustrated in Fig. \ref{Qual_haze_rain}, we evaluate the restoration performance on dehazing (SOTS \cite{RESIDE}) and deraining (Rain13K \cite{MPRNet}) tasks.
Existing methods demonstrate satisfactory capability in handling global atmospheric haze and rain streaks, yet closer inspection exposes their struggles with fine-detail preservation. 
Furthermore, RAM \cite{RAM}, DGSolver \cite{DGSolver} and CPLIR \cite{CPLIR} exhibit impressive performance in the deraining task but leave residual rain streaks in local regions of the image.
In contrast, our PaAgent not only recovers sharper building boundaries in complex haze but also removes dense rain streaks that competing methods fail to address.
Fig. \ref{Qual_low_blur} shows the results on LLIE (LOL-V1 \cite{LOL-v1}) and deblurring (GoPro \cite{GoPro}) tasks.
RAM \cite{RAM} DGSolver \cite{DGSolver}, DFPIR \cite{DFPIR}, DA-RCOT \cite{DA-RCOT}, and AdaIR \cite{AdaIR} demonstrate strong capability in enhancing global visibility; however, they inevitably introduce noticeable color casts in the stuffed toy regions or fail to recover the local fine details of the drum.
AWRaCLe \cite{AWRaCLe}, AdaIR \cite{AdaIR}, and AgenticIR \cite{AgenticIR} successfully restore the general shape of the petals and reduce motion blur, yet struggle with fine texture reconstruction.
In comparison, our proposed PaAgent method efficiently addresses illumination deficiency and motion blurring, yielding outputs with richer semantic content and sharper structural details.
As shown in Fig. \ref{Qual_composit}, we present the results under composite degradation scenarios (i.e., LOL\_Blur \cite{LEDNet} and CDD-11 \cite{OneRestore}).
DGSolver \cite{DGSolver} and DFPIR \cite{DFPIR} demonstrate competence in illumination adjustment, but they fail to simultaneously eliminate motion blur.
GridFormer \cite{GridFormer} and DGSolver \cite{DGSolver} exhibit impressive performance in handling snow+haze+low-light degradations, but they sacrifice fine textural details such as tree barks. 
AgenticIR \cite{AgenticIR} excels in recovering high-frequency information yet exhibits limited capability in contrast adjustment. 
By intelligently coordinating specialized restoration experts, our PaAgent method can restore results of both fine details and clear structures, which suggests that our method is more robust in handling diverse real-world degradations.

\begin{table*}[!ht]
\centering
\caption{Quantitative results of different methods on desnowing and denoising datasets. $\sigma$ denotes the noise level The \textcolor{red}{best} results are marked in red.}
\label{tab_snow_noise}
\resizebox{1\linewidth}{!}{
\begin{tabular}{c|c|cc|cc|cc|cc|cc|cc|cc}

\hline

\rowcolor[HTML]{FFCCC9} 
\cellcolor[HTML]{FFCCC9}                         & 
\cellcolor[HTML]{FFCCC9}                        & 
\multicolumn{2}{c|}{\cellcolor[HTML]{FFCCC9}\textbf{Desnowing}} & 
\multicolumn{12}{c}{\cellcolor[HTML]{FFCCC9}\textbf{Denoising} (BSD68 \cite{BSD68})}   \\ \cline{3-16}

\rowcolor[HTML]{FFCCC9} 
\cellcolor[HTML]{FFCCC9}                         & 
\cellcolor[HTML]{FFCCC9}                        & 
\multicolumn{2}{c|}{\cellcolor[HTML]{FFCCC9}CSD \cite{CSD}}       & 
\multicolumn{2}{c|}{\cellcolor[HTML]{FFCCC9}$\sigma$=5} & 
\multicolumn{2}{c|}{\cellcolor[HTML]{FFCCC9}$\sigma$=10} & 
\multicolumn{2}{c|}{\cellcolor[HTML]{FFCCC9}$\sigma$=15} & 
\multicolumn{2}{c|}{\cellcolor[HTML]{FFCCC9}$\sigma$=25} & 
\multicolumn{2}{c|}{\cellcolor[HTML]{FFCCC9}$\sigma$=35} & 
\multicolumn{2}{c}{\cellcolor[HTML]{FFCCC9}$\sigma$=50} \\ \cline{3-16}

\rowcolor[HTML]{FFCCC9} 
\multirow{-3}{*}{\cellcolor[HTML]{FFCCC9}Method} & 
\multirow{-3}{*}{\cellcolor[HTML]{FFCCC9}Venue} & 
PSNR                                                 & SSIM                                                & 
PSNR                                                 & SSIM                                                & 
PSNR                                                 & SSIM                                                & 
PSNR                                                 & SSIM                                                & 
PSNR                                                 & SSIM                                                & 
PSNR                                                 & SSIM                                                & 
PSNR                                                 & SSIM                                                \\

\hline
\hline

PromptIR \cite{PromptIR}                                        & 
NIPS'23                                         & 
20.45                      & 0.79                     & 27.00                    & 0.80                    & 33.11                     & 0.93                    & 31.64                     & 0.90                    & 28.41                     & 0.82                    & 26.22                     & 0.75                    & 23.62                     & 0.65                    \\

GridFormer \cite{GridFormer}                                      & 
IJCV'24                                         & 
15.94                      & 0.79                     & 24.64                    & 0.69                    & 28.82                     & 0.88                    & 27.00                     & 0.81                    & 23.85                     & 0.67                    & 21.68                     & 0.56                    & 19.18                     & 0.44                    \\

RAM \cite{RAM}                                             & 
ECCV'24                                        & 
20.08                      & 
0.79                     & 
30.60                    & 
0.86                    & 
33.16                     & 
0.93                    & 
31.94                     & 
0.91                    & 
30.06                     & 
0.86                    & 
28.60                     & 
0.82                    & 
26.74                     & 
0.76                    \\

NDR-Restore \cite{NDR-Restore}                                     & TIP'24                                          & 20.06                      & 0.78                     & 27.38                    & 0.79                    & 32.29                     & 0.91                    & 31.98                     & 0.90                    & 29.29                     & 0.84                    & 27.67                     & 0.79                    & 23.78                     & 0.65                    \\
DGSolver \cite{DGSolver}                                        & NIPS'25                                         & 32.69                      & 0.96                     & 25.47                    & 0.65                    & 29.62                     & 0.85                    & 26.82                     & 0.76                    & 22.91                     & 0.59                    & 20.29                     & 0.47                    & 17.73                     & 0.36                    \\
AWRaCLe \cite{AWRaCLe}                                         & AAAI'25                                         & 15.46                      & 0.78                     & 24.92                    & 0.69                    & 29.65                     & 0.89                    & 27.29                     & 0.82                    & 23.43                     & 0.68                    & 20.51                     & 0.56                    & 17.73                     & 0.43                    \\
DFPIR \cite{DFPIR}                                           & CVPR'25                                         & 13.03                      & 0.69                     & 21.19                    & 0.67                    & 26.75                     & 0.88                    & 24.95                     & 0.82                    & 23.83                     & 0.70                    & 21.53                     & 0.59                    & 17.85                     & 0.44                    \\
DA-RCOT \cite{DA-RCOT}                                         & TPAMI'25                                        & 20.21                      & 0.78                     & 18.46                    & 0.66                    & 19.11                     & 0.79                    & 31.00                     & 0.88                    & 30.21                     & 0.86                    & 27.03                     & 0.76                    & 21.98                     & 0.56                    \\
CPLIR \cite{CPLIR}                                           & TPAMI'25                                        & 15.56                      & 0.73                     & 24.03                    & 0.74                    & 29.30                     & 0.88                    & 26.47                     & 0.83                    & 21.55                     & 0.74                    & 19.11                     & 0.67                    & 17.02                     & 0.58                    \\
AdaIR \cite{AdaIR}                                           & ICLR'25                                         & 20.44                      & 0.79                     & 24.19                    & 0.74                    & 33.28                     & 0.93                    & 31.93                     & 0.91                    & 28.05                     & 0.81                    & 26.32                     & 0.75                    & 22.99                     & 0.61                    \\
AgenticIR \cite{AgenticIR}                                       & ICLR'25                                         & 34.06                      & 0.97                     & 10.13                    & 0.31                    & 12.35                     & 0.36                    & 9.59                      & 0.30                    & 10.24                     & 0.33                    & 10.05                     & 0.33                    & 9.77                      & 0.33                    \\

\hline

Our                                              & -                                               & 
\textcolor{red}{37.95}                      & 
\textcolor{red}{0.98}                     & 
\textcolor{red}{33.89}                    & 
\textcolor{red}{0.88}                    & 
\textcolor{red}{36.74}                     & 
\textcolor{red}{0.96}                    & 
\textcolor{red}{34.39}                     & 
\textcolor{red}{0.94}                    & 
\textcolor{red}{31.40}                     & 
\textcolor{red}{0.89}                    & 
\textcolor{red}{29.36}                     & 
\textcolor{red}{0.85}                    & 
\textcolor{red}{27.03}                     & 
\textcolor{red}{0.79}                   \\

\hline

\end{tabular}
}
\end{table*}

\begin{table*}[!ht]
\centering
\caption{Quantitative results of different methods on dehazing and deraining datasets. The \textcolor{red}{best} results are marked in red.}
\label{tab_haze_rain}
\resizebox{1\linewidth}{!}{
\begin{tabular}{c|c|cc|cc|cc|cc|cc|cc|cc}

\hline

\rowcolor[HTML]{FFCCC9} 
\cellcolor[HTML]{FFCCC9}                         & 
\cellcolor[HTML]{FFCCC9}                        & 
\multicolumn{4}{c|}{\cellcolor[HTML]{FFCCC9}\textbf{Dehazing} (SOTS \cite{RESIDE})}&
\multicolumn{10}{c}{\cellcolor[HTML]{FFCCC9}\textbf{Deraining} (Rain13K \cite{MPRNet})}\\ \cline{3-16}

\rowcolor[HTML]{FFCCC9} 
\cellcolor[HTML]{FFCCC9}                         & 
\cellcolor[HTML]{FFCCC9}                        & 
\multicolumn{2}{c|}{\cellcolor[HTML]{FFCCC9}Indoor}       & 
\multicolumn{2}{c|}{\cellcolor[HTML]{FFCCC9}Outdoor} & 
\multicolumn{2}{c|}{\cellcolor[HTML]{FFCCC9}Rain100L} & 
\multicolumn{2}{c|}{\cellcolor[HTML]{FFCCC9}Rain100H} & 
\multicolumn{2}{c|}{\cellcolor[HTML]{FFCCC9}Test100} & 
\multicolumn{2}{c|}{\cellcolor[HTML]{FFCCC9}Test1200} & 
\multicolumn{2}{c}{\cellcolor[HTML]{FFCCC9}Test2800} \\ \cline{3-16}

\rowcolor[HTML]{FFCCC9} 
\multirow{-3}{*}{\cellcolor[HTML]{FFCCC9}Method} & 
\multirow{-3}{*}{\cellcolor[HTML]{FFCCC9}Venue} & 
PSNR                                                 & SSIM                                                & 
PSNR                                                 & SSIM                                                & 
PSNR                                                 & SSIM                                                & 
PSNR                                                 & SSIM                                                & 
PSNR                                                 & SSIM                                                & 
PSNR                                                 & SSIM                                                & 
PSNR                                                 & SSIM                                                \\

\hline
\hline

PromptIR \cite{PromptIR}                                        & NIPS'23                                         & 23.52 & 0.93 & 27.12 & 0.93 & 37.34 & \textcolor{red}{0.98} & 15.66 & 0.53 & 22.37 & 0.73 & 24.25 & 0.76 & 25.46 & 0.83 
\\

GridFormer \cite{GridFormer}                                      & IJCV'24                                         & 11.96 & 0.70 & 17.30 & 0.84 & 25.40 & 0.85 & 12.22 & 0.39 & 21.80 & 0.69 & 23.12 & 0.71 & 23.87 & 0.77 
\\

RAM \cite{RAM}                                             & ECCV'24                                        & 13.09 & 0.74 & 28.23 & 0.95 & 28.16 & 0.90 & 26.44 & 0.85 & 25.72 & 0.86 & 31.14 & 0.90 & 31.20 & 0.92 
\\

NDR-Restore \cite{NDR-Restore}                                     & TIP'24                                          & 17.27 & 0.85 & 27.12 & 0.95 & 36.58 & \textcolor{red}{0.98} & 14.63 & 0.49 & 22.17 & 0.72 & 23.28 & 0.73 & 25.29 & 0.84 
\\

DGSolver \cite{DGSolver}                                        & NIPS'25                                         & 36.41 & 0.98 & 31.78 & 0.95 & 36.02 & 0.97 & 26.98 & 0.85 & 22.69 & 0.79 & 32.15 & 0.92 & 29.87 & 0.89 
\\

AWRaCLe \cite{AWRaCLe}                                         & AAAI'25                                         & 12.30 & 0.71 & 28.00 & 0.93 & 35.48 & 0.97 & 27.29 & 0.86 & 28.01 & 0.88 & \textcolor{red}{31.78} & 0.91 & 31.06 & \textcolor{red}{0.91}
\\

DFPIR \cite{DFPIR}                                           & CVPR'25                                         & 10.56 & 0.66 & 12.68 & 0.71 & 24.09 & 0.84 & 11.06 & 0.38 & 19.17 & 0.65 & 19.69 & 0.68 & 19.92 & 0.74 
\\

DA-RCOT \cite{DA-RCOT}                                         & TPAMI'25                                        & 19.20 & 0.87 & 27.45 & 0.93 & 
\textcolor{red}{37.92} & 
\textcolor{red}{0.98} & 
14.85 & 
0.50 & 
21.68 & 
0.68 & 
23.55 & 
0.71 & 
24.70 & 
0.79 
\\

CPLIR \cite{CPLIR}                                           & TPAMI'25                                        & 12.06 & 0.69 & 16.62 & 0.77 & 30.32 & 0.92 & 17.71 & 0.60 & 22.63 & 0.80 & 26.78 & 0.84 & 27.69 & 0.87 
\\

AdaIR \cite{AdaIR}                                           & ICLR'25                                         & 18.99 & 0.88 & 26.80 & 0.93 & 37.88 & \textcolor{red}{0.98} & 15.59 & 0.53 & 22.40 & 0.73 & 24.03 & 0.74 & 25.54 & 0.83 
\\

AgenticIR \cite{AgenticIR}                                       & ICLR'25                                         & 
33.46 & 
\textcolor{red}{0.99} & 
20.43 & 
0.86 & 
35.04 & 
0.96 & 
28.63 & 
0.88 & 
28.66 & 
0.88 & 
30.47 & 
0.89 & 
31.45 & 
0.92 
\\

\hline

Our                                              & -                                               & 
\textcolor{red}{40.65} & 
\textcolor{red}{0.99}& 
\textcolor{red}{37.01} & 
\textcolor{red}{0.99}& 
37.64 & 
\textcolor{red}{0.98}& 
\textcolor{red}{29.57} & 
\textcolor{red}{0.90}& 
\textcolor{red}{30.55} & 
\textcolor{red}{0.91}& 
31.74 & 
\textcolor{red}{0.91}& 
\textcolor{red}{32.37} & 
\textcolor{red}{0.93}
\\

\hline

\end{tabular}
}
\end{table*}

\begin{table*}[!ht]
\centering
\caption{Quantitative results of different methods on low-light image enhancement (LLIE) and deblurring datasets. Blur\_gamma denotes blurred images with gamma correction. The \textcolor{red}{best} results are marked in red.}
\label{tab_lol_blur}
\begin{tabular}{c|c|cc|cc|cc|cc|cc}
\hline
\rowcolor[HTML]{FFCCC9} 
\cellcolor[HTML]{FFCCC9}                         & 
\cellcolor[HTML]{FFCCC9}                        & 
\multicolumn{2}{c|}{\cellcolor[HTML]{FFCCC9}\textbf{LLIE}}&
\multicolumn{4}{c|}{\cellcolor[HTML]{FFCCC9}\textbf{Deblurring} (GoPro \cite{GoPro})}&
\multicolumn{4}{c}{\cellcolor[HTML]{FFCCC9}\textbf{LLIE+Deblurring} (LOL\_Blur \cite{LEDNet})}\\ \cline{3-12}

\rowcolor[HTML]{FFCCC9} 
\cellcolor[HTML]{FFCCC9}                         & 
\cellcolor[HTML]{FFCCC9}                        & 
\multicolumn{2}{c|}{\cellcolor[HTML]{FFCCC9}LOL-V1 \cite{LOL-v1}}       & 
\multicolumn{2}{c|}{\cellcolor[HTML]{FFCCC9}Blur\_gamma} & 
\multicolumn{2}{c|}{\cellcolor[HTML]{FFCCC9}Blur} & 
\multicolumn{2}{c|}{\cellcolor[HTML]{FFCCC9}Low-light} & 
\multicolumn{2}{c}{\cellcolor[HTML]{FFCCC9}Low-light+Blur}   \\ \cline{3-12}

\rowcolor[HTML]{FFCCC9} 
\multirow{-3}{*}{\cellcolor[HTML]{FFCCC9}Method} & 
\multirow{-3}{*}{\cellcolor[HTML]{FFCCC9}Venue} & 
PSNR                                                 & SSIM                                                & 
PSNR                                                 & SSIM                                                & 
PSNR                                                 & SSIM                                                & 
PSNR                                                 & SSIM                                                & 
PSNR                                                 & SSIM                                                \\

\hline
\hline

PromptIR \cite{PromptIR}                                        & NIPS'23                                         & 7.66 & 0.18 & 22.14 & 0.75 & 22.15 & 0.75 & 8.18 & 0.18 & 8.15 & 0.16 
\\

GridFormer \cite{GridFormer}                                      & IJCV'24                                         & 7.73 & 0.18 & 24.89 & 0.78 & 25.14 & 0.78 & 8.75 & 0.27 & 8.77 & 0.25 
\\

RAM \cite{RAM}                                             & ECCV'24                                        & 26.24 & 0.90 & 27.08 & 0.86 & 27.63 & 0.86 & 9.10 & 0.29 & 8.79 & 0.26 
\\

NDR-Restore \cite{NDR-Restore}                                     & TIP'24                                          & 7.61 & 0.17 & 23.72 & 0.77 & 23.84 & 0.77 & 8.48 & 0.23 & 8.45 & 0.21 
\\

DGSolver \cite{DGSolver}                                        & NIPS'25                                         & 24.89 & 0.91 & 27.16 & 0.87 & 27.91 & 0.87 & 17.38 & 0.74 & 14.67 & 0.59 
\\

AWRaCLe \cite{AWRaCLe}                                         & AAAI'25                                         & 7.55 & 0.16 & 23.72 & 0.77 & 23.85 & 0.78 & 8.57 & 0.24 & 8.60 & 0.22 
\\

DFPIR \cite{DFPIR}                                           & CVPR'25                                         & 9.96 & 0.36 & 21.95 & 0.76 & 21.71 & 0.75 & 12.46 & 0.50 & 12.53 & 0.47 
\\

DA-RCOT \cite{DA-RCOT}                                         & TPAMI'25                                        & 29.61 & 0.91 & 27.37 & 0.87 & 
28.03 & 
0.88 & 
12.57 & 
0.51 & 
10.44 & 
0.35 
\\

CPLIR \cite{CPLIR}                                           & TPAMI'25                                        & 7.79 & 0.19 & 24.30 & 0.78 & 24.52 & 0.78 & 8.73 & 0.26 & 8.73 & 0.24 
\\

AdaIR \cite{AdaIR}                                           & ICLR'25                                         & 29.33 & 0.91 & 27.01 & 0.86 & 27.47 & 0.86 & 13.31 & 0.54 & 10.61 & 0.36 
\\

AgenticIR \cite{AgenticIR}                                       & ICLR'25                                         & 
9.32 & 
0.39 & 
29.11 & 
0.92 & 
30.59 & 
0.93 & 
11.02 & 
0.50 & 
10.93 & 
0.45 
\\

\hline

Our                                              & -                                               & 
\textcolor{red}{31.95} & 
\textcolor{red}{0.93}& 
\textcolor{red}{29.97} & 
\textcolor{red}{0.94}& 
\textcolor{red}{31.81} & 
\textcolor{red}{0.94}& 
\textcolor{red}{21.44} & 
\textcolor{red}{0.88}& 
\textcolor{red}{20.82} & 
\textcolor{red}{0.83}
\\

\hline

\end{tabular}
\end{table*}

\begin{table*}[!ht]
\centering
\Large
\caption{Quantitative results of different methods on the CDD-11 \cite{OneRestore} dataset. The \textcolor{red}{best} results are marked in red.}
\label{tab_cdd11}
\resizebox{1\linewidth}{!}{
\begin{tabular}{c|cc|cc|cc|cc|cc|cc|cc|cc|cc|cc|cc}
\hline

\rowcolor[HTML]{FFCCC9} 
\cellcolor[HTML]{FFCCC9}                         & 
\multicolumn{8}{c|}{\cellcolor[HTML]{FFCCC9}\textbf{Single Degradation}}& 
\multicolumn{10}{c|}{\cellcolor[HTML]{FFCCC9}\textbf{Double Degradation}}&  
\multicolumn{4}{c}{\cellcolor[HTML]{FFCCC9}\textbf{Triple Degradation}}\\ \cline{2-23} 

\rowcolor[HTML]{FFCCC9} 
\cellcolor[HTML]{FFCCC9}                         & 
\multicolumn{2}{c|}{\cellcolor[HTML]{FFCCC9}Haze (H)}       & 
\multicolumn{2}{c|}{\cellcolor[HTML]{FFCCC9}Rain (R)} & 
\multicolumn{2}{c|}{\cellcolor[HTML]{FFCCC9}Snow (S)} & 
\multicolumn{2}{c|}{\cellcolor[HTML]{FFCCC9}Low-light (L)} & 
\multicolumn{2}{c|}{\cellcolor[HTML]{FFCCC9}H+R} & 
\multicolumn{2}{c|}{\cellcolor[HTML]{FFCCC9}H+S}& 
\multicolumn{2}{c|}{\cellcolor[HTML]{FFCCC9}L+H}& 
\multicolumn{2}{c|}{\cellcolor[HTML]{FFCCC9}L+R}& 
\multicolumn{2}{c|}{\cellcolor[HTML]{FFCCC9}L+S}&    
\multicolumn{2}{c|}{\cellcolor[HTML]{FFCCC9}L+H+R}&
\multicolumn{2}{c}{\cellcolor[HTML]{FFCCC9}L+H+S}\\ \cline{2-23}

\rowcolor[HTML]{FFCCC9} 
\multirow{-3}{*}{\cellcolor[HTML]{FFCCC9}Method} & 
PSNR                                                 & SSIM                                                & 
PSNR                                                 & SSIM                                                & 
PSNR                                                 & SSIM                                                & 
PSNR                                                 & SSIM                                                & 
PSNR                                                 & SSIM                                                & 
PSNR                                                 & SSIM                                                & 
PSNR                                                 & SSIM                                                & 
PSNR                                                 & SSIM                                                & 
PSNR                                                 & SSIM                                                & 
PSNR                                                 & SSIM                                                & 
PSNR                                                 & SSIM                                                \\

\hline
\hline

PromptIR \cite{PromptIR}                                        & 17.52 & 0.76 & 18.76 & 0.74 & 17.95 & 0.73 & 14.49 & 0.63 & 17.43 & 0.72 &  16.83 &0.73 &  15.27 &0.65 &  15.12 &0.61 &  13.80 &0.57 &  15.67 &0.64 & 14.96 &0.62 
\\

GridFormer \cite{GridFormer}                                      & 17.32 & 0.78 & 18.92 & 0.77 & 18.83 & 0.77 & 14.62 & 0.64 & 18.21 & 0.79 &  16.79 &0.76 &  15.67 &0.68 &  14.85 &0.60 &  13.34 &0.60 &  15.95 &0.66 & 15.51 &0.67 
\\

RAM \cite{RAM}                                             & 15.73 & 0.72 & 18.49 & 0.76 & 16.90 & 0.69 & 14.41 & 0.64 & 15.89 & 0.72 &  15.15 &0.68 &  15.25 &0.65 &  14.54 &0.66 &  13.87 &0.57 &  15.66 &0.65 & 14.89 &0.61 
\\

NDR-Restore \cite{NDR-Restore}                                     & 17.02 & 0.74 & 18.40 & 0.73 & 17.53 & 0.72 & 14.43 & 0.62 & 17.19 & 0.72 &  16.17 &0.72 &  15.16 &0.63 &  15.12 &0.61 &  13.78 &0.57 &  15.58 &0.63 & 14.85 &0.60 
\\

DGSolver \cite{DGSolver}                                        & 16.70 & 0.75 & 17.98 & 0.74 & 18.21 & 0.74 & 13.86 & 0.64 & 16.14 & 0.72 &  16.75 &0.74 &  13.84 &0.65 &  13.81 &0.62 &  12.79 &0.56 &  14.83 &0.65 & 13.68 &0.63 
\\

AWRaCLe \cite{AWRaCLe}                                         & 15.89 & 0.74 & 18.19 & 0.75 & 18.95 & 0.74 & 13.50 & 0.62 & 15.81 & 0.73 &  15.59 &0.72 &  14.35 &0.65 &  13.46 &0.59 &  12.46 &0.58 &  15.59 &0.66 & 15.13 &0.65 
\\

DFPIR \cite{DFPIR}                                           & 14.85 & 0.69 & 16.47 & 0.69 & 15.62 & 0.68 & 14.93 & 0.62 & 15.25 & 0.69 &  14.25 &0.66 &  14.98 &0.64 &  15.55 &0.60 &  14.20 &0.55 &  15.19 &0.63 & 14.82 &0.61 
\\

DA-RCOT \cite{DA-RCOT}                                         & 15.10 & 0.72 & 17.40 & 0.73 & 
17.00 & 
0.72 & 
13.33 & 
0.64 & 
15.05 & 
0.70 &  14.65 &0.70 &  11.36 &0.60 &  15.70 &0.66 &  15.15 &0.62 &  11.56 &0.60 & 11.49 &0.59 
\\

CPLIR \cite{CPLIR}                                           & 16.02 & 0.71 & 18.58 & 0.77 & 17.84 & 0.71 & 15.27 & 0.66 & 16.31 & 0.72 &  15.51 &0.67 &  15.70 &0.65 &  15.33 &0.70 &  14.76 &0.62 &  16.09 &0.67 & 15.28 &0.61 
\\

AdaIR \cite{AdaIR}                                           & 17.10 & 0.75 & 18.40 & 0.73 & 17.72 & 0.72 & 14.49 & 0.63 & 16.91 & 0.71 &  16.50 &0.72 &  14.88 &0.64 &  15.25 &0.62 &  14.02 &0.57 &  15.34 &0.64 & 14.58 &0.61 
\\

AgenticIR \cite{AgenticIR}                                       & 
11.99 & 
0.36 & 
12.68 & 
0.31 & 
12.31 & 
0.37 & 
10.62 & 
0.29 & 
12.00 & 
0.29 &  11.94 &0.37 &  10.60 &0.28 &  11.16 &0.28 &  10.68 &0.28 &  11.07 &0.25 & 10.37 &0.28 
\\

\hline

Our                                              & 
\textcolor{red}{28.37} & 
\textcolor{red}{0.92}& 
\textcolor{red}{29.57} & 
\textcolor{red}{0.92}& 
\textcolor{red}{29.39} & 
\textcolor{red}{0.91}& 
\textcolor{red}{27.24} & 
\textcolor{red}{0.88}& 
\textcolor{red}{28.10} & 
\textcolor{red}{0.92}&  
\textcolor{red}{27.83} &
\textcolor{red}{0.91}&  
\textcolor{red}{26.46} &
\textcolor{red}{0.87}&  
\textcolor{red}{27.70} &
\textcolor{red}{0.88}&  
\textcolor{red}{27.02} &
\textcolor{red}{0.86}&  
\textcolor{red}{26.46} &
\textcolor{red}{0.87}& 
\textcolor{red}{25.95} &
\textcolor{red}{0.86}
\\

\hline

\end{tabular}
}
\end{table*}

\subsection{Quantitative Assessment}
We quantify the performance of different IR methods using PSNR and SSIM \cite{PSNR_SSIM} metrics.
As reported in Table \ref{tab_snow_noise}, our PaAgent yields the highest scores in both desnowing and denoising tasks, which reflects its superior capabilities in recovering structural details and preserving color fidelity.
Moreover, even under extreme noise conditions (e.g., $\sigma=50$), the proposed PaAgent still maintains a leading PSNR of 27.03 dB.
As summarized in Table \ref{tab_haze_rain}, compared with other competitors, our PaAgent achieves optimal or suboptimal results in both dehazing and deraining tasks, demonstrating its superior capability in effectively removing haze and rain streaks.
From the results in Table \ref{tab_lol_blur}, our method consistently achieves the highest performance in both low-light image enhancement (LLIE) and deblurring tasks, suggesting richer image details and better visibility of our results.
Furthermore, in the more challenging LLIE+deblurring task, our method maintains leading performance, which demonstrates the effectiveness of our PaAgent in addressing composite degradations. 
To further evaluate the robustness of our PaAgent in handling complex degradation scenes, we conduct experiments on the CDD-11 dataset. 
As shown in Table \ref{tab_cdd11}, our PaAgent consistently outperforms all competitors across all 11 evaluation scenarios, including single, double, and triple degradations.
This significant performance gap proves its generalization capability in complex degradations.
The comprehensive quantitative results across the aforementioned diverse degradation tasks demonstrate that our PaAgent offers a powerful solution for IR.

\begin{table}[t]
    \centering
    \Large
    \caption{The average PSNR and SSIM scores of ablation study of the tool portrait bank and the subjective-objective reinforcement learning strategy on the CDD-11 \cite{OneRestore} dataset. The \textcolor{red}{best} results are marked in red.}
    \label{tab_Ablation}
    \resizebox{1\linewidth}{!}{
    \begin{tabular}{c|c|c|c|c|c|c|c|c}
        \hline
        \rowcolor[HTML]{FFCCC9} &  
        \multicolumn{4}{c|}{\textbf{Double Degradation}}&
        \multicolumn{4}{c}{\textbf{Triple Degradation}}\\ \cline{2-9}
        
        \rowcolor[HTML]{FFCCC9} &  
        \multicolumn{2}{c}{H+S}& 
        \multicolumn{2}{c|}{L+S}&
        \multicolumn{2}{c}{L+H+R}& 
        \multicolumn{2}{c}{L+H+S}\\ \cline{2-9}

        \rowcolor[HTML]{FFCCC9} 
        \multirow{-3}{*}{\cellcolor[HTML]{FFCCC9}Method}&  
        PSNR& SSIM& 
        PSNR& SSIM&
        PSNR& SSIM&
        PSNR& SSIM\\ \cline{2-9}
        
        \hline \hline
        
        \textit{w/ RS}&         19.33& 0.71& 18.27& 0.70& 17.57& 0.68& 17.39& 0.65\\
        \textit{w/ MS}&         23.36& 0.74& 22.94& 0.72& 21.58& 0.70& 20.09& 0.68\\
        \textit{w/ SFT}&        25.31& 0.86& 24.84& 0.78& 24.10& 0.75& 23.59& 0.71\\
        \textit{w/ SE}&         26.68& 0.88& 26.01& 0.82& 25.53& 0.81& 24.21& 0.79\\
        
        \textit{full model}&        
        \textcolor{red}{27.83}& \textcolor{red}{0.91}& 
        \textcolor{red}{27.02} & \textcolor{red}{0.86}&  
        \textcolor{red}{26.46} & \textcolor{red}{0.87}& 
        \textcolor{red}{25.95} & \textcolor{red}{0.86}\\
        
        \hline
    
    \end{tabular}
    }
\end{table}

\begin{figure*}[!ht]
    \Large
    \centering
    \resizebox{1\linewidth}{!}{
        \begin{tabular}{c@{ }c@{ }c@{ }c@{ }c@{ }c@{ }c@{ }}
            \includegraphics[height=3cm,width=4cm]{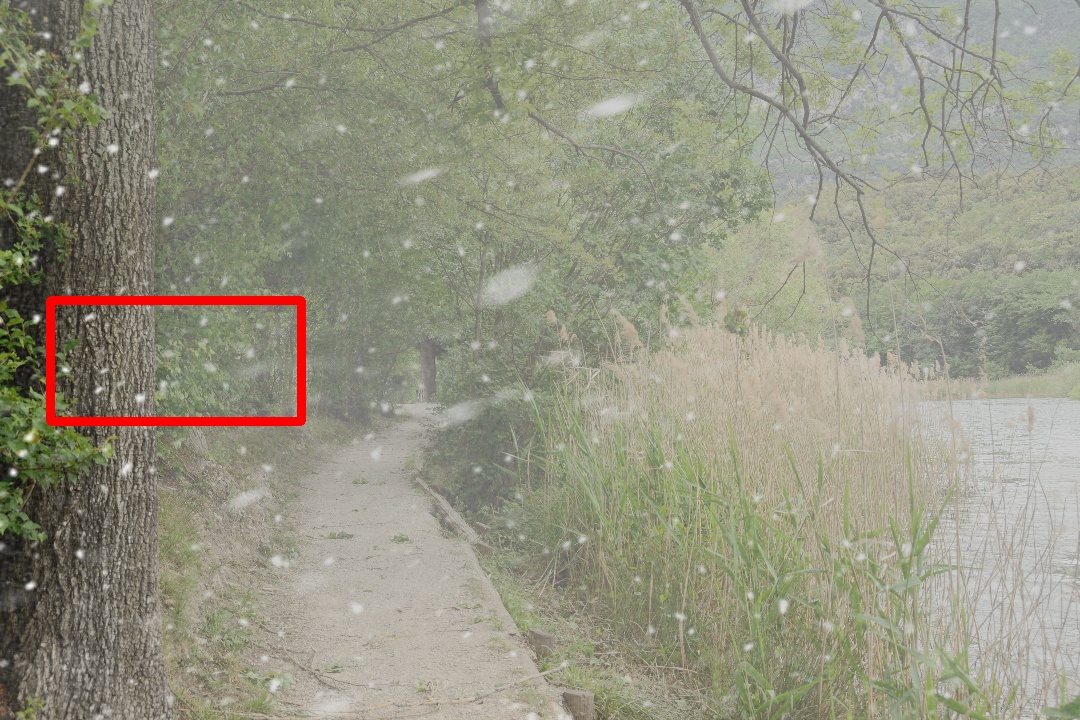} &
            \includegraphics[height=3cm,width=4cm]{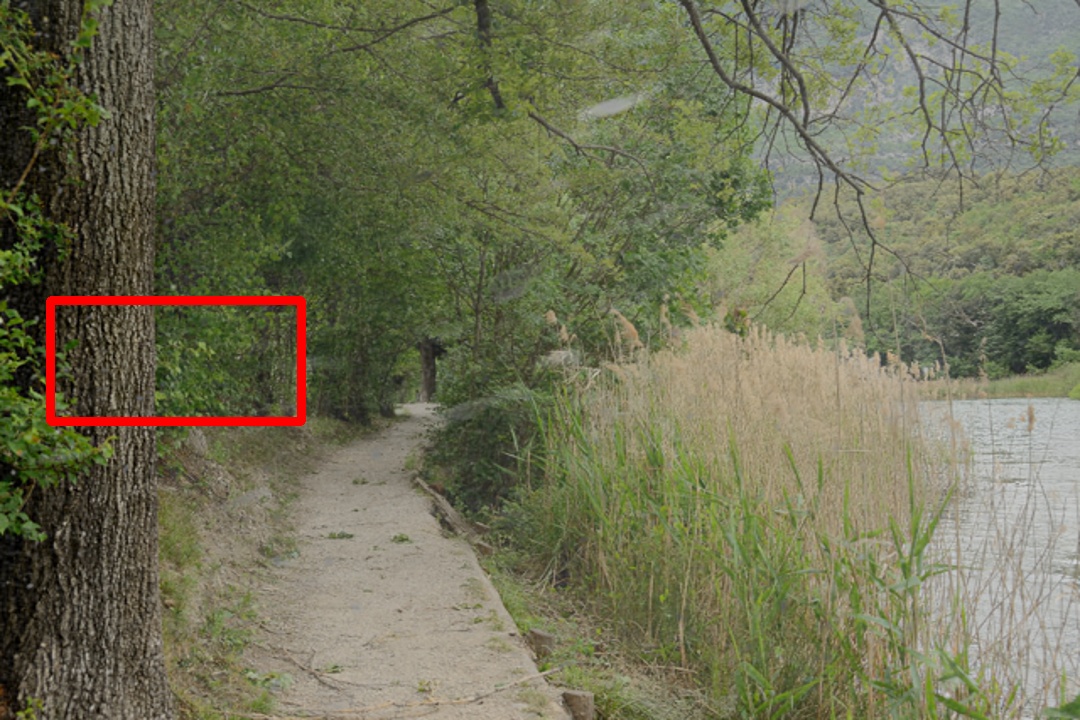} &
            \includegraphics[height=3cm,width=4cm]{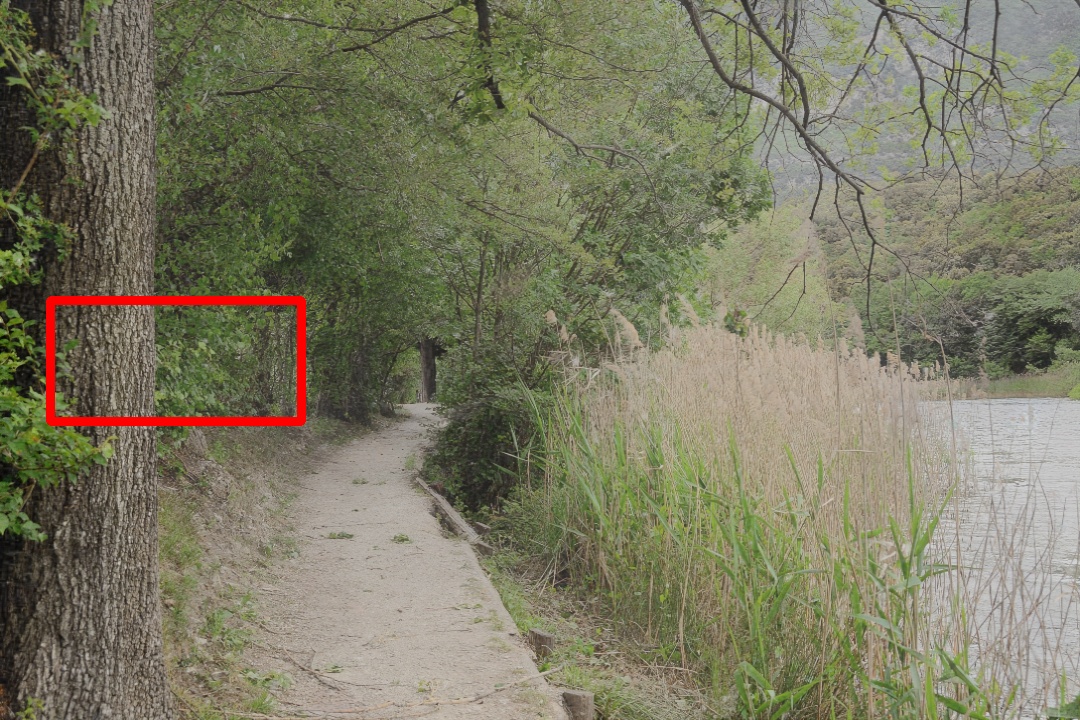} &
            \includegraphics[height=3cm,width=4cm]{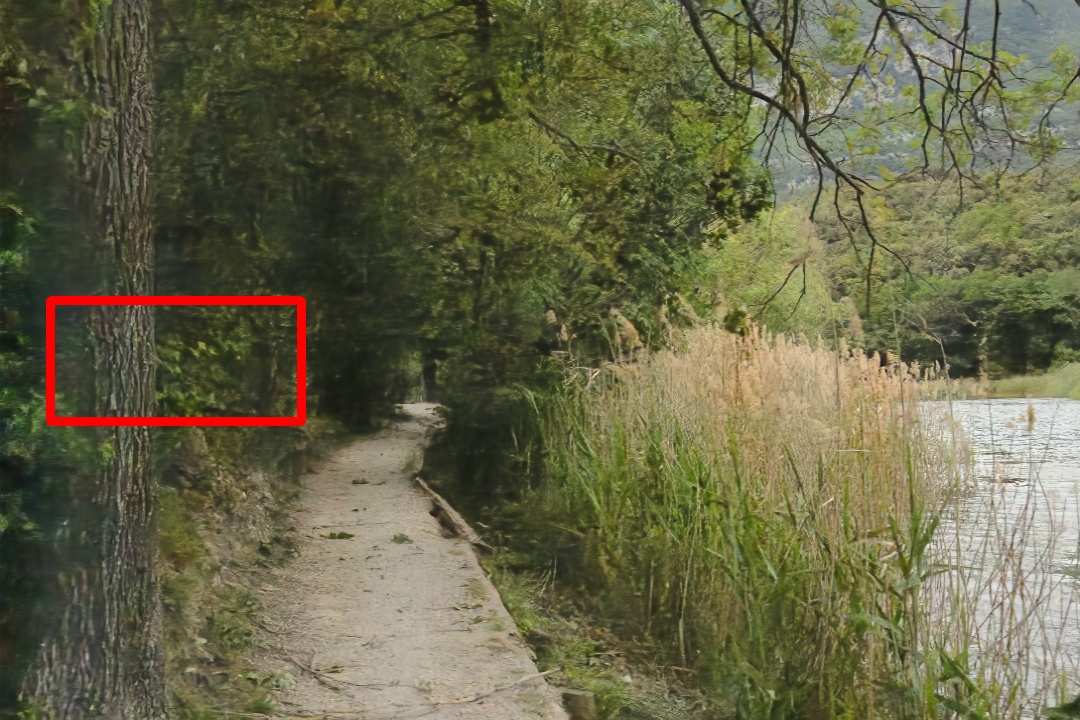} &
            \includegraphics[height=3cm,width=4cm]{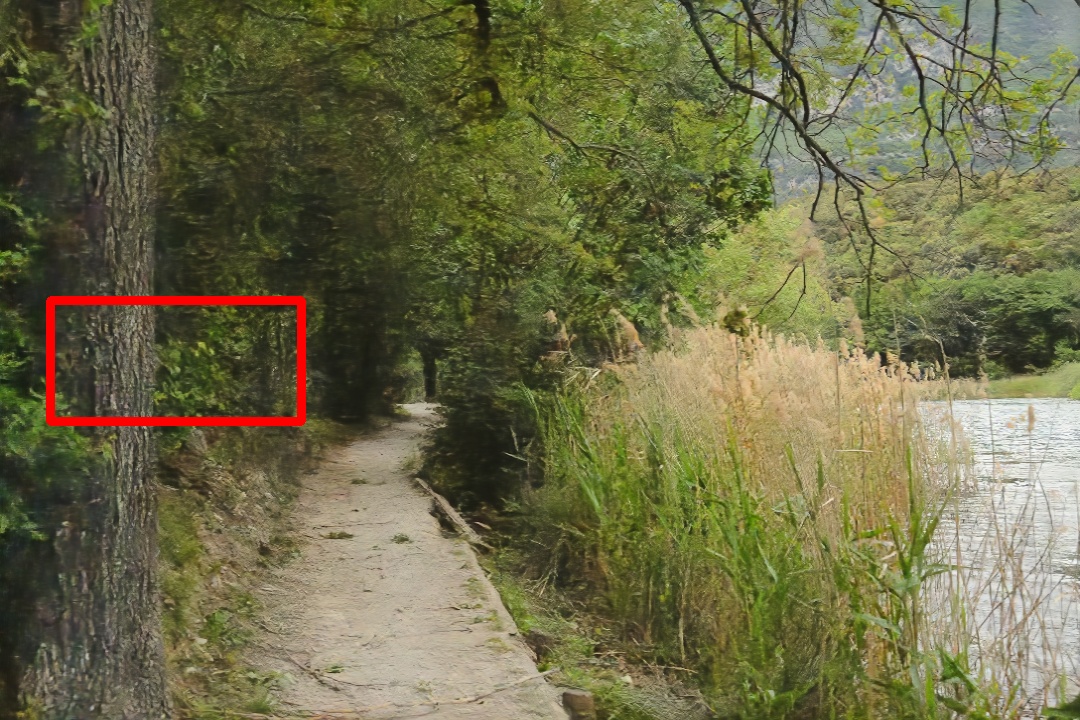} &
            \includegraphics[height=3cm,width=4cm]{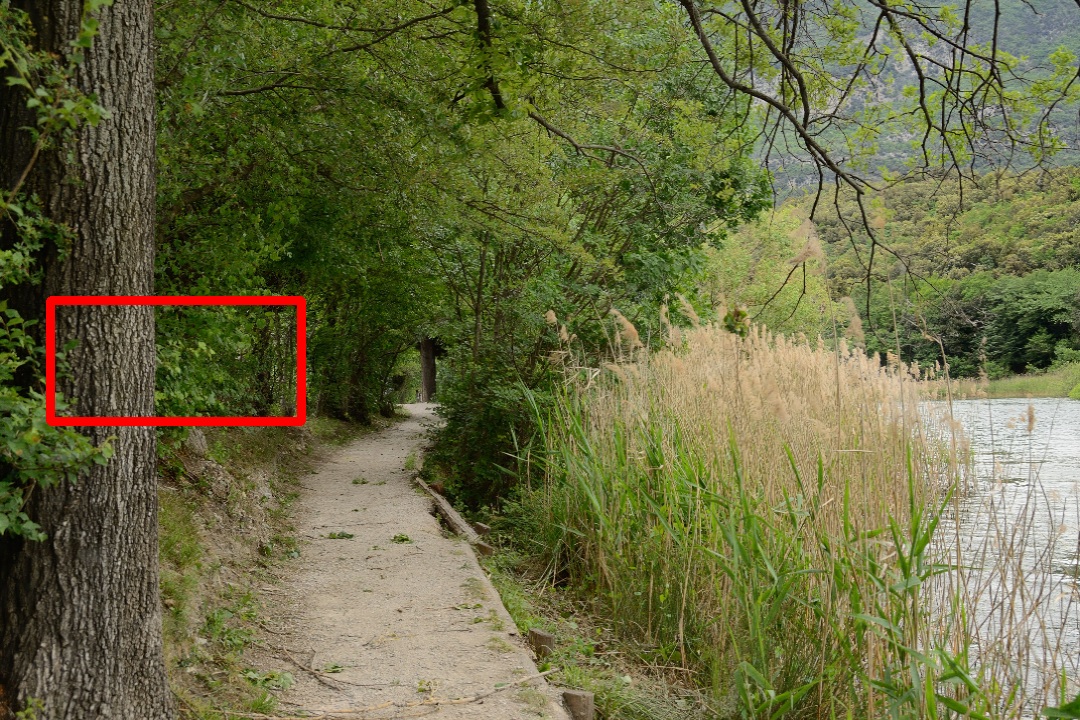} &
            \includegraphics[height=3cm,width=4cm]{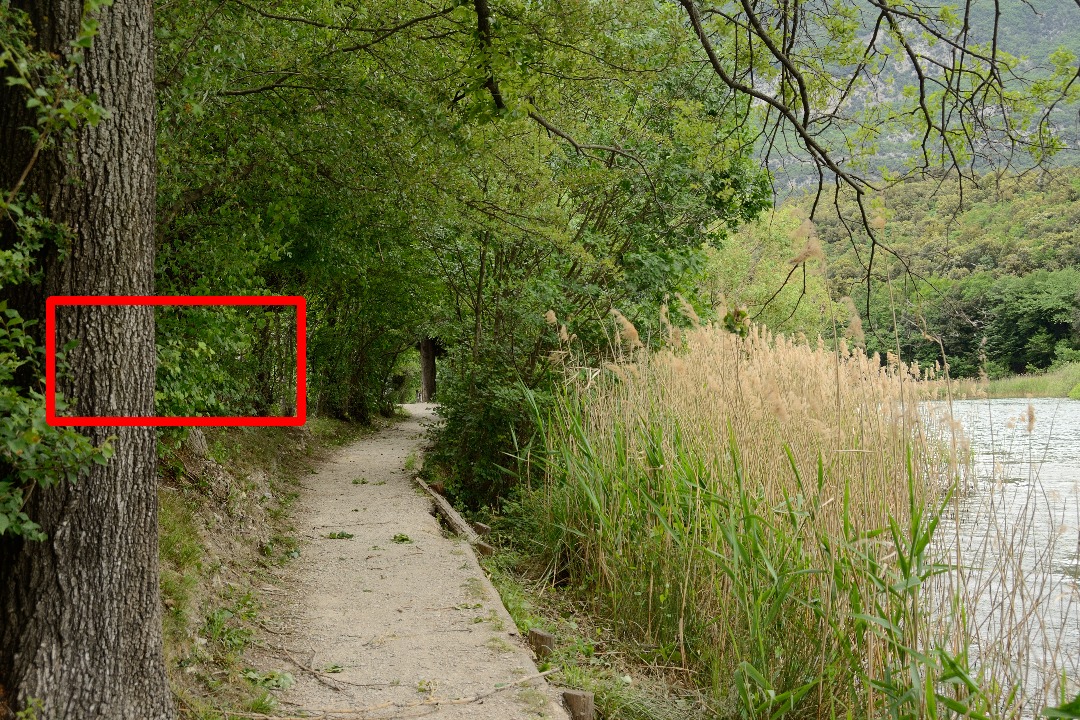} \\

            \includegraphics[height=2cm,width=4cm]{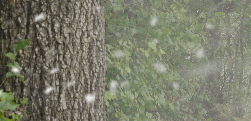} &
            \includegraphics[height=2cm,width=4cm]{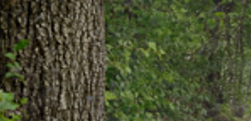} &
            \includegraphics[height=2cm,width=4cm]{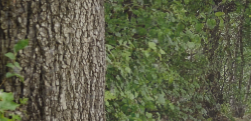} &
            \includegraphics[height=2cm,width=4cm]{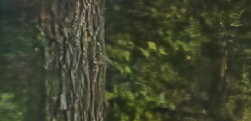} &
            \includegraphics[height=2cm,width=4cm]{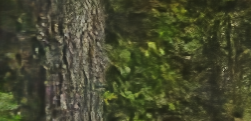} &
            \includegraphics[height=2cm,width=4cm]{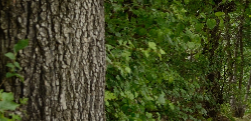} &
            \includegraphics[height=2cm,width=4cm]{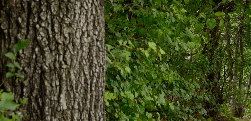} \\

            \includegraphics[height=3cm,width=4cm]{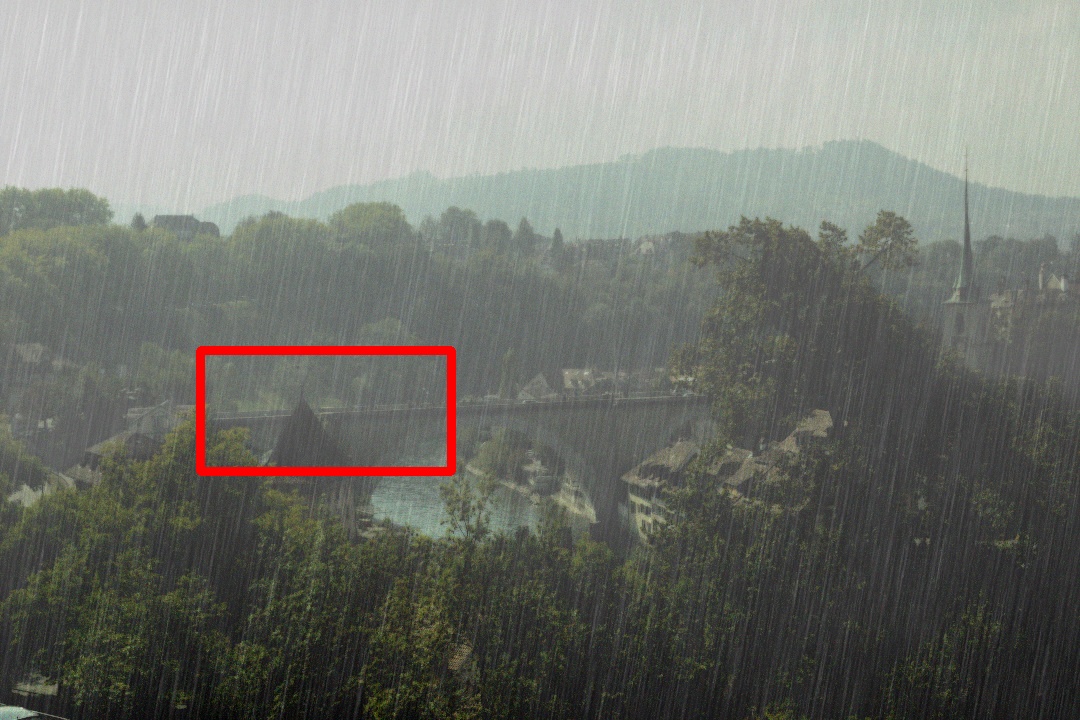} &
            \includegraphics[height=3cm,width=4cm]{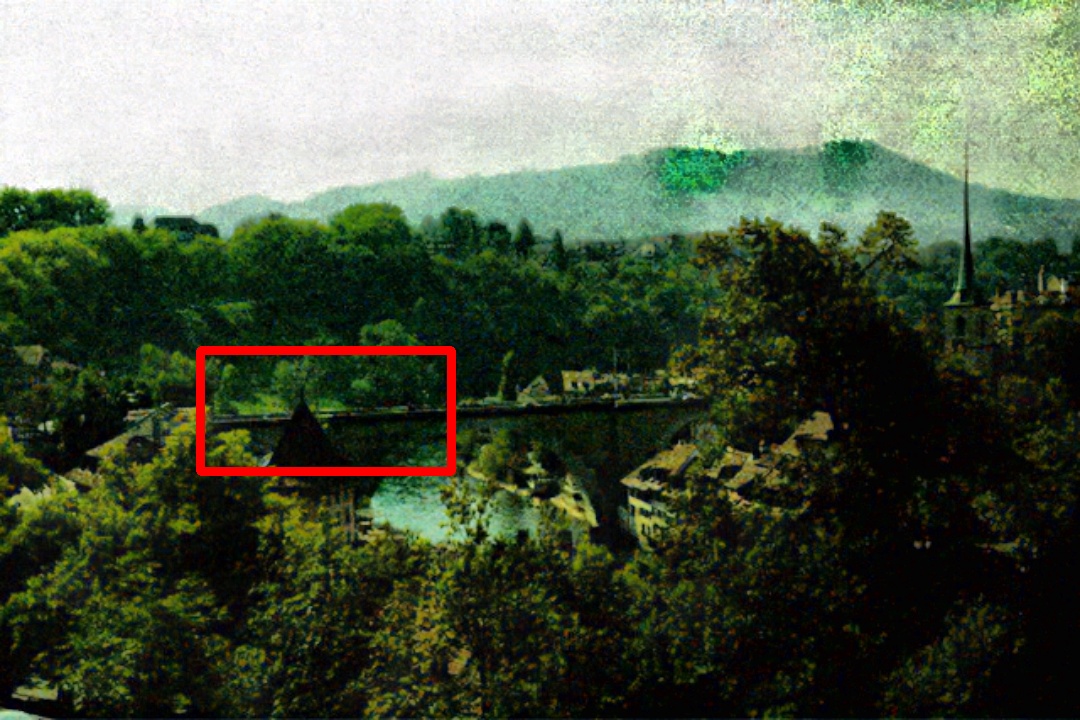} &
            \includegraphics[height=3cm,width=4cm]{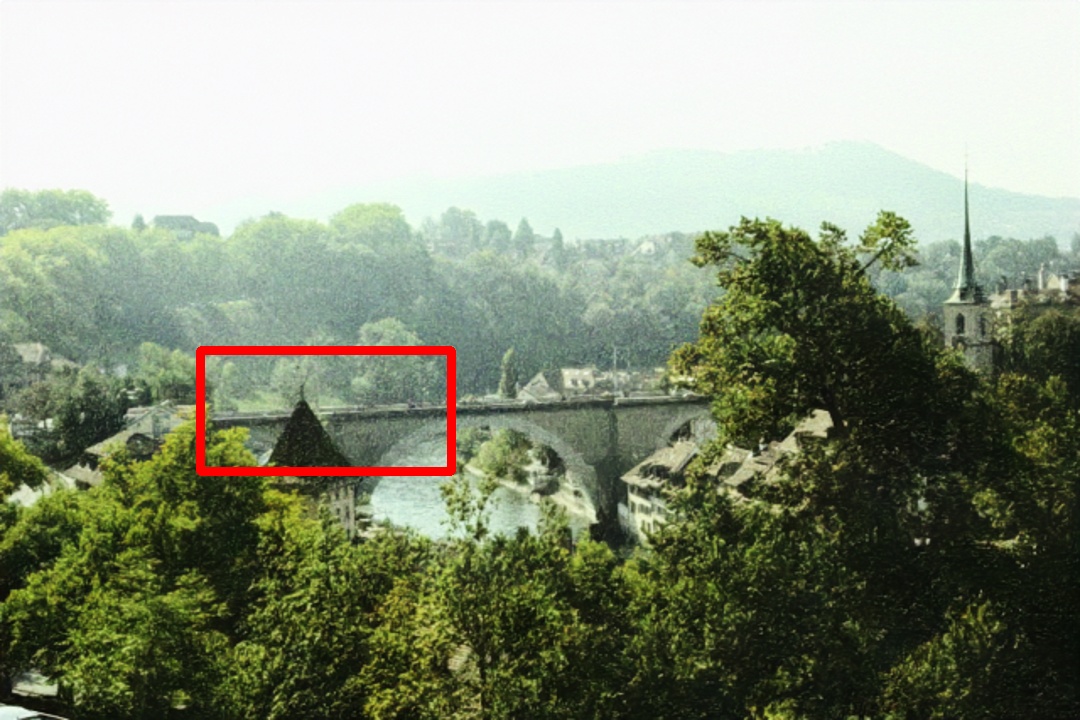} &
            \includegraphics[height=3cm,width=4cm]{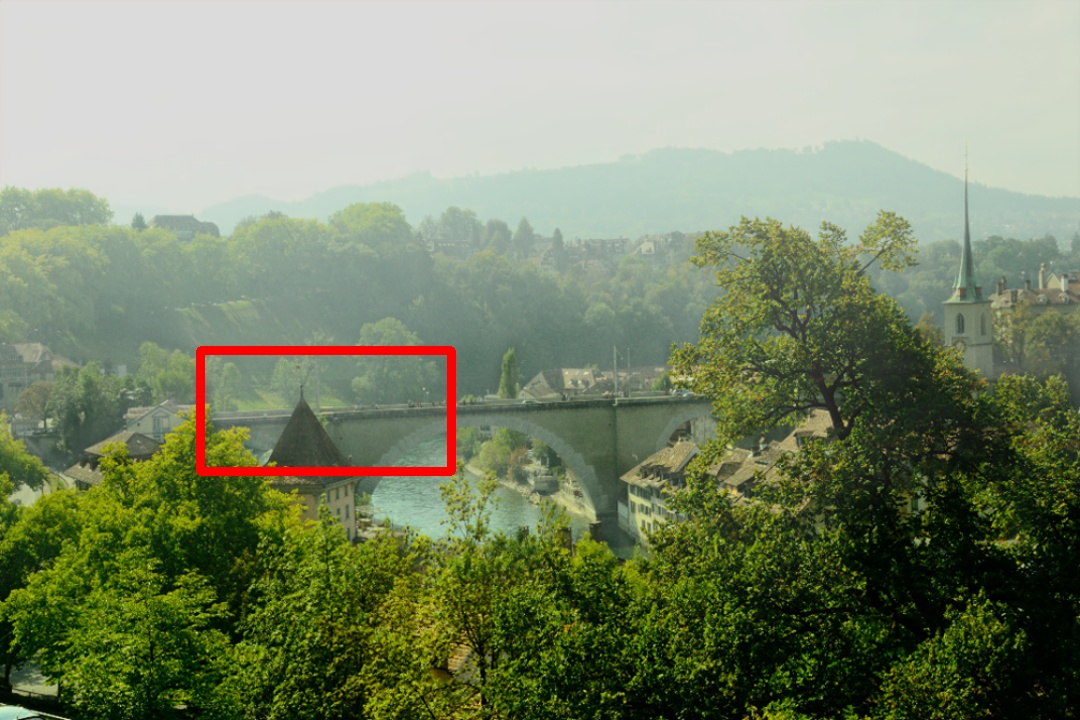} &
            \includegraphics[height=3cm,width=4cm]{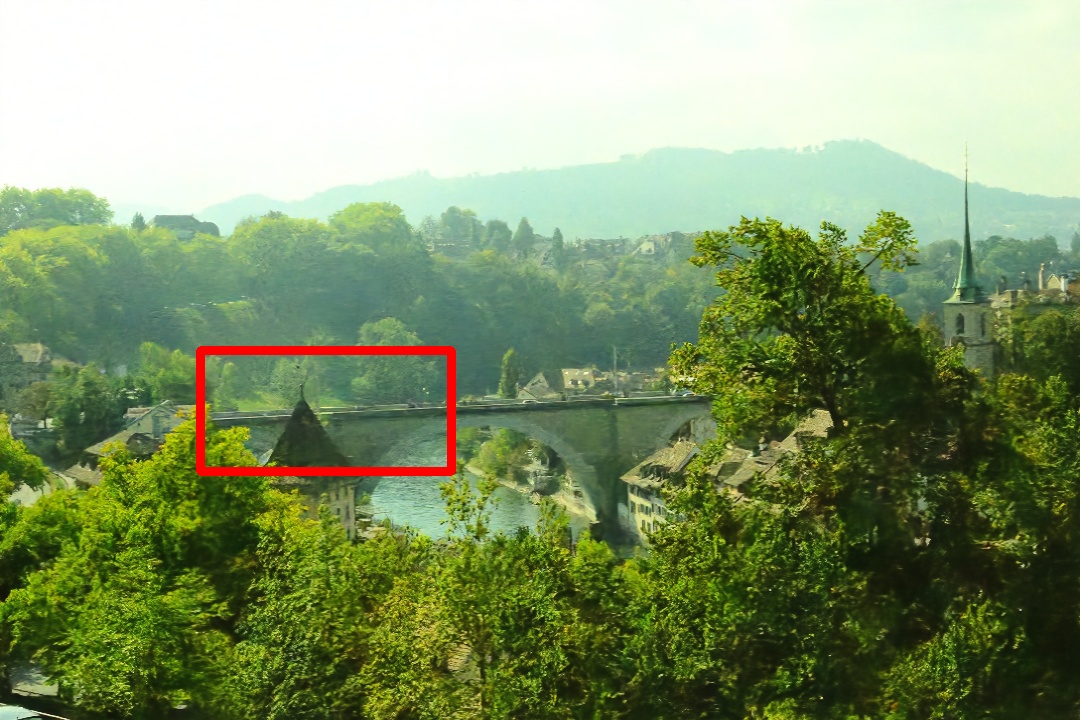} &
            \includegraphics[height=3cm,width=4cm]{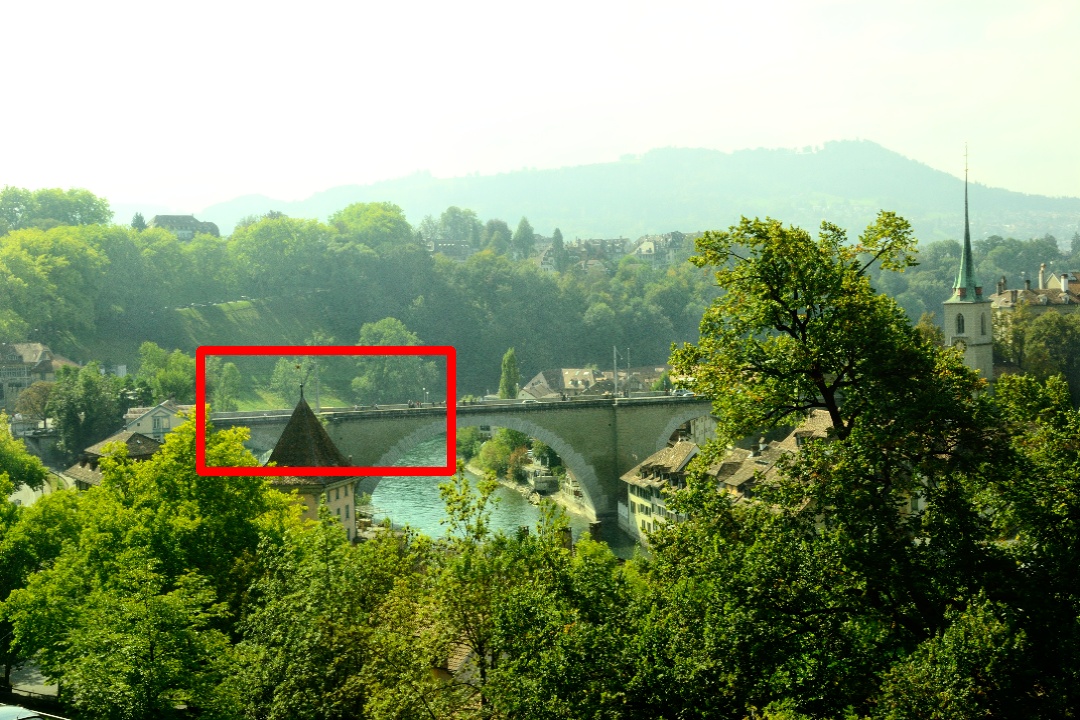} &
            \includegraphics[height=3cm,width=4cm]{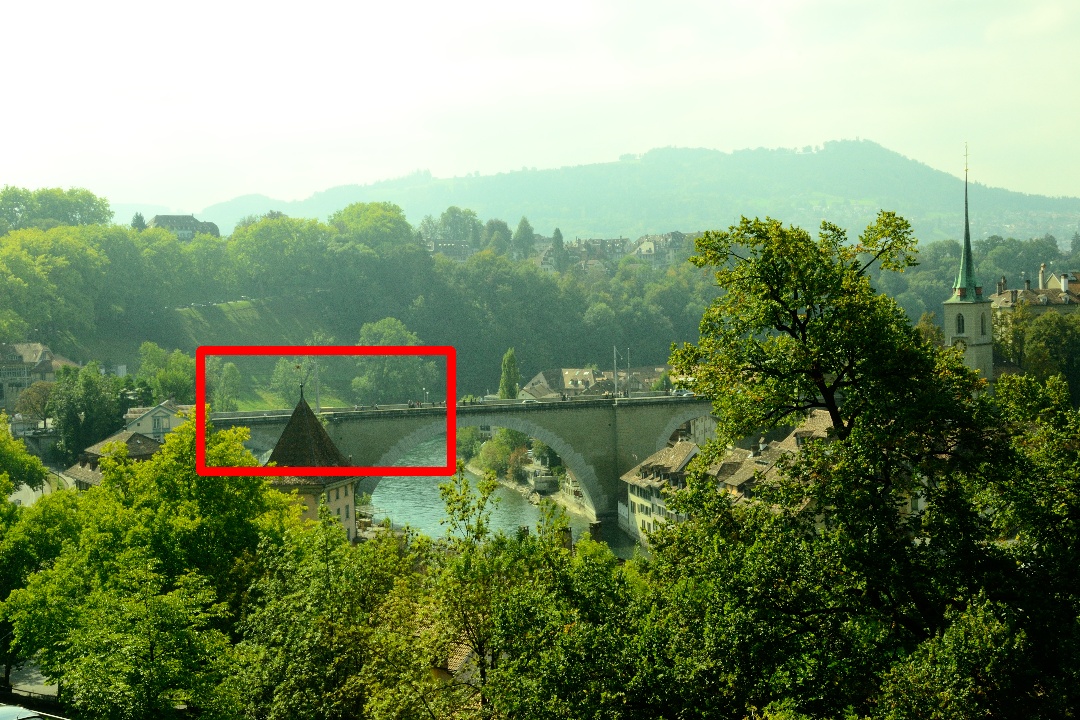} \\

            \includegraphics[height=2cm,width=4cm]{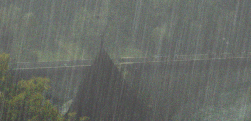} &
            \includegraphics[height=2cm,width=4cm]{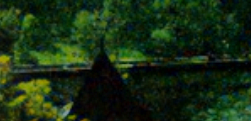} &
            \includegraphics[height=2cm,width=4cm]{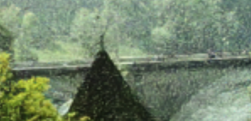} &
            \includegraphics[height=2cm,width=4cm]{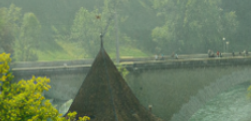} &
            \includegraphics[height=2cm,width=4cm]{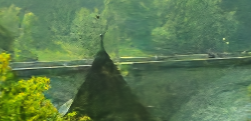} &
            \includegraphics[height=2cm,width=4cm]{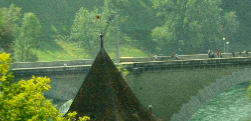} &
            \includegraphics[height=2cm,width=4cm]{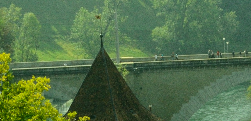} \\

            (a) Input & 
            (b) \textit{w/ RS} &
            (c) \textit{w/ MS} & 
            (d) \textit{w/ SFT} & 
            (e) \textit{w/ SE} & 
            (f) \textit{full model} & 
            (g) Reference \\

        \end{tabular}
    }
    \caption{Visualization results of ablation study on the tool portrait bank and the subjective-objective reinforcement learning strategy. In the zoomed-in areas, our \textit{full model} yields better visual results on different degraded images, with outputs closer to the reference images.}
    \label{Qual_Ablation}
\end{figure*}

\begin{figure*}[ht]
    \centering
    \includegraphics[width=1\linewidth]{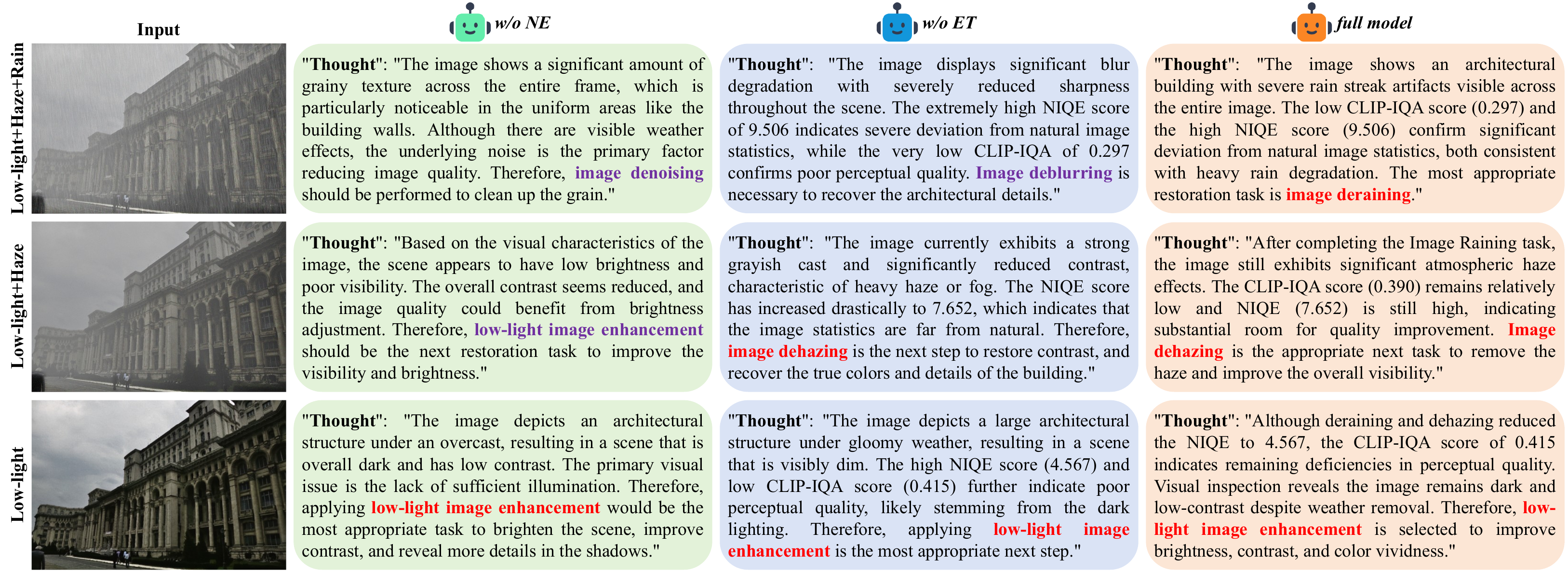}
    \caption{Ablation study on the input context of our proposed PaAgent. Incorrect and correct restoration task suggestions are marked in \textbf{{\color[HTML]{7030A0} purple}} and \textbf{\textcolor{red}{red}}, respectively. Our full model consistently generates accurate restoration strategies across various degradation scenarios.}
    \label{Qual_Abl_Input}
\end{figure*}

\subsection{Ablation Study}
We conduct an ablation study on the tool portrait bank, the subjective-objective reinforcement learning strategy, and the input context to show the effects of each key component.
The following ablation experiments are presented:
\begin{itemize}
\item \textit{w/ RS:} Randomly select one from the IR tool candidates. 
\item \textit{w/ MS:} Manually select one from the IR tool candidates by human experts.
\item \textit{w/ SFT}: Apply the SFT strategy to improve PaAgent's degradation perception capability.
\item \textit{w/ SE}: Apply the GRPO \cite{DeepSeekMath} algorithm driven by MLLMs' subjective evaluation to improve PaAgent's degradation perception capability.
\item \textit{w/o NE}: Exclude NR-IQA scores \cite{CLIP-IQA, Hyper-IQA, NIQE, CPBD, BRISQUE, MUSIQ, LIQE} and executed tasks from PaAgent's input context.
\item \textit{w/o ET}: Exclude executed tasks from PaAgent's input context.
\item \textit{w/ OSD}: Determine the image restoration sequence through the one-step decision.
\item \textit{full model}: Using our full PaAgent.
\end{itemize}

\emph{1) Effects of the Tool Portrait bank}:
The qualitative and quantitative results are shown in Fig. \ref{Qual_Ablation} and Table \ref{tab_Ablation}, respectively.
\textit{w/ RS} produces the poorest performance, as random tool selection fails to address specific degradations, leaving heavy noise and artifacts, as shown in Fig.\ref{Qual_Ablation} (b). 
In Fig.\ref{Qual_Ablation} (c), \textit{w/ MS} improves restoration performance, but it lacks the flexibility to handle complex degradations. 
In contrast, the results of our \textit{full model} are closer to the reference images in Fig. \ref{Qual_Ablation} (f).
Moreover, our \textit{full model} possesses the best PSNR and SSIM scores in Table \ref{tab_Ablation}.
This indicates that the insight-driven tool selection effectively narrows the search space to optimal restoration strategies, avoiding both the blindness of random selection and the rigidity of fixed matching schemes, thereby substantially improving restoration quality across diverse degradation scenarios.

\emph{2) Effects of the SORL strategy}:
As shown in Table \ref{tab_Ablation}, \textit{w/ SFT} yields significant gains in PSNR and SSIM compared to \textit{w/ RS} and \textit{w/ MS}.
This demonstrates that the tool portrait bank requires accurate degradation perception to effectively deploy the appropriate IR tools for specific restoration tasks.
As shown in Fig. \ref{Qual_Ablation} (e), \textit{w/ SE} recovers finer textures and more natural illumination than \textit{w/ SFT}, demonstrating that the subjective insights of MLLMs are crucial for improving PaAgent's degradation perception capability.
By contrast, our \textit{full model} eliminates the subtle color biases and over-smoothing issues sometimes found in \textit{w/ SE}, and produces results that are visually pleasing and structurally accurate.
This indicates that the SORL strategy can further enhance PaAgent’s degradation perception capability.

\emph{3) Effects of the input context}:
As shown in Fig. \ref{Qual_Abl_Input}, the agent's restoration path varies significantly depending on the richness of the input context. 
Specifically, when processing images with rain streaks, \textit{w/o NE} confuses them with noise and erroneously selects image denoising; similarly, it mistakes haze degradation for low-light conditions, resulting in fundamentally incorrect restoration strategies. 
While \textit{w/o ET} enables correct distinction between atmospheric haze and noise, the absence of executed task history still causes erroneous sequencing in multi-step restoration pipelines, such as attempting to deblur before deraining.
In contrast, our \textit{full model}, equipped with a comprehensive input context, accurately perceives degradation types and determines the optimal restoration task. 
This indicates that NR-IQA scores provide essential quantitative guidance for degradation perception, while executed task history provides sequential guidance for multi-step restoration, where both components are crucial for robust restoration planning.

\begin{figure*}[ht]
    \centering
    \includegraphics[width=1\linewidth]{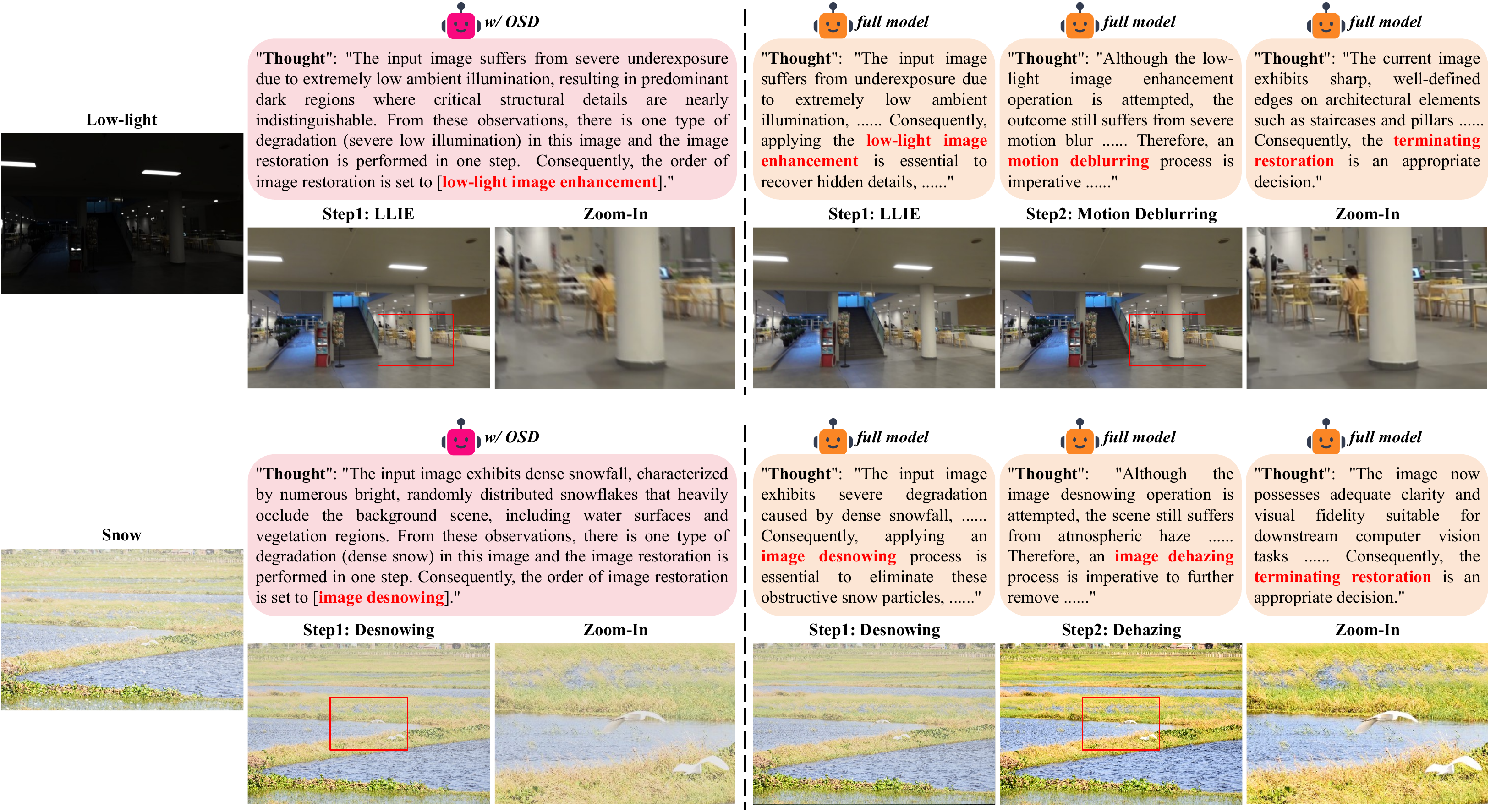}
        \addvspace{-6pt}  
    \caption{Ablation study on the multi-step decision mechanism of our PaAgent. Our full model dynamically evaluates intermediate results and triggers additional restoration steps when necessary, thereby avoiding premature termination caused by the fixed restoration sequence.}
    \label{Qual_Abl_mdm}
\end{figure*}

\emph{4) Effects of multi-step decision mechanism}:
As shown in Fig. \ref{Qual_Abl_mdm}, \textit{w/ OSD} demonstrates the capability to generate a complete restoration sequence in a single forward pass, which significantly improves computational efficiency. 
However, \textit{w/ OSD} follows a predetermined sequence without adapting to the actual restoration outcomes. 
When the enhanced image still exhibits motion blur or when the low-light enhancement produces suboptimal results, \textit{w/ OSD} proceeds to execute the next operation or terminates the restoration process, leading to incomplete restoration.
By contrast, our \textit{full model} employs the multi-step decision mechanism that dynamically evaluates the intermediate results after each restoration operation. 
As illustrated in Fig. \ref{Qual_Abl_mdm}, after completing image desnowing, our \textit{full model} detects the residual atmospheric haze and determines that an additional image dehazing step is imperative.
This progressive perceive-restore-evaluate cycle process continues until the image achieves adequate visual quality, as evidenced by the final terminating restoration decision.
These observations confirm that the multi-step decision mechanism is crucial for handling complex degradations and ensuring thorough restoration.

\section{Conclusion}
In this paper, we present PaAgent, a portrait-aware IR agent designed to boost the paradigm of intelligent IR. 
The cornerstone of our approach is a RAG-powered framework that enables precise IR tool invocation by leveraging past interaction insights.
Moreover, we propose a subjective-objective reinforcement learning strategy based on the GRPO algorithm to refine PaAgent’s degradation perception. 
This strategy aligns policy updates with both NR-IQA scores and human visual preferences, successfully achieving a fine-grained perception of degradation characteristics.
Extensive experiments across eight benchmark datasets demonstrate that our PaAgent achieves superior performance across diverse degradation scenarios, underscoring its robustness in complex degradations.

\bibliographystyle{ieeetr} 
\bibliography{ref}

\end{document}